\documentclass{article}

\usepackage{microtype}
\usepackage{graphicx}
\usepackage{booktabs}
\usepackage{hyperref}

\usepackage[utf8]{inputenc}
\usepackage[T1]{fontenc}
\usepackage{adjustbox}
\usepackage{amsfonts}
\usepackage[font=small]{subfig}
\usepackage{amsmath}
\usepackage{amssymb}
\usepackage{balance}
\usepackage{bbm}
\usepackage{bm}
\usepackage[font=small]{caption}
\usepackage{contour}
\usepackage[bottom,flushmargin]{footmisc}
\usepackage{cite}
\usepackage{enumitem}
\usepackage{float}
\usepackage{makecell}
\usepackage{mathbbol}
\usepackage{mathtools, nccmath}
\usepackage{microtype}
\usepackage{multirow}
\usepackage{nicefrac}
\usepackage{pifont}
\usepackage{setspace}
\usepackage{thmtools}
\usepackage{thm-restate}
\usepackage{url}
\usepackage{wrapfig}
\usepackage{xcolor}

\DeclareSymbolFontAlphabet{\mathbb}{AMSb}
\DeclareSymbolFontAlphabet{\mathbbl}{bbold}

\usepackage[noend]{algorithmic}
\renewcommand\algorithmiccomment[1]{
\hfill$\triangleright$ {#1}\hspace{-0.6em}
}

\usepackage[accepted]{icml2020}

\definecolor{crimson}{rgb}{0.7,0.01,0.02}
\definecolor{fern}{rgb}{0.05,0.5,0.15}
\definecolor{prussian}{rgb}{0,0.08,0.45}
\definecolor{faintgray}{gray}{0.9}
\definecolor{faintblue}{rgb}{0,0.08,0.65}

\newcommand{\pix}{\kern 0.1em}
\newcommand{\pmm}{\kern 0.25em$\pm$\kern 0.15em}
\newcommand{\pms}{\kern 0.10em$\pm$\kern 0.05em}
\newcommand{\cm}{\ding{51}}
\newcommand{\xm}{{\color{faintgray}\ding{55}}}

\newcommand{\dayum}[1]{{#1\parfillskip=0pt\par}}

\declaretheorem[name=Theorem]{retheorem}
\declaretheorem[name=Proposition,numberlike=retheorem]{reproposition}
\declaretheorem[name=Lemma,numberlike=retheorem]{relemma}

\declaretheorem[name=Definition]{redefinition}

\makeatletter
\newcommand{\raisemath}[1]{\mathpalette{\raisem@th{#1}}}
\newcommand{\raisem@th}[3]{\raisebox{#1}{$#2#3$}}
\makeatother

\usepackage[normalem]{ulem}

\contourlength{0.6pt}
\newcommand{\muline}[1]{%
\uline{\phantom{#1}}%
\llap{\contour{white}{#1}}%
}

\usepackage{etoolbox}

\usepackage{tikz}

\newrobustcmd*{\mytriangle}[1]{\tikz{\filldraw[draw=#1,fill=#1] (0,0) --
(0.15cm,0) -- (0.08cm,0.15cm);}}

\let\OLDthebibliography\thebibliography

\renewcommand\thebibliography[1]{
  \OLDthebibliography{#1}
  \setlength{\itemsep}{3.0pt plus 2pt minus 0pt}
}

\usepackage{tikz}

\usetikzlibrary{calc}
\usetikzlibrary{decorations.pathreplacing}
\usetikzlibrary{intersections}
\usetikzlibrary{patterns}
\usetikzlibrary{shapes.geometric}

\tikzset{x=1pt,y=1pt,font=\small}
\tikzset{fonttiny/.style={font=\tiny}}
\pgfdeclarepatternformonly{custom dots}
    {\pgfqpoint{-1pt}{-1pt}}
    {\pgfqpoint{2.5pt}{2.5pt}}
    {\pgfqpoint{3pt}{3pt}}{
        \pgfpathcircle{\pgfqpoint{0pt}{0pt}}{.75pt}
        \pgfpathcircle{\pgfqpoint{1.5pt}{1.5pt}}{.75pt}
        \pgfusepath{fill}}
\pgfdeclarepatternformonly{custom lines}
    {\pgfpointorigin}
    {\pgfqpoint{1.5pt}{100pt}}
    {\pgfqpoint{1.5pt}{100pt}}{
        \pgfsetlinewidth{0.75pt}
        \pgfpathmoveto{\pgfqpoint{0.75pt}{0pt}}
        \pgfpathlineto{\pgfqpoint{0.75pt}{100pt}}
        \pgfusepath{stroke}}

\newcommand{\mytitle}{Inverse Decision Modeling:\\
\mbox{Learning Interpretable Representations of Behavior}}

\icmltitlerunning{Inverse Decision Modeling}

\begin{document}

\twocolumn[
\icmltitle{\mytitle}
\icmlsetsymbol{equal}{*}

\begin{icmlauthorlist}
\vspace{0.5em}
\icmlauthor{Daniel Jarrett}{cam,equal}
\icmlauthor{Alihan H\"uy\"uk}{cam,equal}
\icmlauthor{Mihaela van der Schaar}{cam,ucla}
\vspace{0.25em}
\end{icmlauthorlist}

\icmlaffiliation{cam}{\pix Department of Applied Mathematics and Theoretical Physics, University of Cambridge, UK;}
\icmlaffiliation{ucla}{Department of Electrical Engineering, University of California, Los Angeles, USA. $^*$Authors contributed equally}
\icmlcorrespondingauthor{\hspace{-2pt}}{daniel.jarrett@maths.cam.ac.uk}
\icmlkeywords{Machine Learning, ICML}

\vskip 0.3in
]

\printAffiliationsAndNotice{}

\allowdisplaybreaks

\setcounter{footnote}{2}

\begin{abstract}

\dayum{
Decision analysis deals with modeling and enhan- cing decision processes. A principal challenge in improving behavior is in obtaining a transparent \textit{description} of existing behavior in the first place. In this paper, we develop an expressive, unifying perspective on \textit{inverse decision modeling}: a framework for learning parameterized representations of sequential decision behavior. First, we formalize the \textit{forward} problem (as a normative standard), subsuming common classes of control behavior. Second, we use this to formalize the \textit{inverse} problem (as a descriptive model), generalizing existing work on imitation/reward learning---while opening up a much broader class of research problems in behavior representation. Finally, we instantiate this approach with an example (\mbox{\textit{inverse bounded}} \textit{rational control}), illustrating how this structure enables learning (interpretable) representations of (bounded) rationality---while naturally capturing intuitive notions of suboptimal actions, biased beliefs, and imperfect knowledge of environments.}

\vspace{-0.5em}
\end{abstract}\section{Introduction}\label{sec:intro}

Modeling and enhancing decision-making behavior is a fundamental concern in computational and behavioral science, with real-world applications to healthcare \cite{li2015sequential}, economics \cite{clithero2018response}, and cognition \cite{drugowitsch2014relation}. A principal challenge in improving decision processes is in obtaining a transparent \textit{understanding} of existing behavior to begin with. In this pursuit, a key com- plication is that agents are often \textit{boundedly rational} due to biological, psychological, and computational factors \cite{wheeler2018bounded,griffiths2015rational,genewein2015bounded,augenblick2018belief,ortega2015information}, the precise mechanics of which are seldom known. As such, how can we intelligibly characterize imperfect behavior?

\dayum{
Consider the ``lifecycle'' of decision analysis \cite{keller1989role} in the real world. First, \textit{normative analysis} deals with modeling rational decision-making. It asks the question: What constitutes ideal behavior? To this end, a prevailing approach is given by von Neumann-Morgenstern's expected utility theory, and the study of optimal control is its incarnation in sequential decision-making \cite{neumann1947theory}. But judgment rendered by real-world agents is often imperfect, so \textit{prescriptive analysis} deals with improving existing decision behavior. It asks the question: How can we move closer toward the ideal? To this end, the study of decision engineering seeks to design \scalebox{0.9}{``}human-in-the-loop\scalebox{0.9}{''} techniques that nudge or assist decision-makers, such as medical guidelines and best practices \cite{mellers1998judgment}. Importantly, however, this first requires a quantitative account of current practices and the imperfections that necessitate correcting.}

\dayum{
To take this crucial first step, we must therefore start with \textit{descriptive analysis}---that is, with understanding observed decision-making from demonstration. We ask the question: What does existing behavior look like---relative to the ideal? Most existing work on imitation learning (i.e. to replicate expert actions) \cite{yue2018imitation} and apprenticeship learning (i.e. to match expert returns) \cite{abbeel2004apprenticeship} offers limited help, as our objective is instead in understanding (i.e. to interpret imperfect behavior). In particular, beyond the utility-driven nature of rationality for agent behaviors, we wish to quantify intuitive notions of \textit{boundedness}---such as the apparent flexibility of decisions, tolerance for surprise, or optimism in beliefs. At the same time, we wish that such representations be \textit{interpretable}---that is, that they be projections of observed behaviors onto parameterized spaces that are meaningful and parsimonious.}

\textbf{Contributions}~
In this paper, our mission is to explicitly relax normative assumptions of optimality when modeling decision behavior from observations.\footnote{\dayum{Our terminology is borrowed from economics: By ``descriptive'' models, we refer to those that capture \textit{observable} decision-making behavior as-is (e.g. an imitator policy in behavioral cloning), and by ``normative'' models, we refer to those that specify \textit{optimal} de- cision-making behavior (e.g. with respect to some utility function).}\vspace{-1.15em}} First, we develop an expressive, unifying perspective on \textit{inverse decision modeling}: a general framework for learning parameterized representations of sequential decision-making behavior. Specifically, we begin by formalizing the \textit{forward problem} $F$ (as a normative standard), showing that this subsumes common classes of control behavior in literature. Second, we use this to formalize the \textit{inverse problem} $G$ (as a descriptive model), showing that it generalizes existing work on imitation and reward learning. Importantly, this opens up a much broader variety of research problems in behavior representation learning---beyond simply learning optimal utility functions. Finally, we instantiate this approach with an example that we term \textit{inverse bounded rational control}, illustrating how this structure enables learning (interpretable) representations of (bounded) rationality---capturing familiar notions of decision complexity, subjectivity, and uncertainty.

\vspace{-0.5em}\section{Related Work}\label{sec:related}

\begin{table}[t]\small
\newcolumntype{A}{>{          \arraybackslash}m{2.0 cm}}
\newcolumntype{B}{>{\centering\arraybackslash}m{1.15cm}}
\newcolumntype{C}{>{\centering\arraybackslash}m{0.65cm}}
\newcolumntype{D}{>{\centering\arraybackslash}m{0.65cm}}
\newcolumntype{E}{>{\centering\arraybackslash}m{0.75cm}}
\newcolumntype{F}{>{\centering\arraybackslash}m{0.75cm}}
\newcolumntype{G}{>{\centering\arraybackslash}m{0.7 cm}}
\newcolumntype{H}{>{\centering\arraybackslash}m{0.7 cm}}
\newcolumntype{I}{>{\centering\arraybackslash}m{0.7 cm}}
\newcolumntype{J}{>{\centering\arraybackslash}m{0.7 cm}}
\newcolumntype{K}{>{\centering\arraybackslash}m{0.7 cm}}
\newcolumntype{L}{>{\centering\arraybackslash}m{0.7 cm}}
\setlength{\cmidrulewidth}{0.5pt}
\setlength\tabcolsep{0pt}
\renewcommand{\arraystretch}{0.85}
\vspace{-0.8em}
\caption{\textit{Inverse Decision Modeling}. Comparison of primary class- es of imitation$\pix$/$\pix$reward learning (IL$\pix$/$\pix$IRL) versus our prototypical example (i.e. inverse bounded rational control) as instantiations of inverse decision modeling. Constraints on agent behavior include: \smash{$^{\dagger}$}environment dynamics (extrinsic), and \smash{$^{\ddagger}$}bounded rationality (intrinsic). \textit{Legend}: deterministic (Det.), stochastic (Stoc.), subjective dynamics (Subj.), behavioral cloning (BC), distribution matching (DM), risk-sensitive (RS), partially-observable (PO), maximum entropy (ME). All terms$\pix$/$\pix$notation are developed over Sections \ref{sec:unified}--\ref{sec:example}.}
\vspace{-0.8em}
\label{tab:project}
\begin{center}
\begin{adjustbox}{max width=\linewidth}
\begin{tabular}{ACDEFGHJKLB}
\toprule
  \multirow{8}{*}{\scalebox{0.9}{\makecell[l]{\pix\textbf{Inverse}\\\pix\textbf{Decision Model}}}}
& \multicolumn{2}{c}{\scalebox{0.9}{\textbf{Extrinsic}\smash{$^{\dagger}$}\hspace{-4pt}}}
& \multicolumn{7}{c}{\scalebox{0.9}{\textbf{Intrinsic}\smash{$^{\ddagger}$}\hspace{-4pt}}}
& \multirow{8.5}{*}{\scalebox{0.9}{\hspace{1pt}\textbf{Examples}}}
\\
\cmidrule{2-3}
\cmidrule(l{0.45em}){4-10}
& \rotatebox[origin=c]{90}{\makecell{\scalebox{0.88}{Partially~~}\\[-0.5ex]\scalebox{0.88}{Controllable}}}
& ~\rotatebox[origin=c]{90}{\makecell{\scalebox{0.88}{Partially}\\[-0.5ex]\scalebox{0.88}{Observable}}}
& ~~\rotatebox[origin=c]{90}{\makecell{\scalebox{0.88}{Purposeful}\\[-0.35ex]\scalebox{0.88}{Behavior}}}
& ~\rotatebox[origin=c]{90}{\makecell{\scalebox{0.88}{Subjective}\\[-0.5ex]\scalebox{0.88}{Dynamics}}}
& \rotatebox[origin=c]{90}{\makecell{\scalebox{0.88}{Action}\\[-0.55ex]\scalebox{0.88}{Stochasticity}}}
& \rotatebox[origin=c]{90}{\makecell{\scalebox{0.88}{Knowledge}\\[-0.5ex]\scalebox{0.88}{Uncertainty}}}
& \rotatebox[origin=c]{90}{\makecell{\scalebox{0.88}{~Decision}\\[-0.55ex]\scalebox{0.88}{Complexity}}}
& \rotatebox[origin=c]{90}{\makecell{\scalebox{0.88}{Specification}\\[-0.35ex]\scalebox{0.88}{Complexity}}}
& \rotatebox[origin=c]{90}{\makecell{\scalebox{0.88}{Recognition~}\\[-0.55ex]\scalebox{0.88}{Complexity}}}
&
\\
\cmidrule{2-3}
\cmidrule(l{0.45em}){4-10}
& \smash{$\tau_{\text{env}}$}
& \smash{$\omega_{\text{env}}$}
& \smash{$\upsilon$}
& \smash{$\tau\hspace{-1pt},\hspace{-1pt}\omega$}
& \smash{$\pi$}
& \smash{$\rho,\hspace{-1pt}\sigma$}
& \smash{$\alpha$}
& \smash{$\beta$}
& \smash{$\eta$}
&
\\
\midrule
  \scalebox{0.95}{\pix BC-IL}
& ~\cm & ~\cm
& \xm & \xm
& \cm & \xm 
& \xm & \xm & \xm
& \scalebox{0.95}{\cite{pomerleau1991efficient,bain1999framework,syed2007imitation,ross2010efficient,syed2010reduction,ross2011reduction,piot2014boosted,jarrett2020strictly}}
\\
  \scalebox{0.95}{\pix Subj. BC-IL}
& ~\cm & ~\cm
& \xm & \cm
& \cm & \xm 
& \xm & \xm & \xm
& \scalebox{0.95}{\cite{huyuk2021explaining}}
\\
\midrule
  \scalebox{0.95}{\pix Det. DM-IL}
& ~\cm & ~\xm
& \xm & \xm
& \xm & \xm 
& \xm & \xm & \xm
& \scalebox{0.95}{\cite{blonde2019sample,kostrikov2019discriminator}}
\\
  \scalebox{0.95}{\pix Stoc. DM-IL}
& ~\cm & ~\xm
& \xm & \xm
& \cm & \xm 
& \xm & \xm & \xm
& \scalebox{0.95}{\cite{ho2016generative,jeon2018bayesian,ghasemipour2019understanding,ghasemipour2019divergence,ke2019imitation,ke2020wafr,kim2018imitation,xiao2019wasserstein,dadashi2021primal,kostrikov2020imitation,arenz2020non,srinivasan2020interpretable,zhang2020f,baram2016model,baram2017model}}
\\
\midrule
  \scalebox{0.95}{\pix Det. IRL}
& ~\cm & ~\xm
& \cm & \xm
& \xm & \xm 
& \xm & \xm & \xm
& \scalebox{0.95}{\cite{ng2000algorithms,syed2008game,syed2008apprenticeship,klein2011batch,mori2011model,lee2019truly,piot2017bridging}}
\\
  \scalebox{0.95}{\pix Stoc. IRL}
& ~\cm & ~\xm
& \cm & \xm
& \cm & \xm 
& \xm & \xm & \xm
& \scalebox{0.95}{\cite{klein2012inverse,klein2013cascaded,tossou2013probabilistic,jain2019model,neu2007apprenticeship,babes2011apprenticeship,ho2016model,finn2016guided,pirotta2016inverse,metelli2017compatible,tateo2017gradient,neu2009training,ramachandran2007bayesian,choi2011map,dimitrakakis2011bayesian,rothkopf2011preference,balakrishnan2020efficient,tanwani2013inverse,andrus2019inverse,belogolovsky2019learning}}
\\
  \scalebox{0.95}{\pix Subj. IRL}
& ~\cm & ~\xm
& \cm & \cm
& \cm & \xm 
& \xm & \xm & \xm
& \scalebox{0.95}{\cite{reddy2018you}}
\\
  \scalebox{0.95}{\pix RS-IRL}
& ~\cm & ~\xm
& \cm & \cm
& \xm & \cm 
& \xm & \xm & \xm
& \scalebox{0.95}{\cite{majumdar2017risk,singh2018risk}}
\\
\midrule
  \scalebox{0.95}{\pix Det. PO-IRL}
& ~\cm & ~\cm
& \cm & \xm
& \xm & \xm 
& \xm & \xm & \xm
& \scalebox{0.95}{\cite{choi2009inverse,choi2011inverse,chinaei2012inverse,bica2021learning}}
\\
  \scalebox{0.95}{\pix Stoc. PO-IRL}
& ~\cm & ~\cm
& \cm & \xm
& \cm & \xm 
& \xm & \xm & \xm
& \scalebox{0.95}{\cite{makino2012apprenticeship,jarrett2020inverse,pattanayak2020inverse}}
\\
  \scalebox{0.95}{\pix Subj. PO-IRL}
& ~\cm & ~\cm
& \cm & \cm
& \cm & \xm 
& \xm & \xm & \xm
& \scalebox{0.95}{\cite{golub2013learning,wu2018inverse,daptardar2019inverse,kwon2020inverse}}
\\
\midrule
  \scalebox{0.95}{\pix ME-IRL}
& ~\cm & ~\xm
& \cm & \xm
& \cm & \xm 
& \cm & \xm & \xm
& \scalebox{0.95}{\cite{ziebart2008maximum,boularias2011relative,kalakrishnan2013learning,wulfmeier2015maximum,finn2016connection,fu2018learning,qureshi2019adversarial,barde2020adversarial,ziebart2010modeling,zhou2017infinite,lee2018maximum,mai2019generalized}}
\\
  \scalebox{0.95}{\pix Subj. ME-IRL}
& ~\cm & ~\xm
& \cm & \cm
& \cm & \xm 
& \cm & \xm & \xm
& \scalebox{0.95}{\cite{herman2016inverse,herman2016thesis}}
\\
\midrule
  \scalebox{0.88}{\makecell[l]{\pix\textbf{Inverse Bounded}\\\pix\textbf{Rational Control}}}
& \raisebox{-3pt}{~\cm} & \raisebox{-3pt}{~\cm}
& \raisebox{-3pt}{\cm} & \raisebox{-3pt}{\cm}
& \raisebox{-3pt}{\cm} & \raisebox{-3pt}{\cm} 
& \raisebox{-3pt}{\cm} & \raisebox{-3pt}{\cm} & \raisebox{-3pt}{\cm}
& \raisebox{-3pt}{\scalebox{0.93}{\textbf{Section \ref{sec:example}}}}
\\
\bottomrule
\end{tabular}
\end{adjustbox}
\end{center}
\vspace{-1.8em}
\end{table}

\dayum{
As specific forms of descriptive modeling, imitation learning and apprenticeship learning are popular paradigms for learning policies that mimic the behavior of a demonstrator.
\textit{Imitation learning} focuses on replicating an expert's actions. Classically, ``behavioral cloning'' methods directly seek to learn a mapping from input states to output actions \cite{pomerleau1991efficient,bain1999framework,syed2007imitation}, using assistance from interactive experts or auxiliary regularization to improve generalization \cite{ross2010efficient,syed2010reduction,ross2011reduction,piot2014boosted,jarrett2020strictly}. More recently, ``distribution-matching'' methods have been proposed for learning an imitator policy whose induced state-action occupancy measure is close to that of the demonstrator \cite{blonde2019sample,kostrikov2019discriminator,ho2016generative,jeon2018bayesian,ghasemipour2019understanding,ghasemipour2019divergence,ke2019imitation,ke2020wafr,kim2018imitation,xiao2019wasserstein,dadashi2021primal,kostrikov2020imitation,arenz2020non,srinivasan2020interpretable,zhang2020f,baram2016model,baram2017model}.
\textit{Apprenticeship learning} focuses on matching the cumulative returns of the expert---on the basis of some ground-truth reward function not known to the imitator policy. This is most popularly approached by inverse reinforcement learning (IRL), which seeks to infer the reward function for which the demonstrated behavior appears most optimal, and using which an apprentice policy may itself be optimized via reinforcement learning. This includes maximum-margin methods based on feature expectations \cite{ng2000algorithms,abbeel2004apprenticeship,syed2008game,syed2008apprenticeship,klein2011batch,mori2011model,lee2019truly}, maximum likelihood soft policy matching \cite{neu2007apprenticeship,babes2011apprenticeship}, maximum entropy policies
\cite{ziebart2010modeling,zhou2017infinite,lee2018maximum,mai2019generalized,jain2019model}, and Bayesian maximum a posteriori inference \cite{ramachandran2007bayesian,choi2011map,dimitrakakis2011bayesian,rothkopf2011preference,balakrishnan2020efficient}, as well as methods that leverage preference models and additional annotations for assistance \cite{asri2016score,burchfiel2016distance,jacq2019learning,brown2019extrapolating,brown2020better}. We defer to surveys of \cite{yue2018imitation,osa2018imitation} for more detailed overviews of imitation learning and inverse reinforcement learning.}

\dayum{
\textit{Inverse decision modeling} subsumes most of the standard approaches to imitation and apprenticeship learning as specific instantiations, as we shall see (cf. Table \ref{tab:project}). Yet---with very few exceptions \cite{wu2018inverse,daptardar2019inverse,kwon2020inverse}---the vast majority of these works are limited to cases where demonstrators are assumed to be ideal or close to ideal. Inference is therefore limited to that of a single utility function; after all, its primary purpose is less for introspection than simply as a mathematical intermediary for mimicking the demonstrator's exhibited behavior. To the contrary, we seek to inspect and understand the demonstrator's behavior, rather than simply producing a faithful copy of it. In this sense, the novelty of our work is two-fold. First, we shall formally define ``inverse decision models'' much more generally as \textit{projections} in the space of \textit{behaviors}. These projections depend on our conscious choices for forward and inverse planners, and the explicit structure we choose for their parameterizations allows asking new classes of targeted research questions based on normative factors (which we impose) and descriptive factors (which we learn). Second, we shall model an agent's behavior as induced by both a \textit{recognition policy} (committing observations to internal states) and a \textit{decision policy} (emitting actions from internal states). Importantly, not only may an agent's mapping from internal states into actions be suboptimal (viz. the latter), but that their mapping from observations into beliefs may also be subjective (viz. the former). This greatly generalizes the idea of ``boundedness'' in sequential decision-making---that is, instead of commonly-assumed forms of noisy optimality, we arrive at precise notions of subjective dynamics and biased belief-updates. Appendix \ref{app:a} gives a more detailed treatment of related work.}

\vspace{-0.25em}\section{Inverse Decision Modeling}\label{sec:unified}

First, we describe our formalism for \textit{planners} (Section \ref{sub:forward}) and \textit{inverse planners} (Section \ref{sub:inverse})---together constituting our framework for inverse decision modeling (Section \ref{sub:project}). Next, we instantiate this with a prototypical example to spotlight the wider class of research questions that this unified perspective opens up (Section \ref{sec:example}). Table \ref{tab:project} summarizes related work subsumed, and contextualizes our later example.

\begin{table*}[t]\small
\newcolumntype{A}{>{          \arraybackslash}m{4.6 cm}}
\newcolumntype{B}{>{\centering\arraybackslash}m{2.4 cm}}
\newcolumntype{C}{>{\centering\arraybackslash}m{2.2 cm}}
\newcolumntype{D}{>{\centering\arraybackslash}m{2.3 cm}}
\newcolumntype{E}{>{\centering\arraybackslash}m{6.5 cm}}
\setlength{\cmidrulewidth}{0.5pt}
\setlength\tabcolsep{0pt}
\renewcommand{\arraystretch}{1.09}
\vspace{-0.85em}
\caption{\textit{Planners}. Formulation of primary classes of planner algorithms in terms of our (forward) formalism, incl. the boundedly rational planner in our example (Section \ref{sec:example}). \textit{Legend}: controlled Markov process (CMP); Markov decision process (MDP); input-output hidden Markov model (IOHMM); partially-observable (PO); Dirac delta (\smash{$\delta$}); any mapping into policies ($f$); decision-rule parameterization ($\chi$).}
\vspace{-0.8em}
\label{tab:forward}
\begin{center}
\begin{adjustbox}{max width=\textwidth}
\begin{tabular}{ACDEB}
\toprule
  ~~\textbf{Planner} ($F$)
& \textbf{Setting} ($\psi$)
& \textbf{Parameter} ($\theta$)
& \textbf{Optimization ($\pi^{*}\hspace{-2pt},\rho^{*}$)}
& \textbf{Examples}
\\
\midrule
  ~~Decision-Rule CMP Policy
& \scalebox{0.95}{$\mathcal{S},\mathcal{U},\mathcal{T}$}
& $\chi$
& \scalebox{0.95}{$\text{argmax}_{\pi}\delta(\pi-f_{\text{decision}}(\chi))$}
& \cite{pomerleau1991efficient}
\\
  ~~Model-Free MDP Learner
& \scalebox{0.95}{$\mathcal{S},\mathcal{U},\mathcal{T}$}
& $\upsilon,\gamma$
& \scalebox{0.95}{$\text{argmax}_{\pi}\mathbb{E}_{\pi,\tau_{\text{env}}}[\sum_{t}\gamma^{t}\upsilon(s_{t},u_{t})]$}
& (any RL agent)
\\
  ~~Max. Entropy MDP Learner
& \scalebox{0.95}{$\mathcal{S},\mathcal{U},\mathcal{T}$}
& $\upsilon,\gamma,\alpha$
& \scalebox{0.95}{$\text{argmax}_{\pi}\mathbb{E}_{\pi,\tau_{\text{env}}}[\sum_{t}\gamma^{t}\upsilon(s_{t},u_{t})$\pix$+$\pix$\alpha\mathcal{H}(\pi(\cdot|s_{t}))]$}
& \cite{haarnoja2017reinforcement,haarnoja2018soft,eysenbach2019if,shi2019soft}
\\
  ~~Model-Based MDP Planner
& \scalebox{0.95}{$\mathcal{S},\mathcal{U},\mathcal{T}$}
& $\upsilon,\gamma,\tau$
& \scalebox{0.95}{$\text{argmax}_{\pi}\mathbb{E}_{\pi,\tau}[\sum_{t}\gamma^{t}\upsilon(s_{t},u_{t})]$}
& (any MDP solver)
\\
  ~~Differentiable MDP Planner
& \scalebox{0.95}{$\mathcal{S},\mathcal{U},\mathcal{T}$}
& $\upsilon,\gamma,\tau$
& \scalebox{0.95}{$\text{argmax}_{\pi}\delta(\pi-\text{neural-network}(\psi,\upsilon,\gamma,\tau))$}
& \cite{shah2018inferring,shah2019feasibility}
\\
  ~~KL-Regularized MDP Planner
& \scalebox{0.95}{$\mathcal{S},\mathcal{U},\mathcal{T}$}
& $\upsilon,\gamma,\tau,\alpha,\tilde{\pi}$
& \scalebox{0.95}{$\text{argmax}_{\pi}\mathbb{E}_{\pi,\tau}[\sum_{t}\gamma^{t}(\upsilon(s_{t},u_{t})$\pix$-$\pix$\alpha D_{_{\text{KL}}}(\pi(\cdot|s_{t})\|\tilde{\pi}))]$}
& \cite{rubin2012trading,galashov2019information,ho2020efficiency,tiomkin2017unified,leibfried2017information}
\\
  ~~Decision-Rule IOHMM Policy
& \scalebox{0.95}{$\mathcal{S},\mathcal{X},\mathcal{Z},\mathcal{U},\mathcal{T},\mathcal{O}$}
& $\chi,\tau,\omega$
& \scalebox{0.95}{$\text{argmax}_{\pi}\delta(\pi-f_{\text{decision}}(\chi),\rho-f_{\text{recognition}}(\tau,\omega)$)}
& \cite{huyuk2021explaining}
\\
  ~~Model-Free POMDP Learner
& \scalebox{0.95}{$\mathcal{S},\mathcal{X},\mathcal{Z},\mathcal{U},\mathcal{T},\mathcal{O}$}
& $\upsilon,\gamma$
& \scalebox{0.95}{$\text{argmax}_{\pi,\rho\in\{\rho\text{~is black-box}\}}\mathbb{E}_{\pi,\tau_{\text{env}},\rho}[\sum_{t}\gamma^{t}\upsilon(s_{t},u_{t})]$}
& \cite{hausknecht2015deep,zhu2017improving,igl2018deep,zhang2019learning,han2019variational,futoma2020popcorn}
\\
  ~~Model-Based POMDP Planner
& \scalebox{0.95}{$\mathcal{S},\mathcal{X},\mathcal{Z},\mathcal{U},\mathcal{T},\mathcal{O}$}
& $\upsilon,\gamma,\tau,\omega$
& \scalebox{0.95}{$\text{argmax}_{\pi,\rho\in\{\rho\text{~is unbiased}\}}\mathbb{E}_{\pi,\tau,\rho}[\sum_{t}\gamma^{t}\upsilon(s_{t},u_{t})]$}
& \cite{smallwood1973optimal,hauskrecht2000value,pineau2003point,kurniawati2008sarsop}
\\
  ~~Belief-Aware $\upsilon$-POMDP Planner
& \scalebox{0.95}{$\mathcal{S},\mathcal{X},\mathcal{Z},\mathcal{U},\mathcal{T},\mathcal{O}$}
& $\upsilon_{\scalebox{0.5}{$\mathcal{Z}$}},\gamma,\tau,\omega$
& \scalebox{0.95}{$\text{argmax}_{\pi,\rho\in\{\rho\text{~is unbiased}\}}\mathbb{E}_{\pi,\tau,\rho}[\sum_{t}\gamma^{t}\upsilon_{\scalebox{0.5}{$\mathcal{Z}$}}(s_{t},z_{t},u_{t})]$}
& \cite{araya2010pomdp,fehr2018rho}
\\
\midrule
  ~~\textbf{Bounded Rational Control}
& \scalebox{0.95}{$\mathcal{S},\mathcal{X},\mathcal{Z},\mathcal{U},\mathcal{T},\mathcal{O}$}
& $\upsilon,\gamma,\alpha,\beta,$ $\eta,\tilde{\pi},\tilde{\sigma},\tilde{\varrho}$
& \scalebox{0.95}{$\text{argmax}_{\pi,\rho\in\{\rho\text{~is possibly-biased}\}}\mathbb{E}_{\pi,\rho}[\sum_{t}\gamma^{t}\upsilon(s_{t},u_{t})]$} \scalebox{0.95}{$-~\alpha\mathbb{I}_{\pi,\rho}[\pi;\tilde{\pi}]-\beta\mathbb{I}_{\pi,\rho}[\sigma;\tilde{\sigma}]-\eta\mathbb{I}_{\pi,\rho}[\varrho;\tilde{\varrho}]$}
& \textbf{Theorems \ref{thm:values}--\ref{thm:policies}}
\\
\midrule
  ~~\textbf{General Formulation}
& \scalebox{0.95}{$\mathcal{S},\mathcal{X},\mathcal{Z},\mathcal{U},\mathcal{T},\mathcal{O}$}
& (any)
& \scalebox{0.95}{$\text{argmax}_{\pi,\rho}\mathcal{F}_{\psi}(\pi,\rho;\theta)$}
& \textbf{Section \ref{sub:forward}}
\\
\bottomrule
\end{tabular}
\end{adjustbox}
\end{center}
\vspace{-1.2em}
\end{table*}

\vspace{-0.40em}
\subsection{Forward Problem}\label{sub:forward}

\dayum{
Consider the standard setup for sequential decision-making, where an agent interacts with a (potentially partially-obser- vable) environment.
First, let \smash{$\psi$\pix$\doteq$\pix$(\mathcal{S},\mathcal{X},\mathcal{Z},\mathcal{U},\mathcal{T},\mathcal{O})$} give the \textit{problem setting}, where $\mathcal{S}$ denotes the space of (external) environment states, $\mathcal{X}$ of environment observables, $\mathcal{Z}$ of (internal) agent states, $\mathcal{U}$ of agent actions, $\mathcal{T}$\pix$\doteq$ \smash{$\Delta(\mathcal{S})^{\mathcal{S}\times\mathcal{U}}$} of environment transitions, and \smash{$\mathcal{O}$\pix$\doteq$\pix$\Delta(\mathcal{X})^{\mathcal{U}\times\mathcal{S}}$} of environment emissions.
Second, denote with $\theta$ the \textit{planning parameter}{\kern 0.03em}: the parameterization of (subjective) factors that a planning algorithm uses to produce behavior, e.g. utility functions \smash{$\upsilon$\pix$\in$\pix$\mathbb{R}^{\mathcal{S}\times\mathcal{U}}$}\hspace{-2pt}, discount factors \smash{$\gamma$\pix$\in$\pix$[0,1)$}, or any other biases that an agent might be subject to, such as imperfect knowledge $\tau,\omega$ of true environment dynamics \smash{$\tau_{\text{env}},\omega_{\text{env}}$\pix$\in$\pix$\mathcal{T}$\pix$\times$\pix$\mathcal{O}$}. Note that access to the true dynamics is only (indirectly) possible via such knowledge, or by sampling online/from batch data. Now, a planner is a mapping producing observable behavior:}
\vspace{-0.5em}
\begin{redefinition}[name=Behavior]\upshape
Denote the space of (observation-action) trajectories with \smash{$\mathcal{H}\doteq\cup_{t=0}^{\infty}(\mathcal{X}\times\mathcal{U})^{t}\times\mathcal{X}$}. Then a \textit{behavior} $\phi$ manifests as a distribution over trajectories (indu- ced by an agent's policies interacting with the environment):
\vspace{-0.75em}
\begin{equation}
\Phi\doteq\Delta(\mathcal{H})
\end{equation}
\vspace{-2.0em}

\dayum{
Consider behaviors induced by an agent operating under a \textit{recognition policy} \smash{$\rho$\pix$\in$\pix$\Delta(\mathcal{Z})^{\mathcal{Z}\times\mathcal{U}\times\mathcal{X}}$} (i.e. committing obser- vation-action trajectories to internal states), together with a \textit{decision policy} \smash{$\pi\in\Delta(\mathcal{U})^{\mathcal{Z}}$} (i.e. emitting actions from internal states). We shall denote behaviors induced by $\pi,\rho$:}
\vspace{-1.0em}
\begin{equation}
\phi_{\pi,\rho}\big((x_{0},u_{0},...)\big)
\doteq
\mathbb{P}
\hspace{-5pt}
_{\substack{
u\sim\pi(\cdot|z)\\
s^{\prime}\sim\tau_{\text{env}}(\cdot|s,u)\\
x^{\prime}\sim\omega_{\text{env}}(\cdot|u,s^{\prime})\\
z^{\prime}\sim\rho(\cdot|z,u,x^{\prime})}}
\hspace{-5pt}
\big(h=(x_{0},u_{0},...)\big)
\end{equation}
\vspace{-1.75em}
\end{redefinition}

\dayum{(\textit{Note}: Our notation may not be immediately familiar as we seek to unify terminology across multiple fields. For reference, a summary of notation is provided in Appendix \ref{app:0}).}

\begin{redefinition}[name=Planner]\upshape\label{def:planner}
Given problem setting $\psi$ and planning parameter $\theta$, a \textit{planner} is a mapping into behaviors:
\vspace{-0.75em}
\begin{equation}
F:\Psi\times\Theta\rightarrow\Phi
\end{equation}
\vspace{-2.0em}

where $\Psi$ indicates the space of settings, and $\Theta$ the space of parameters. Often, behaviors of the form $\phi_{\pi,\rho}$ can be naturally expressed in terms of the solution to an optimization:
\vspace{-1.25em}
\begin{equation}\label{eq:forward2}
F(\psi,\theta)\doteq\phi_{\pi^{*}\hspace{-2pt},\rho^{*}}
:
\pi^{*}\hspace{-2pt},\rho^{*}
\doteq
\text{argmax}_{\pi,\rho}
\mathcal{F}_{\psi}(\pi,\rho;\theta)
\end{equation}
\vspace{-2.0em}

\dayum{
of some real-valued function \smash{$\mathcal{F}_{\psi}$} (e.g. this includes all cases where a utility function $\upsilon$ is an element of $\theta$). So, we shall write $\phi^{*}\doteq\phi_{\pi^{*}\hspace{-2pt},\rho^{*}}$ to indicate the behavior produced by $F$.}
\end{redefinition}

\dayum{
This definition is very general: It encapsulates a wide range of standard algorithms in the literature (see Table \ref{tab:forward}), including decision-rule policies and neural-network planners. Importantly, however, observe that in most contexts, a global optimizer for $\rho$ is (trivially) either an identity function, or perfect Bayesian inference (with the practical caveat, of course, that in model-free contexts actually reaching such an optimum may be difficult, such as with a deep recurrent network). Therefore in addition to just $\pi$, what Definition \ref{def:planner} makes explicit is the potential for $\rho$ to be \textit{biased}---that is, to deviate from (perfect) Bayes updates; this will be one of the important developments made in our subsequent example.}

\dayum{
Note that by equating a planner with such a mapping, we are implicitly assuming that the embedded optimization (Equation \ref{eq:forward2}) is \textit{well-defined}---that is, that there exists a single global optimum. In general if the optimization is non-trivial, this requires that the spaces of policies \smash{$\pi,\rho\in\mathcal{P}$\pix$\times$\pix$\mathcal{R}$} be suitably restricted: This is satisfied by the usual (hard-/ soft-$Q$) Boltzmann-rationality for decision policies, and by uniquely fixing the semantics of internal states as (subjective) beliefs, i.e. probability distributions over states, with recognition policies being (possibly-biased) Bayes updates.}

\dayum{
A more practical question is whether this optimum is reachable. While this may seem more difficult (at least in the most general case), for our \textit{interpretative} purposes it is rarely a problem, because (simple) human-understandable models are what we desire to be working with in the first instance. In healthcare, for example, diseases are often modeled in terms of \textit{discrete} states, and subjective beliefs over those states are eminently transparent factors that medical practitioners can readily comprehend and reason about \cite{sonnenberg1983markov,jackson2003multistate}. This is prevalent in research and practice, e.g. two-to-four states in progressive dementia \cite{obryant2008staging,jarrett2018match,jarrett2019dynamic}, cancer screening \cite{petousis2019using,cardoso2019early}, cystic fibrosis \cite{alaa2019attentive}, as well as pulmonary disease \cite{wang2014unsupervised}. Of course, this is not to say our exposition is incompatible with model-free, online settings with complex spaces and black-box approximators. But our focus here is to establish an interpretative paradigm---for which simple state-based models are most amenable to human reasoning.}

\begin{table*}[t]\small
\newcolumntype{A}{>{          \arraybackslash}m{3.55cm}}
\newcolumntype{B}{>{\centering\arraybackslash}m{2.8 cm}}
\newcolumntype{C}{>{\centering\arraybackslash}m{3.2 cm}}
\newcolumntype{D}{>{\centering\arraybackslash}m{2.55cm}}
\newcolumntype{E}{>{\centering\arraybackslash}m{5.9 cm}}
\setlength{\cmidrulewidth}{0.5pt}
\setlength\tabcolsep{0pt}
\renewcommand{\arraystretch}{1.15}
\vspace{-0.8em}
\caption{\textit{Inverse Planners}. Formulation of primary classes of identification strategies in terms of our (inverse) formalism. \textit{Legend}: value functions for $\phi$ under $\theta$ (\smash{$V_{\theta}^{\phi},Q_{\theta}^{\phi}$}); regularizer (\smash{$\zeta$}); shaped-reward error ($\Delta\upsilon$); $p$-norm (\smash{$\|\cdot\|_{p}$}); preference relation ($\prec$); $f$-divergence ($D_{f}$). Note that while our notation is general, virtually \textit{all} original works here have \smash{$\theta_{\text{desc}}$\pix$=$\pix$\upsilon$} and assume full observability (whence $\mathcal{S}$\pix$=$\pix$\mathcal{X}$\pix$=$\pix$\mathcal{Z}$).}
\vspace{-0.8em}
\label{tab:inverse}
\begin{center}
\begin{adjustbox}{max width=\textwidth}
\begin{tabular}{ACDEB}
\toprule
  ~~\textbf{Inverse Planner} ($G$)
& \textbf{Demonstrator} ($\phi_{\text{demo}}$)
& \textbf{Helper}
& \textbf{Optimization} ($\theta_{\text{desc}}^{*}$)
& \textbf{Examples}
\\
\midrule
  ~~Minimum Perturbation
& Deterministic, Optimal
& \scalebox{0.95}{Default $\tilde{\theta}_{\text{desc}}$}
& \smash{\scalebox{0.95}{$\text{argmin}_{\theta_{\text{desc}}}\hspace{-1pt}\|\theta_{\text{desc}}$$-$\pix$\tilde{\theta}_{\text{desc}}\|_{p}$\pix$:$\pix$\phi_{\text{demo}}$\pix$=$\pix$F(\psi,\theta)$}}
& \cite{heuberger2004inverse}
\\
  ~~Maximum Margin
& Deterministic, Optimal
& -
& \smash{\scalebox{0.95}{$\text{argmin}_{\theta_{\text{desc}}}\pix\mathbb{E}_{z\sim\rho_{0}}[\pix V_{\theta}^{\phi_{\text{imit}}}(z)-V_{\theta}^{\phi_{\text{demo}}}(z)\pix]$}}
& \cite{ng2000algorithms,syed2008game,syed2008apprenticeship,klein2011batch,mori2011model,lee2019truly,choi2009inverse,choi2011inverse,chinaei2012inverse,bica2021learning,ho2016model}
\\
  ~~Regularized Max. Margin
& Stochastic, Optimal
& -
& \smash{\scalebox{0.95}{$\text{argmin}_{\theta_{\text{desc}}}\mathbb{E}_{z\sim\rho_{0}}[V_{\hspace{-1pt}\text{soft},\theta}^{\phi_{\text{imit}}}(z)-V_{\theta}^{\phi_{\text{demo}}}(z)]+\zeta(\theta)$}}
& \cite{ho2016generative}
\\
  ~~Multiple Experimentation
& Deterministic, Optimal
& \scalebox{0.95}{Environments $\mathcal{V}$}
& \smash{\scalebox{0.95}{$\text{argmin}_{\theta_{\text{desc}}}\textstyle$\hspace{-0.5pt}$\int$\hspace{-0.4pt}$\textstyle\text{max}_{\mathcal{V},u}(Q_{\mathcal{V},\theta}^{\phi_{\text{demo}}}\hspace{-2pt}(z,u)$$-$$V_{\mathcal{V},\theta}^{\phi_{\text{demo}}}(z))dx$}}
& \cite{amin2016towards,amin2017repeated}
\\
  ~~Distance Minimization
& Individually-Scored
& \scalebox{0.95}{Scores \smash{$\tilde{\upsilon}(h)\in\mathbb{R}$}}
& \smash{\scalebox{0.95}{$\text{argmin}_{\theta_{\text{desc}}}\hspace{-1.5pt}\mathbb{E}_{h\sim\phi_{\text{demo}}}\|\tilde{\upsilon}(h)-\textstyle\sum_{s,u\in h}\upsilon(s,u)\|_{p}$}}
& \cite{asri2016score,burchfiel2016distance}
\\
  ~~Soft Policy Inversion
& Stoc., Batch-Ordered
& \scalebox{0.95}{\smash{$\{\phi_{\text{demo}}^{\scalebox{0.6}{($1$)}},...,\phi_{\text{demo}}^{\scalebox{0.6}{($K$)}}\}$}}
& \smash{\scalebox{0.95}{$\text{argmin}_{\theta_{\text{desc}}}\hspace{-1.5pt}\textstyle\sum_{k}\mathbb{E}_{s,u,s^{\prime}\sim\phi_{\text{demo}}^{\scalebox{0.6}{($k$)}}}$\hspace{1pt}\raisebox{-0.75pt}{$\|\Delta\upsilon^{\scalebox{0.6}{($k$)}}(s,u,s^{\prime})\|_{p}$}}}
& \cite{jacq2019learning}
\\
  ~~Preference Extrapolation
& Stoc., Pairwise-Ranked
& \scalebox{0.95}{\smash{$\{(i,j)|h_{i}\prec h_{j}\}$}}
& \smash{\scalebox{0.95}{$\text{argmin}_{\theta_{\text{desc}}}\mathbb{E}_{(h_{i}\prec h_{j})\sim\phi_{\text{demo}}}\log\mathbb{P}_{\upsilon}(h_{i}\prec h_{j})$}}
& \cite{brown2019extrapolating,brown2020better}
\\
  ~~Soft Policy Matching
& Stochastic, Optimal
& -
& \smash{\scalebox{0.95}{$\text{argmin}_{\theta_{\text{desc}}}\hspace{-5pt}\raisebox{-0pt}{$D_{\text{KL}}(\mathbb{P}_{\phi_{\text{demo}}}(u_{\scalebox{0.5}{0:$T$}}\|x_{\scalebox{0.5}{0:$T$}})\|\mathbb{P}_{\phi_{\text{imit}}}(u_{\scalebox{0.5}{0:$T$}}\|x_{\scalebox{0.5}{0:$T$}}))$}$}}
& \cite{ziebart2010modeling,zhou2017infinite,lee2018maximum,mai2019generalized,neu2007apprenticeship,babes2011apprenticeship,jain2019model,pattanayak2020inverse,klein2012inverse,klein2013cascaded,tossou2013probabilistic,herman2016inverse,herman2016thesis}
\\
  ~~Distribution Matching
& Stochastic, Optimal
& -
& \scalebox{0.95}{$\text{argmin}_{\theta_{\text{desc}}}D_{f}(\phi_{\text{demo}}\|\phi_{\text{imit}})$}
& \cite{ho2016generative,jeon2018bayesian,ghasemipour2019understanding,ghasemipour2019divergence,ke2019imitation,ke2020wafr,kim2018imitation,xiao2019wasserstein,dadashi2021primal,kostrikov2020imitation,arenz2020non,srinivasan2020interpretable,zhang2020f,baram2016model,baram2017model,ziebart2008maximum,boularias2011relative,kalakrishnan2013learning,wulfmeier2015maximum,finn2016connection,fu2018learning,qureshi2019adversarial,barde2020adversarial,kostrikov2019discriminator,blonde2019sample,finn2016guided}
\\
\midrule
  ~~\textbf{General Formulation}
& (any)
& (any)
& \scalebox{0.95}{$\text{argmin}_{\theta_{\text{desc}}}\mathcal{G}_{\psi}(\phi_{\text{demo}},\phi_{\text{imit}})$}
& \textbf{Section \ref{sub:inverse}}
\\
\bottomrule
\end{tabular}
\end{adjustbox}
\end{center}
\vspace{-1.2em}
\end{table*}

\begin{figure}[t]
\centering
\vspace{-0.8em}
\caption{\textit{Forward, Inverse, and Projection Mappings}. In the forward direction (i.e. generation): Given planning parameters $\theta$, a \textit{planner} $F$ generates observable behavior $\phi$ (Definition \ref{def:planner}). In the opposite direction (i.e. inference): Given observed behavior $\phi$, an \textit{inverse planner} $G$ infers the planning parameters $\theta$ that produced it---subject to normative specifications (Definition \ref{def:iplanner}). Finally, given observed behavior $\phi$, the composition of $F$ and $G$ gives its \textit{projection} onto the space of behaviors that are parameterizable by $\theta$ (Definition \ref{def:project}): This is the \textit{inverse decision model} (Definition \ref{def:model}).}
\label{fig:cycle}
\vspace{0.75em}
\includegraphics[width=\linewidth]{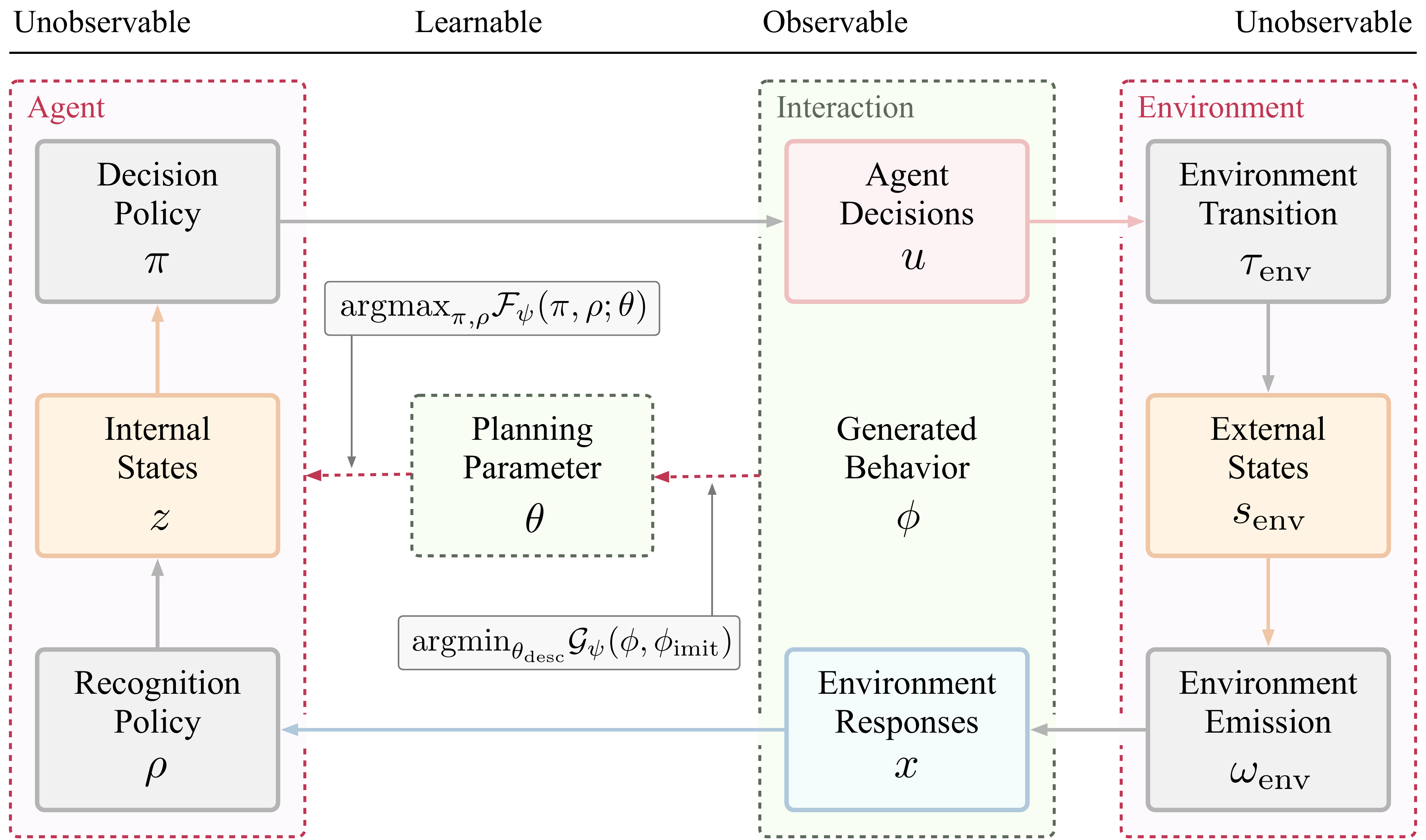}
\vspace{-2.0em}
\end{figure}

\vspace{-0.30em}
\subsection{Inverse Problem}\label{sub:inverse}

Given any setting and appropriate planner, $\theta$ gives a complete account of \smash{$\phi^{*}=F(\psi,\theta)$}: This deals with \textit{generation} ---that is, of behavior from its parameterization. In the opposite, given observed behavior $\phi_{\text{demo}}$ produced by some planner, we can ask what its $\theta$ appears to be: This now deals with \textit{inference}---that is, of parameterizations from behavior.

First, note that absent any restrictions, this endeavor immediately falls prey to the celebrated ``no free lunch'' result: It is in general \textit{impossible} to infer anything of use from $\phi_{\text{demo}}$ alone, if we posit nothing about $\theta$ (or $F$) to begin with \cite{armstrong2018occam,christiano2015medium}. The only close attempt has recruited inductive biases requiring multiple environments, and is \textit{not} interpretable due to the use of differentiable planners \cite{shah2018inferring,shah2019feasibility}.

\dayum{
On the other extreme, the vast literature on IRL has largely restricted attention to perfectly optimal agents---that is, with full visibility of states, certain knowledge of dynamics, and perfect ability to optimize $\upsilon$. While this indeed fends off the impossibility result, it is \textit{overly restrictive} for understanding behavior: Summarizing $\phi_{\text{demo}}$ using $\upsilon$ alone is not informative as to specific types of biases we may be interested in. How aggressive does this clinician seem? How flexible do their actions appear? It is difficult to tease out such nuances from just $\upsilon$---let alone comparing between agents \cite{michaud2020understanding,gleave2021quantifying}.}

We take a generalized approach to allow any middle ground of choice. While some normative specifications are required to fend off the impossibility result \cite{armstrong2018occam,shah2019feasibility}, they need not be so strong as to restrict us to perfect optimality. Formally:

\vspace{-0.50em}
\begin{redefinition}[name=Inverse Planner]\upshape\label{def:iplanner}
\dayum{
Let $\Theta\doteq\Theta_{\text{norm}}\times\Theta_{\text{desc}}$ decompose the parameter space into a \textit{normative} component (i.e. whose values $\theta_{\text{norm}}\in\Theta_{\text{norm}}$ we wish to clamp), and a \textit{descriptive} component (i.e. whose values $\theta_{\text{desc}}\in\Theta_{\text{desc}}$ we wish to infer). Then an \textit{inverse planner} is given as follows:}
\vspace{-0.75em}
\begin{equation}
G:\Phi\times\Theta_{\text{norm}}\rightarrow\Theta_{\text{desc}}
\end{equation}
\vspace{-2.1em}

Often, the descriptive parameter can be naturally expressed as the solution to an optimization (of some real-valued $\mathcal{G}_{\psi}$):
\vspace{-0.25em}
\begin{equation}
G(\phi_{\text{demo}},\theta_{\text{norm}})\doteq\text{argmin}_{\theta_{\text{desc}}}\mathcal{G}_{\psi}(\phi_{\text{demo}},\phi_{\text{imit}})
\end{equation}
\vspace{-1.25em}

where we denote by $\phi_{\text{imit}}$\pix$\doteq$\pix$F(\psi,(\theta_{\text{norm}},\theta_{\text{desc}}))$ the \textit{imitation} behavior generated on the basis of $\theta_{\text{desc}}$. So, we shall write $\theta_{\text{desc}}^{*}$\pix for the (minimizing) descriptive parameter output by\pix\pix $G$.
\end{redefinition}

\dayum{
As with the forward case, this definition is broad: It encapsulates a wide range of inverse optimization techniques in the literature (see Table \ref{tab:inverse}). Although not all techniques entail learning imitating policies in the process, by far the most dominant paradigms do (i.e. maximum margin, soft policy matching, and distribution matching). Moreover, it is \textit{normatively flexible} in the sense of the middle ground we wanted: $\theta_{\text{norm}}$ can encode precisely the information we desire.\footnote{We can verify that $\theta_{\text{desc}}$\pix$=$\pix$\upsilon$ alone recovers the usual IRL paradigm.} This opens up new possibilities for interpretative research.
%
For instance, contrary to IRL for imitation or apprenticeship, we may often \textit{not} wish to recover $\upsilon$ at all. Suppose---as an investigator---we believe that a certain $\upsilon$ we defined is the ``ought-to-be'' ideal. By allowing $\upsilon$ to be encoded in $\theta_{\text{norm}}$ (instead of $\theta_{\text{desc}}$), we may now ask questions of the form: How ``consistently'' does $\phi_{\text{demo}}$ appear to be in pursuing $\upsilon$? Does it seem ``optimistic'' or ``pessimistic'' relative to neutral beliefs about the world? All that is required is for appropriate measures of such notions (and any others) to be represented in $\theta_{\text{desc}}$. (Section \ref{sec:example} shall provide one such exemplar).}

\dayum{
Note that parameter identifiability depends on the degrees of freedom in the target $\theta_{\text{desc}}$ and the nature of the identifi- cation strategy $G$. From our generalized standpoint, we sim- ply note that---beyond the usual restrictions (e.g. on scaling, shifting, reward shaping) in conjunction with $G$---Bayesian inference remains a valid option to address ambiguities, as in \cite{jeon2018bayesian} for distribution matching,\cite{ramachandran2007bayesian,choi2011map,dimitrakakis2011bayesian,rothkopf2011preference,balakrishnan2020efficient
,makino2012apprenticeship,jarrett2020inverse}
for soft policy matching, and \cite{brown2019deep,brown2020safe} for preference extrapolation.}

\vspace{-0.30em}
\subsection{Behavior Projection}\label{sub:project}
\vspace{-0.10em}

Now we have the ingredients to formally define the business of inverse decision modeling. Compacting notation, denote $F_{\theta_{\text{norm}}}(\pix\pix\cdot\pix\pix)$\pix$\doteq$\pix$F(\psi,(\theta_{\text{norm}},\pix\cdot\pix\pix))$, and $G_{\theta_{\text{norm}}}(\pix\pix\cdot\pix\pix)$\pix$\doteq$\pix$G(\pix\pix\pix\cdot\pix\pix\pix,\theta_{\text{norm}})$. First, we require a projection operator that maps onto the sp- ace of behaviors that are \textit{parameterizable} by $\theta$ given $F_{\theta_{\text{norm}}}$:

\vspace{-0.50em}
\begin{redefinition}[name=Behavior Projection]\upshape\label{def:project}
Denote the image of $\Theta_{\text{desc}}$ under $F_{\theta_{\text{norm}}}$ by the following: $\Phi_{\theta_{\text{norm}}}\doteq F_{\theta_{\text{norm}}}[\Theta_{\text{desc}}]\leq\Phi$. Then the projection map onto this subspace is given by:
\vspace{-0.4em}
\begin{equation}
\text{proj}_{\Phi_{\theta_{\text{norm}}}}
\doteq
F_{\theta_{\text{norm}}}\circ G_{\theta_{\text{norm}}}
\end{equation}
\end{redefinition}

\vspace{-1.20em}
\begin{redefinition}[name=Inverse Decision Model]\upshape\label{def:model}
Given a specified method of parameterization $\Theta$, normative standards $\theta_{\text{norm}}$, (and appropriate planner $F$ and identification strategy $G$), the resulting \textit{inverse decision model} of $\phi_{\text{demo}}$ is given by:
\vspace{-0.5em}
\begin{equation}
\phi_{\text{imit}}^{*}
\doteq
\text{proj}_{\Phi_{\theta_{\text{norm}}}}(\phi_{\text{demo}})
\end{equation}
\vspace{-2.20em}

In other words, the model $\phi_{\text{imit}}^{*}$ serves as a complete (genera- tive) account of $\phi_{\text{demo}}$ as its \textit{behavior projection} onto $\Phi_{\theta_{\text{norm}}}$.
\end{redefinition}

\vspace{-0.50em}
\dayum{
\textbf{Interpretability}~
What dictates our choices?
For pure imitation (i.e. replicating expert actions), a black-box decision-rule fitted by soft policy matching may do well.
For apprenticeship (i.e. matching expert returns), a perfectly optimal planner inversed by distribution matching may do well.
But for \textit{understanding}, however, we wish to place appropriate structure on $\Theta$ depending on the question of interest: Precisely, the mission here is to choose some (interpretable) $F_{\theta_{\text{norm}}},G_{\theta_{\text{norm}}}$ such that $\phi_{\text{imit}}^{*}$ is amenable to human reasoning.}

\dayum{
Note that these are not passive \textit{assumptions}: We are not making the (factual) claim that $\theta$ gives a scientific explanation of the complex neurobiological processes in a clinician's head. Instead, these are active \textit{specifications}: We are making the (effective) claim that the learned $\theta$ is a parameterized ``as-if'' interpretation of the observed behavior. For instance, while there exist a multitude of commonly studied human biases in psychology, it is difficult to measure their magnitudes---much less compare them among agents. Section \ref{sec:example} shows an example of how inverse decision modeling can tackle this. (Figure \ref{fig:cycle} visualizes inverse decision modeling in a nutshell).}

\section{Bounded Rationality}\label{sec:example}

\dayum{
We wish to understand observed behavior through the lens of \textit{bounded rationality}. Specifically, let us account for the following facts: that (1) an agent's \textit{knowledge} of the environment is uncertain and possibly biased; that (2) the agent's \textit{capacity} for information processing is limited, both for decisions and recognition; and---as a result---that (3) the agent's (subjective) beliefs and (suboptimal) actions \textit{deviate} from those expected of a perfectly rational agent. We shall see, this naturally allows quantifying such notions as flexibility of decisions, tolerance for surprise, and optimism in beliefs.}

First, Section \ref{sub:rational} describes inference and control under environment uncertainty (cf. 1). Then, \ref{sub:bounded} \mbox{develops the forward} model ($F$) for agents bounded by information constraints (cf. 2--3). Finally, \ref{sub:estimate} learns parameterizations of such bo- undedness from behavior by inverse decision modeling ($G$).

\vspace{-0.30em}
\subsection{Inference and Control}\label{sub:rational}
\vspace{0.12em}

\dayum{
Consider that an agent has \textit{uncertain} knowledge of the environment, captured by a prior over dynamics \smash{$\tilde{\sigma}$\pix$\in$\pix$\Delta(\mathcal{T}$\pix$\times$\pix$\mathcal{O})$}. As a normative baseline, let this be given by some (unbiased) posterior \smash{$\tilde{\sigma}\doteq p(\tau,\omega|\mathcal{E})$}, where $\mathcal{E}$ refers to any manner of experience (e.g. observed data about environment dynamics) with which we may come to form such a neutral belief.}

\dayum{
Now, an agent may \textit{deviate} from $\tilde{\sigma}$ depending on the situation, relying instead on \smash{$\tau,\omega$\pix$\sim$\pix$\sigma(\cdot|z,u)$}---where $z,u$ allows the (biased) \smash{$\sigma$\pix$\in$\pix$\Delta(\mathcal{T}$\pix$\times$\pix$\mathcal{O})^{\mathcal{Z}\times\mathcal{U}}$} to be context-dependent. Consider recognition policies thereby parameterized by $\sigma$:}
\vspace{-0.9em}
\begin{equation}\label{eq:recognition}
\rho(z^{\prime}|z,u,x^{\prime})
\doteq
\mathbb{E}_{\tau,\omega\sim\sigma(\cdot|z,u)}\rho_{\tau,\omega}(z^{\prime}|z,u,x^{\prime})
\end{equation}

\vspace{-0.9em}
where $\rho_{\tau,\omega}$ denotes the policy for adapting $z$ to $x^{\prime}$ given (a point value for) $\tau,\omega$. For interpretability, we let $\rho_{\tau,\omega}$ be the usual Bayes belief-update. Importantly, however, $\rho$ can now effectively be biased (i.e. by $\sigma$) even while $\rho_{\tau,\omega}$ is Bayesian.

\dayum{
\textbf{Forward Process}~
The forward (``inference'') process yields the occupancy measure. First, the \textit{stepwise conditional} is:}
\vspace{-1.2em}
\begin{equation}
p(z^{\prime}|z)
=
\mathbb{E}_{\substack{
u\sim\pi(\cdot|z)\\
\tau,\omega\sim\sigma(\cdot|z,u)\\
s^{\prime}\sim\tau(\cdot|s,u)\\
x^{\prime}\sim\omega(\cdot|u,s^{\prime})
}}
\rho_{\tau,\omega}(z^{\prime}|z,u,x^{\prime})
\end{equation}

\vspace{-1.0em}
\dayum{
Define Markov operator $\mathbb{M}_{\pi,\rho}$\pix$\in$\pix$\Delta(\mathcal{Z})^{\Delta(\mathcal{Z})}$ such that for any distribution $\mu$\pix$\in$\pix$\Delta(\mathcal{Z})$ : \smash{$
(\mathbb{M}_{\pi,\rho}\mu)(z^{\prime})
\doteq
\mathbb{E}_{z\sim\mu}
p(z^{\prime}|z)
$}. Then}
\vspace{-0.7em}
\begin{equation}
\mu_{\pi,\rho}(z)
\doteq
(1-\gamma)
\textstyle\sum_{t=0}^{\infty}\gamma^{t}
p(z_{t}=z|z_{0}\sim\rho_{0})
\end{equation}

\vspace{-0.7em}
defines the \textit{occupancy measure} \smash{$\mu_{\pi,\rho}$\pix$\in$\pix$\Delta(\mathcal{Z})$} for any initial (internal-state) distribution $\rho_{0}$, and discount rate $\gamma\in[0,1)$.

\vspace{-0.5em}
\begin{relemma}[restate=forward,name=Forward Recursion]\label{thm:forward}\upshape
\dayum{Define the forward oper- ator \smash{$\mathbb{F}_{\pi,\rho}:\Delta(\mathcal{Z})\raisebox{-1pt}{$^{\Delta(\mathcal{Z})}$}$} such that for any given \smash{$\mu\in\Delta(\mathcal{Z})$}:}
\vspace{-1.0em}
\begin{equation}
(\mathbb{F}_{\pi,\rho}\mu)(z)
\doteq
(1-\gamma)\rho_{0}(z)
+
\gamma(\mathbb{M}_{\pi,\rho}\mu)(z)
\end{equation}

\vspace{-1.1em}
\dayum{
Then the occupancy $\mu_{\pi,\rho}$ is the (unique) fixed point of $\mathbb{F}_{\pi,\rho}$.}
\end{relemma}

\dayum{
\textbf{Backward Process}~
The backward (``control'') process yie- lds the value function. We want that $\mu_{\pi,\rho}$ \textit{maximize utility}:}
\vspace{-1.2em}
\begin{equation}\label{eq:rational}
\text{maximize}_{\mu_{\pi,\rho}}
J_{\pi,\rho}\doteq
\mathbb{E}_{\substack{
z\sim\mu_{\pi,\rho}\\
s\sim p(\cdot|z)\\
u\sim\pi(\cdot|z)
}}
\upsilon(s,u)
\end{equation}

\vspace{-1.0em}
Using \smash{$V$\pix$\in$\pix$\mathbb{R}^{\mathcal{Z}}$} to denote the multiplier, the Lagrangian is giv- en by $
\mathcal{L}_{\pi,\rho}(\mu,V)
\doteq
J_{\pi,\rho}
-
\langle
V
,
\mu
-
\gamma\mathbb{M}_{\pi,\rho}\mu
-
(1-\gamma)\rho_{0}
\rangle
$.

\vspace{-0.5em}
\begin{relemma}[restate=backward,name=Backward Recursion]\label{thm:backward}\upshape
Define the backward o- perator \smash{$\mathbb{B}_{\pi,\rho}:\mathbb{R}^{\mathcal{Z}}\rightarrow\mathbb{R}^{\mathcal{Z}}$} such that for any given \smash{$V\in\mathbb{R}^{\mathcal{Z}}$}:
\vspace{-0.5em}
\begin{gather}\begin{aligned}
(\mathbb{B}_{\pi,\rho}V)(z)
\doteq
\mathbb{E}_{\substack{
s\sim p(\cdot|z)\\
u\sim\pi(\cdot|z)
}}[
\upsilon(s,u)
+
\mathbb{E}\hspace{-5pt}_{\substack{
\tau,\omega\sim\sigma(\cdot|z,u)\\
s^{\prime}\sim\tau(\cdot|s,u)\\
x^{\prime}\sim\omega(\cdot|u,s^{\prime})\\
z^{\prime}\sim\rho_{\tau,\omega}(\cdot|z,u,x^{\prime})
}}\hspace{-5pt}
\gamma
V(z^{\prime})
]\end{aligned}\raisetag{0.725\baselineskip}
\end{gather}

\vspace{-1.2em}
Then the (dual) optimal $V$ is the (unique) fixed point of $\mathbb{B}_{\pi,\rho}$; this is the \textit{value function} considering knowledge uncertainty:

\vspace{-1.0em}
\begin{equation}
V^{\phi_{\pi\hspace{-1pt},\hspace{-0.5pt}\rho}}(z)
\doteq
\textstyle\sum_{t=0}^{\infty}
\gamma^{t}
\mathbb{E}\hspace{-20pt}_{\substack{
s_{t}\sim p(\cdot|z_{t})\\
u_{t}\sim\pi(\cdot|z_{t})\\
\tau,\omega\sim\sigma(\cdot|z_{t},u_{t})\\
s_{t+1}\sim\tau(\cdot|s_{t},u_{t})\\
x_{t+1}\sim\omega(\cdot|u_{t},s_{t+1})\\
z_{t+1}\sim\rho_{\tau,\omega}(\cdot|z_{t},u_{t},x_{t+1})
}}\hspace{-20pt}
[
\upsilon(s_{t},u_{t})
|z_{0}=z]
\end{equation}

\vspace{-1.0em}
\dayum{
so we can equivalently write targets \smash{$J\hspace{-1pt}_{\pi,\rho}$$=$\pix$\mathbb{E}_{z\sim\rho_{0}}\hspace{-2pt}V^{\phi_{\pi\hspace{-1pt},\hspace{-0.5pt}\rho}}(z)$}. Likewise, we can also define the (state-action) value func- tion \smash{$Q^{\phi_{\pi\hspace{-1pt},\hspace{-0.5pt}\rho}}$$\in$$\mathbb{R}^{\mathcal{Z}\times\mathcal{U}}$}---that is, \smash{$Q^{\phi_{\pi\hspace{-1pt},\hspace{-0.5pt}\rho}}\hspace{-1pt}(z,\hspace{-2pt}u)$\pix$\doteq$\pix$\mathbb{E}_{s\sim p(\cdot|z)}[\upsilon(s,\hspace{-2pt}u)$\pix$+$}
\smash{$\mathbb{E}_{\tau,\omega\sim\sigma(\cdot|z,u),...,z^{\prime}\hspace{-0.5pt}\sim\rho_{\tau,\omega}(\cdot|z,u,x^{\prime}\hspace{-0.5pt})}
\gamma
V^{\phi_{\pi\hspace{-1pt},\hspace{-0.5pt}\rho}}(z^{\prime})]$} given an action.}
\end{relemma}

\begin{table*}[t]\small
\newcolumntype{A}{>{          \arraybackslash}m{4.28cm}}
\newcolumntype{B}{>{\centering\arraybackslash}m{1.62cm}}
\newcolumntype{C}{>{\centering\arraybackslash}m{1.59cm}}
\newcolumntype{D}{>{\centering\arraybackslash}m{1.59cm}}
\newcolumntype{E}{>{\centering\arraybackslash}m{1.64cm}}
\newcolumntype{F}{>{\centering\arraybackslash}m{1.73cm}}
\newcolumntype{G}{>{\centering\arraybackslash}m{1.73cm}}
\newcolumntype{H}{>{\centering\arraybackslash}m{1.73cm}}
\newcolumntype{I}{>{\centering\arraybackslash}m{2.09cm}}
\setlength{\cmidrulewidth}{0.5pt}
\setlength\tabcolsep{0pt}
\renewcommand{\arraystretch}{1.0}
\vspace{-1.0em}
\caption{\textit{Boundedly Rational Agents}. Formulation of common decision agents as instantiations of our (boundedly rational) formalism. Note that either \smash{$\beta^{-1}$$\rightarrow$\pix$0$} or \smash{$\tilde{\sigma}$\pix$=$\pix$\delta$} is sufficient to guarantee $\forall z,u:\sigma(\cdot|z,\hspace{-2pt}u)$\pix$=$\pix$\tilde{\sigma}$.\pix \smash{$^{\dagger}$}Softmax added on top of deterministic, optimal $Q$-functions.}
\vspace{-0.8em}
\label{tab:agent}
\begin{center}
\begin{adjustbox}{max width=\textwidth}
\begin{tabular}{ACDEFGHIB}
\toprule
  \multirow{2.5}{*}{~~\textbf{Boundedly Rational Agent}}
& \textbf{Flexibility}
& \textbf{Optimism}
& \textbf{Adaptivity}~~
& \scalebox{0.97}{\raisebox{0.15pt}{\text{(\hspace{-0.5pt}Action Prior\hspace{-0.5pt})}}}
& \scalebox{0.97}{\raisebox{0.15pt}{\text{(\hspace{-0.5pt}Model Prior\hspace{-0.5pt})}}}
& \scalebox{0.97}{\raisebox{0.15pt}{\text{(\hspace{-0.5pt}Belief Prior\hspace{-0.5pt})}}}
& \multirow{2.5}{*}{~~\textbf{Observability}}
& \multirow{2.5}{*}{~~\textbf{Examples}}
\\
\cmidrule(r{0.3em}){2-4}
\cmidrule(l{0.1em}){5-7}
& \smash{$\alpha^{-1}$}
& \smash{$\beta^{-1}$}
& \smash{$\eta^{-1}$}
& \smash{$\tilde{\pi}$}
& \smash{$\tilde{\sigma}$}
& \smash{$\tilde{\varrho}$}
&
&
\\
\midrule
  ~~Uniformly Random Agent
& $\rightarrow~0~$
& $~\rightarrow~0~$
& $\rightarrow$$\scalebox{0.7}{$\pm$}\infty~$
& Uniform
& Dirac $\delta$
& -
& ~~Full\pix /\pix Partial
& -
\\
  ~~Deterministic, Optimal Agent
& $\rightarrow\infty$
& $~\rightarrow~0~$
& $\rightarrow$$\scalebox{0.7}{$\pm$}\infty~$
& -
& Dirac $\delta$
& -
& ~~Full\pix /\pix Partial
& \mbox{(any)}
\\
  ~~Boltzmann-Exploratory Agent\smash{$^{\dagger}$}
& $\rightarrow\infty$
& $~\rightarrow~0~$
& $\rightarrow$$\scalebox{0.7}{$\pm$}\infty~$
& -
& Dirac $\delta$
& -
& ~~Full\pix /\pix Partial
& \cite{heess2013actor,osband2016generalization,cesa2017boltzmann}
\\
  ~~Minimum-Information Agent
& $=~\pix1~$
& $~\rightarrow~0~$
& \pix\pix$=~~1~~$
& (any)
& Dirac $\delta$
& (any)
& ~~Full
& \cite{globerson2009minimum,tishby2011information,ortega2013thermodynamics}
\\
  ~~Maximum Entropy Agent
& $~\pix(0,\hspace{-0.5pt}\infty)$
& $~\rightarrow~0~$
& $\rightarrow$$\scalebox{0.7}{$\pm$}\infty~$
& Uniform
& Dirac $\delta$
& -
& ~~Full
& \cite{haarnoja2017reinforcement,haarnoja2018soft,eysenbach2019if,shi2019soft}
\\
  ~~(Action) KL-Regularized Agent
& $~\pix(0,\hspace{-0.5pt}\infty)$
& $~\rightarrow~0~$
& $\rightarrow$$\scalebox{0.7}{$\pm$}\infty~$
& (any)
& Dirac $\delta$
& -
& ~~Full
& \cite{rubin2012trading,galashov2019information,ho2020efficiency,tiomkin2017unified,leibfried2017information}
\\
  ~~KL-Penalized Robust Agent
& $\rightarrow\infty$
& $(-\infty,0)~\pix\pix$
& $\rightarrow$$\scalebox{0.7}{$\pm$}\infty~$
& -
& (any)
& -
& ~~Full
& \cite{petersen2000minimax,charalambous2004relations,osogami2012robustness,grau2016planning}
\\
\midrule
  ~~\textbf{General Formulation}
& \hspace{-2pt}~\smash{$\mathbb{R}\hspace{-2pt}\setminus\hspace{-2pt}\{0\}$}
& \hspace{-2pt}~~~\smash{$\mathbb{R}\hspace{-2pt}\setminus\hspace{-2pt}\{0\}$}
& \hspace{-2pt}~\smash{$\mathbb{R}\hspace{-2pt}\setminus\hspace{-2pt}\{0\}$}
& (any)
& (any)
& (any)
& ~~Full\pix /\pix Partial
& \textbf{Section \ref{sec:example}}
\\
\bottomrule
\end{tabular}
\end{adjustbox}
\end{center}
\vspace{-1.3em}
\end{table*}

\vspace{-0.93em}
\subsection{Bounded Rational Control}\label{sub:bounded}

\dayum{
For perfectly rational agents, the best \textit{decision policy} given any $z$ simply maximizes \smash{$V^{\phi_{\pi\hspace{-1pt},\hspace{-0.5pt}\rho}}(z)$}, thus it selects actions according to \smash{$\text{argmax}_{u}Q^{\phi_{\pi\hspace{-1pt},\hspace{-0.5pt}\rho}}(z,u)$}. And the best \textit{recognition policy} simply corresponds to their unbiased knowledge of the world, thus it sets $\sigma(\cdot|z,u)=\tilde{\sigma},\forall z,u$ (in Equation \ref{eq:recognition}).}

\dayum{
\textbf{Information Constraints}~
But control is resource-intensive. We formalize an agent's boundedness in terms of capacities for processing information. First, \textit{decision complexity} captures the informational effort in determining actions $\pi(\cdot|z)$, relative to some prior \smash{$\tilde{\pi}$} (e.g. baseline clinical guidelines):}
\vspace{-0.80em}
\begin{equation}
\mathbb{I}_{\pi,\rho}[\pi;\tilde{\pi}]
\doteq
\mathbb{E}_{z\sim\mu_{\pi,\rho}}
D_{\text{KL}}
(\pi(\cdot|z)\|\tilde{\pi})
\end{equation}
\vspace{-2.0em}

\dayum{
Second, \textit{specification complexity} captures the average regret of their internal model $\sigma(\cdot|z,u)$ deviating from their prior (i.e. unbiased knowledge \smash{$\tilde{\sigma}$}) about environment dynamics:}
\vspace{-0.80em}
\begin{equation}
\mathbb{I}_{\pi,\rho}[\sigma;\tilde{\sigma}]
\doteq
\mathbb{E}_{\substack{z\sim\mu_{\pi,\rho}\\u\sim\pi(\cdot|z)}}
D_{\text{KL}}
(\sigma(\cdot|z,u)\|\tilde{\sigma})
\end{equation}
\vspace{-1.6em}

\dayum{
Finally, \textit{recognition complexity} captures the statistical surprise in adapting to successive beliefs about the partially-ob- servable states of the world (again, relative to some prior \smash{$\tilde{\varrho}$}):}
\vspace{-0.80em}
\begin{equation}
~~~~~\mathbb{I}_{\pi,\rho}[\varrho;\tilde{\varrho}]
\doteq
\mathbb{E}\hspace{-7pt}_{\substack{z\sim\mu_{\pi,\rho}\\u\sim\pi(\cdot|z)\\\tau,\omega\sim\sigma(\cdot|z,u)}}\hspace{-7pt}
D_{\text{KL}}
(\varrho_{\tau,\omega}(\cdot|z,u)\|\tilde{\varrho})
\end{equation}
\vspace{-1.8em}

where \smash{$\varrho_{\tau,\omega}(\cdot|z,u)\doteq\mathbb{E}_{s\sim p(\cdot|z),s^{\prime}\sim\tau(\cdot|s,u),x^{\prime}\sim\omega(\cdot|u,s^{\prime})}\rho_{\tau,\omega}(\cdot$} \smash{$\hspace{-1pt}z,\hspace{-2pt}u,\hspace{-1pt}x^{\prime})$} gives the internal-state update. We shall see, these measures generalize information-theoretic ideas in control.

\textbf{Backward Process}~
With capacity constraints, the maximi- zation in Equation \ref{eq:rational} now becomes subject to \smash{$\mathbb{I}_{\pi,\rho}[\pi;\tilde{\pi}]\leq$} \smash{$A$}, \smash{$\mathbb{I}_{\pi,\rho}[\sigma;\tilde{\sigma}]\leq B$}, and \smash{$\mathbb{I}_{\pi,\rho}[\varrho;\tilde{\varrho}]\leq C$}. So the Lagrangian (now with the additional multipliers \smash{$\alpha,\beta,\eta$\pix$\in$\pix$\mathbb{R}$}) is given by \smash{$
\mathcal{L}_{\pi,\rho}(\mu,\alpha,\beta,\eta,V)
$\pix$\doteq$\pix$
J_{\pi,\rho}
$\pix$-$\pix$
\langle
V
,
\mu
$$-$$
\gamma\mathbb{M}_{\pi,\rho}\mu
$$-$$
(1$$-$$\gamma)\rho_{0}
\rangle
$\pix$-$} \smash{$
\alpha
$$\cdot$$
(
\mathbb{I}_{\pi,\rho}[\pi;\tilde{\pi}]
$$-$$
A
)
-
\beta
$$\cdot$$
(
\mathbb{I}_{\pi,\rho}[\sigma;\tilde{\sigma}]
$$-$$
B
)
-
\eta
$$\cdot$$
(
\mathbb{I}_{\pi,\rho}[\varrho;\tilde{\varrho}]
$$-$$
C
)
$}.

\vspace{-0.5em}
\begin{reproposition}[restate=backwardx,name=Backward Recursion]\upshape\label{thm:prop}
\dayum{
Define the backwa- rd operator \smash{$\mathbb{B}_{\pi,\rho}:\mathbb{R}^{\mathcal{Z}}$\pix$\rightarrow$\pix$\mathbb{R}^{\mathcal{Z}}$} such that for any given function \smash{$V$\pix$\in$\pix$\mathbb{R}^{\mathcal{Z}}$} and for any given coefficient values \smash{$\alpha,\beta,\eta$\pix$\in$\pix$\mathbb{R}$}:}
\vspace{-0.5em}
\begin{gather}
\begin{aligned}
(\mathbb{B}_{\pi,\rho}V)(z)
\doteq
~\mathbb{E}_{\substack{
s\sim p(\cdot|z)\\
u\sim\pi(\cdot|z)
}}&\big[
-\alpha
\log
\scalebox{1.2}{$
\frac{\pi(u|z)}{\tilde{\pi}(u)}
$}
+
\upsilon(s,u)
+ \\[-2.0ex]
\mathbb{E}_{\tau,\omega\sim\sigma(\cdot|z,u)}&\big[
-\beta
\log
\scalebox{1.2}{$
\frac{\sigma(\tau,\omega|z,u)}{\tilde{\sigma}(\tau,\omega)}
$}
+ \\[-0.8ex]
\mathbb{E}_{\substack{s^{\prime}\sim\tau(\cdot|s,u)\\x^{\prime}\sim\omega(\cdot|u,s^{\prime})\\z^{\prime}\sim\rho_{\tau,\omega}(\cdot|z,u,x^{\prime})}}&\big[
-\eta
\log
\raisebox{-2pt}{\scalebox{1.2}{$
\frac{\varrho_{\tau,\omega}(z^{\prime}|z,u)}{\tilde{\varrho}(z^{\prime})}
$}}
\hspace{-9pt}\raisebox{-13pt}{$+\pix
\gamma
V(z^{\prime})
\big]\big]\big]$}
\end{aligned}\raisetag{3.25\baselineskip}
\end{gather}

\vspace{-0.8em}
\dayum{
Then the (dual) optimal $V$ is the (unique) fixed point of $\mathbb{B}_{\pi,\rho}$; as before, this is the \textit{value function} $V^{\phi_{\pi\hspace{-1pt},\hspace{-0.5pt}\rho}}$---which now in- cludes the complexity terms. Likewise, we can also define the (state-action) \smash{$Q^{\phi_{\pi\hspace{-1pt},\hspace{-0.5pt}\rho}}$\pix$\in$\pix$\mathbb{R}^{\mathcal{Z}\times\mathcal{U}}$} as the \smash{$\nicefrac{1}{3}$}-step-ahead expectation, and the (state-action-model) \raisebox{-0.5pt}{\smash{$K^{\phi_{\pi\hspace{-1pt},\hspace{-0.5pt}\rho}}$\pix$\in$\pix$\mathbb{R}$}}\raisebox{-1.5pt}{\smash{$^{\mathcal{Z}\times\mathcal{U}\times\mathcal{T}\times\mathcal{O}}$}} as the \smash{$\nicefrac{2}{3}$}-steps-ahead expectation (which is new in this setup).}
\end{reproposition}

\dayum{
\textbf{Policies and Values}~
The (dis-)/utility-seeking decision policy (min-)/maximizes \smash{$V\hspace{-1pt}^{\phi_{\pi\hspace{-1pt},\hspace{-0.5pt}\rho}}\hspace{-1pt}(z)$}, and a pessimistic/optimis- tic \textit{recognition policy} min-/maximizes \smash{$Q^{\phi_{\pi\hspace{-1pt},\hspace{-0.5pt}\rho}}(z,u)$} via $\sigma$.\footnote{In general, flipping the direction of optimization for $\pi$ or $\rho$ corre- sponds to the \textit{signs} of $\alpha$ or $\beta$, but does not change Theorems \ref{thm:values}--\ref{thm:policies}.} These optimal policies depend on optimal value functions:}

\vspace{-0.5em}
\begin{retheorem}[restate=values,name=Boundedly Rational Values]\upshape\label{thm:values}
Define the bac- kward operator \smash{$\mathbb{B}^{*}:\mathbb{R}^{\mathcal{Z}}\rightarrow\mathbb{R}^{\mathcal{Z}}$} such that for any \smash{$V\in\mathbb{R}^{\mathcal{Z}}$}:
\vspace{-0.2em}
\begin{gather}
\begin{aligned}
(\mathbb{B}^{*}&V)(z)\hspace{-1.5pt}
\doteq\hspace{-1pt}
\alpha\log\mathbb{E}_{u\sim\tilde{\pi}}\exp(\tfrac{1}{\alpha}Q(z,u)) \\[-0.25ex]
&Q\hspace{0.65pt}(z,u\hspace{0.5pt})\hspace{0.5pt}
\doteq\hspace{1pt}
\beta\log\mathbb{E}_{\tau,\omega\sim\tilde{\sigma}}\exp(\tfrac{1}{\beta}K(z\hspace{-0.5pt},\hspace{-1pt}u\hspace{-0.25pt},\hspace{-1.25pt}\tau\hspace{-2pt},\hspace{-1pt}\omega))
\hspace{-35pt}\raisebox{-15pt}{$+~\mathbb{E}_{s\sim p(\cdot|z)}\upsilon(s,u)$} \\[-3.25ex]
&K(z\hspace{-0.5pt},\hspace{-1pt}u\hspace{-0.25pt},\hspace{-1.25pt}\tau\hspace{-2pt},\hspace{-1pt}\omega)\hspace{-1pt}
\doteq \\[-0.5ex]
&\pix~~~~~~~~~~~~~~\mathbb{E}
\hspace{-9pt}
_{\substack{s\sim p(\cdot|z)\\s^{\prime}\sim\tau(\cdot|s,u)\\x^{\prime}\sim\omega(\cdot|u,s^{\prime})\\z^{\prime}\sim\rho_{\tau,\omega}(\cdot|z,u,x^{\prime})}}
\hspace{-9pt}
\big[
\hspace{-2pt}-\hspace{-2pt}\eta
\log
\raisebox{-2pt}{\scalebox{1.2}{$
\frac{\varrho_{\tau,\omega}(z^{\prime}|z,u)}{\tilde{\varrho}(z^{\prime})}
$}}
+\pix
\gamma
V(z^{\prime})
\big]
\end{aligned}\raisetag{5.8\baselineskip}
\end{gather}

\vspace{-0.9em}
\dayum{Then the \textit{boundedly rational value function} \smash{$V^{*}$} for the (primal) optimal $\pi^{*}\hspace{-2pt},\rho^{*}$ is the (unique) fixed point of \smash{$\mathbb{B}_{\pi,\rho}^{*}$}. (Note that both \smash{$Q^{*}$} and \smash{$K^{*}$} are immediately obtainable from this).}
\end{retheorem}

\begin{retheorem}[restate=policies,name=Boundedly Rational Policies]\upshape\label{thm:policies}
\dayum{The \textit{bounded- ly rational decision policy} (i.e. primal optimal) is given by:}
\vspace{-1.2em}
\begin{equation}
\pi^{*}(u|z)
=
\scalebox{1.2}{$\frac{\tilde{\pi}(u)}{Z_{Q^{*}}(z)}$}
\exp\big(\tfrac{1}{\alpha}Q^{*}(z,u)\big)
\end{equation}

\vspace{-1.2em}
and the \textit{boundedly rational recognition policy} is given by:
\vspace{-0.3em}
\begin{gather}
\begin{aligned}
\rho^{*}(z^{\prime}|z,u,x^{\prime})
\hspace{-1pt}=&\hspace{2pt}
\mathbb{E}_{\tau,\omega\sim\sigma^{*}(\cdot|z,u)}\rho_{\tau,\omega}(z^{\prime}|z,u,x^{\prime})\text{ , where~~~} \\[-0.5ex]
\sigma^{*}(\tau,\omega|z,u)
&\hspace{-1pt}\doteq\hspace{-2pt}
\scalebox{1.2}{$\frac{\tilde{\sigma}(\tau,\omega)}{Z_{K^{*}}(z,u)}$}
\exp\big(\tfrac{1}{\beta}K^{*}(z,u,\tau,\omega)\big)
\end{aligned}\raisetag{1.05\baselineskip}
\end{gather}

\vspace{-0.9em}
\dayum{where \smash{$Z_{Q^{*}}(z)=\mathbb{E}_{u\sim\tilde{\pi}}\exp(\tfrac{1}{\alpha}Q^{*}(z,u))$} and \smash{$Z_{K^{*}}(z,u)=$} \smash{$\mathbb{E}_{\tau,\omega\sim\tilde{\sigma}}\exp(\tfrac{1}{\beta}K^{*}(z,u,\tau,\omega))$} give the partition functions.}
\end{retheorem}

\textbf{Interpretation of Parameters}~
This articulation of bound- ed rationality reflects the fact that imperfect behavior results from two sources of ``boundedness'': Firstly, that (1) given a mental model $\rho$ for comprehending the world, an agent's information-processing capacities distort their decision-ma- king $\pi$ (cf. suboptimal actions); and secondly, that (2) the agent's mental model $\rho$ itself is an imperfect characteriza- tion of the world---because prior knowledge $\tilde{\sigma}$ is uncertain, and internal states can be biased by $\sigma$ (cf. subjective beliefs).

\dayum{
Concretely, the parameters in Theorems \ref{thm:values}--\ref{thm:policies} admit intuitive interpretations. First, \smash{$\alpha^{-1}$} captures \textit{flexibility} of decision-making, from a completely inflexible agent (\smash{$\alpha^{-1}$$\rightarrow$\pix$0$}) to an infinitely flexible, utility-seeking (\smash{$\alpha^{-1}$$\rightarrow$$\infty$}) or disutility-seeking (\smash{$\alpha^{-1}$$\rightarrow$$-\infty$}) one. Second, \smash{$\beta^{-1}$} captures \textit{optimism} in internal models, from a completely neutral agent (\smash{$\beta^{-1}$$\rightarrow$} $0$) to an infinitely optimistic (\smash{$\beta^{-1}$$\rightarrow$$\infty$}) or pessimistic (\smash{$\beta^{-1}$} \smash{$\rightarrow$$-\infty$}) one. Lastly, $\eta^{-1}$ captures \textit{adaptivity} of beliefs, from a perfectly adaptive agent (\smash{$\eta^{-1}$$\rightarrow\pm\infty$}) to one with infinite intolerance (\smash{$\eta^{-1}$$\rightarrow0^{+}$}) or affinity (\smash{$\eta^{-1}$$\rightarrow0^{-}$}) for surprise.
Table \ref{tab:agent} underscores the generality of this parameterization.}

\vspace{-0.3em}
\subsection{Inverse Bounded Rational Control}\label{sub:estimate}

\dayum{
We hark back to our framework of Section \ref{sec:unified}: In bounded rational control (``BRC''), the \textit{planning parameter} \smash{$\theta^{\scalebox{0.6}{\pix\text{BRC}}}$} represents \smash{$\{\upsilon,\gamma,\alpha,\beta,\eta,\tilde{\pi},\tilde{\sigma},\tilde{\varrho}\}$}, and the space \smash{$\Theta^{\scalebox{0.6}{\pix\text{BRC}}}$} is again decomposable as \smash{$\Theta_{\text{norm}}^{\scalebox{0.6}{\pix\text{BRC}}}$\pix$\times$\pix$\Theta_{\text{desc}}^{\scalebox{0.6}{\pix\text{BRC}}}$}. The \textit{forward problem} is encapsulated by Theorems \ref{thm:values}--\ref{thm:policies} (which also yield a straight- forward algorithm, i.e. iterate \ref{thm:values} until convergence, then plug into \ref{thm:policies}). Therefore the \textit{forward planner} is given as follows:}

\vspace{-1.0em}
\begin{equation}
F_{\theta_{\text{norm}}^{\scalebox{0.425}{\pix\text{BRC}}}}(\theta_{\text{desc}}^{\scalebox{0.6}{\pix\text{BRC}}})\doteq\phi_{\pi^{*}\hspace{-2pt},\rho^{*}}
\hspace{1pt}:\hspace{2pt}
\pi^{*}\hspace{-2pt},\rho^{*}
\hspace{1pt}\leftarrow
\text{Theorems \ref{thm:values}--\ref{thm:policies}}
\end{equation}

\vspace{-1.0em}
\dayum{
In the opposite direction, the problem is of \textit{inverse bounded rational control}. Consider a minimal setting where we are given access to logged data \smash{$\mathcal{D}\doteq\{h_{n}\sim\phi_{\text{demo}}\}_{n=1}^{N}$} with no additional annotations. While several options from Table \ref{tab:inverse} are available, for simplicity we select soft policy matching for illustration. Thus the \textit{inverse planner} is given as follows:}
\vspace{-1.0em}
\begin{equation}\label{eq:24}
G_{\theta_{\text{norm}}^{\scalebox{0.425}{\pix\text{BRC}}}}(\phi\hspace{0.5pt})\doteq\text{argmin}_{\theta_{\text{desc}}^{\scalebox{0.425}{\pix\text{BRC}}}}\mathbb{E}_{h\sim\phi}\log\mathbb{P}_{\phi_{\text{imit}}}(u_{0:T}\|x_{0:T})\hspace{-5pt}
\end{equation}

\vspace{-0.9em}
\dayum{
where \smash{$\mathbb{P}_{\phi_{\pi,\rho}}(u_{0:T}\|x_{0:T})$} is the causally-conditioned probability \cite{ziebart2010thesis,kramer1998thesis,massey1990causality,marko1973bidirectional}
\smash{$\prod_{t=0}^{T}\mathbb{P}_{\phi_{\pi,\rho}}(u_{t}|x_{1:t},u_{1:t-1})$}---with the conditioning as induced by \smash{$\pi,\rho$}. In the most general case where \smash{$\rho_{\tau,\omega}$} may be stochastic, \smash{$G_{\theta_{\text{norm}}^{\scalebox{0.425}{\pix\text{BRC}}}}$} would require an EM approach; however, since we selected \smash{$\rho_{\tau,\omega}$} to be the (deterministic) Bayes update for interpretability, the likelihood is:}
\vspace{-1.3em}
\begin{equation}\label{eq:likelihood}
\begin{split}
\log\mathbb{P}_{\phi_{\pi,\rho}}(u_{0:T}\|x_{0:T})
&\propto
\textstyle\sum_{t=0}^{T}\log\pi(u_{t}|z_{t})
\end{split}
\end{equation}

\vspace{-1.3em}
where the \smash{$z_{t}$} terms are computed recursively by $\rho$ (see Appendix \ref{app:c}). Finally, here the \textit{inverse decision model} of any \smash{$\phi_{\text{demo}}$} is given by its projection \smash{$\phi_{\text{imit}}^{*}=F_{\theta_{\text{norm}}^{\scalebox{0.425}{\pix\text{BRC}}}}\circ G_{\theta_{\text{norm}}^{\scalebox{0.425}{\pix\text{BRC}}}}(\phi_{\text{demo}})$} onto the space \smash{$\Phi_{\theta_{\text{norm}}^{\scalebox{0.425}{\pix\text{BRC}}}}$} of behaviors thereby \textit{interpretably} para- meterized---i.e. by the structure we designed for \smash{$\Theta^{\scalebox{0.6}{\pix\text{BRC}}}$}, and by the normative standards \smash{$\theta_{\text{norm}}^{\scalebox{0.6}{\pix\text{BRC}}}$} we may choose to specify.


\section{Illustrative Use Case}\label{sec:simulate}

\begin{figure*}
\vspace{-1.2em}
\centering
\captionsetup[subfigure]{justification=centering,singlelinecheck=false,labelformat=empty,position=top}
\subfloat[
{\scriptsize ~~~~~~~~\textbf{Very Flexible Agent}: $\alpha$\pix$=$\pix$10^{-3}$}
]{
\includegraphics[width=0.32\linewidth, trim=1.3em 1.5em 1.1em 1.5em]{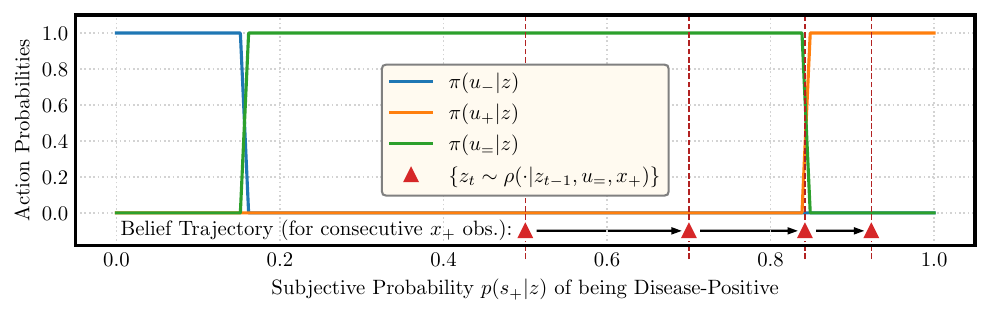}
}\hfill
\subfloat[
{\scriptsize ~~~~~\textbf{Optimistic Agent}: $\beta$\pix$=$\pix$1.25$}
]{
\includegraphics[width=0.32\linewidth, trim=1.3em 1.5em 1.1em 1.5em]{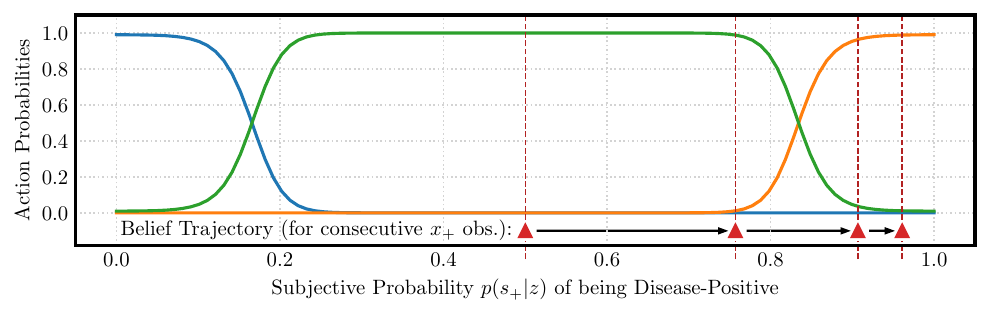}
}\hfill
\subfloat[
{\scriptsize ~~~~~\textbf{Adaptive Agent}: $\eta$\pix$=$\pix$10^{-3}$}
]{
\includegraphics[width=0.32\linewidth, trim=1.3em 1.5em 1.1em 1.5em]{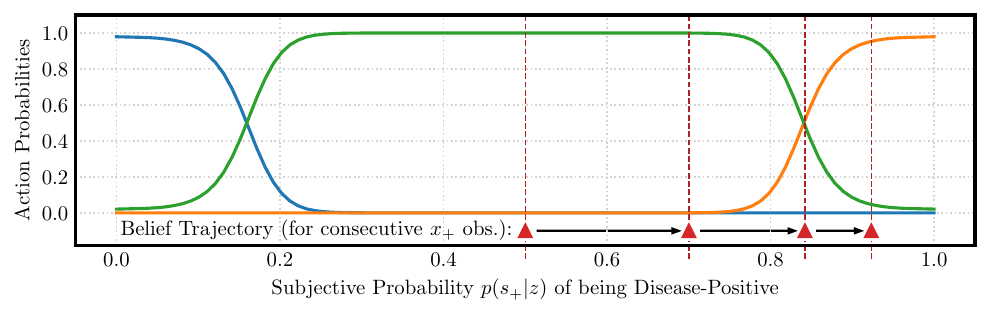}
}
\vspace{-0.6em}
\subfloat[
{\scriptsize ~~~~~~~\textbf{Inflexible Agent}: $\alpha$\pix$=$\pix$10$}
]{
\includegraphics[width=0.32\linewidth, trim=1.3em 1.5em 1.1em 1.5em]{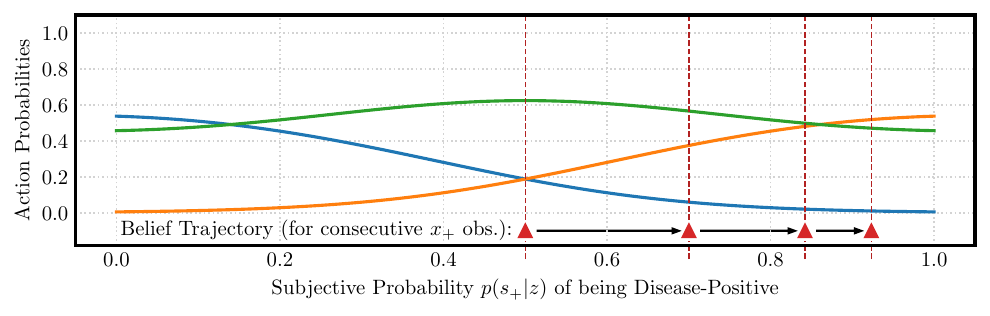}
}\hfill
\subfloat[
{\scriptsize ~~~~~\textbf{Pessimistic Agent}: $\beta$\pix$=$\pix$-0.75$}
]{
\includegraphics[width=0.32\linewidth, trim=1.3em 1.5em 1.1em 1.5em]{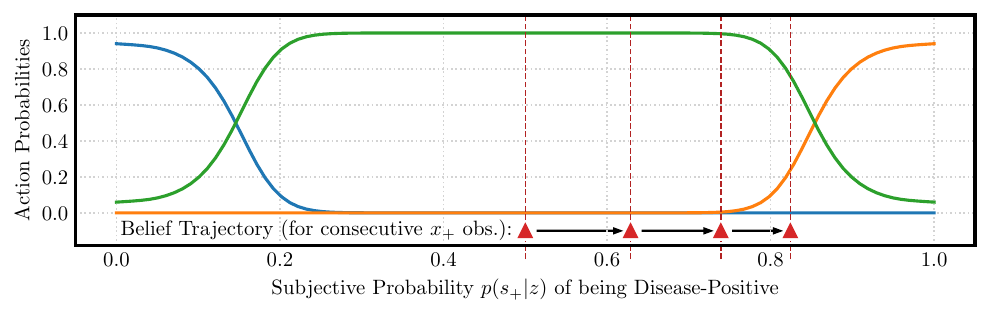}
}\hfill
\subfloat[
{\scriptsize ~~~~~\textbf{Non-adaptive Agent}: $\eta$\pix$=$\pix$75$}
]{
\includegraphics[width=0.32\linewidth, trim=1.3em 1.5em 1.1em 1.5em]{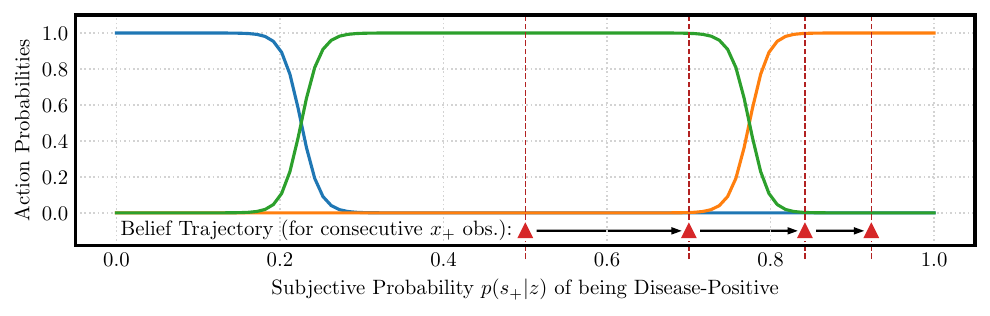}
}
\vspace{-0.4em}
\subfloat[
\scalebox{0.9}{(a) Effect of Flexibility, for a neutral} \scalebox{0.9}{\mbox{~~~}({\scriptsize $\beta$\pix$=$\pix$10^{3}$}), adaptive ({\scriptsize $\eta$\pix$=$\pix$10^{-3}$}) agent}
]{
\rule{0.32\linewidth}{0pt}
}\hfill
\subfloat[
\scalebox{0.9}{(b) Effect of Optimism, for a flexible} \scalebox{0.9}{\mbox{~~~}({\scriptsize $\alpha$\pix$=$\textbf{}$0.5$}), adaptive ({\scriptsize $\eta$\pix$=$\pix$10^{-3}$}) agent}
]{
\rule{0.32\linewidth}{0pt}
}\hfill
\subfloat[
\scalebox{0.9}{(c) Effect of Adaptivity, for a flexible} \scalebox{0.9}{\mbox{~~~}({\scriptsize $\alpha$\pix$=$\pix$0.5$}), neutral ({\scriptsize $\beta$\pix$=$\pix$10^{3}$}) agent}
]{
\rule{0.32\linewidth}{0pt}
}
\vspace{-1.25em}
\caption{\textit{Bounded Rational Control}. Decision agents in DIAG: In each panel, the boundedly rational decision policy $\pi$ is shown in terms of action probabilities ($y$-axis) for different subjective beliefs ($x$-axis). To visualize the boundedly rational recognition policy $\rho$, each panel shows an example trajectory of beliefs (\smash{$z_{0},z_{1},z_{2},z_{3}$}) for the case where three consecutive positive outcomes are observed (\mytriangle{crimson} markers).}
\label{fig:forward}
\label{fig:graphs}
\vspace{-0.8em}
\end{figure*}

So far, we have argued for a systematic, unifying perspective on inverse decision modeling (``IDM'') for behavior representation learning, and presented inverse bounded rational control (``IBRC'') as a concrete example of the formalism. Three aspects of this approach deserve empirical illustration:

\vspace{-0.8em}
\begin{itemize}[leftmargin=*]
\itemsep-1pt
\item \dayum{\muline{\textit{Interpretability}}: IBRC gives a \textit{transparent} parameterization of behavior that can be successfully learned from data.}
\item \dayum{\muline{\textit{Expressivity}}: IBRC more finely \textit{differentiates} between imperfect behaviors, while standard reward learning cannot.}
\item \dayum{\muline{\textit{Applicability}}: IDM can be used in real-world settings, as an investigative device for understanding \textit{human} decisions.}
\end{itemize}
\vspace{-0.7em}

\textbf{Normative-Descriptive Questions}~
Consider \textit{medical diagnosis}, where there is often remarkable regional, institutional, and subgroup-level variability in practice \cite{mckinlay2007sources,bock2016preoperative,osullivan2018variation}, rendering detection and quantification of biases crucial \cite{song2010regional,martin2017routine,allen2017unnecessary}.
Now in modeling an agent's behavior, reward learning asks: \textbf{(1)} ``What does this (perfectly rational) agent appear to be optimizing?'' And the answer takes the form of a function $\upsilon$. However, while $\upsilon$ alone is often sufficient as an \textit{intermediary} for imitation\pix/\pix apprenticeship, it is seldom what we actually want \textit{as an end by itself}---for \mbox{introspective understanding}. Importantly, we often \textit{can} articulate some version of what our preferences $\upsilon$ are. In medical diagnosis, for instance, from the view of an investigator, the average relative healthcare cost/benefit of in-/correct diagnoses is certainly spe- cifiable as a normative standard. So instead, we wish to ask: \textbf{(2)} ``Given that this (boundedly rational) agent should optimize this $\upsilon$, \mbox{\textit{how suboptimally do they appear to behave?}}'' Clearly, such \textit{normative-descriptive} questions are only poss- ible with the generalized perspective of IDM (and IBRC): Here, $\upsilon$ is specified (in $\theta_{\text{norm}}$), whereas one or more behavio- ral parameters $\alpha,\beta,\eta$ are what we wish to recover (in $\theta_{\text{desc}}$).

\vspace{0.2em}
\dayum{
\textbf{Decision Environments}~
For our simulated setting (\textbf{DIAG}), we consider a POMDP where patients are diseased (\smash{$s_{\raisebox{1pt}{\scalebox{0.6}{+}}}$}) or healthy (\smash{$s_{\raisebox{1pt}{\scalebox{0.6}{--}}}$}), and vital-signs measurements taken at each step are noisily indicative of being disease-positive (\smash{$x_{\raisebox{1pt}{\scalebox{0.6}{+}}}$}) or negative (\smash{$x_{\raisebox{1pt}{\scalebox{0.6}{--}}}$}). Actions consist of the decision to continue monitoring the patient (\smash{$u_{\raisebox{1pt}{\scalebox{0.6}{=}}}$})---which yields evidence, but is also costly; or stopping and declaring a final diagnosis---and if so, a diseased (\smash{$u_{\raisebox{1pt}{\scalebox{0.6}{+}}}$}) or healthy (\smash{$u_{\raisebox{1pt}{\scalebox{0.6}{--}}}$}) call. Importantly, note that since we simulate $\tau,\omega$\pix$\sim$\pix$\sigma(\cdot|z,u)$, DIAG is a strict generalization of the diagnostic environment from \cite{huyuk2021explaining} with a point-valued, subjective $\tau,\omega$\pix$\neq$\pix$\tau_{\text{env}},\omega_{\text{env}}$, and of the classic Tiger Problem in POMDP literature where $\tau,\omega$\pix$=$\pix$\tau_{\text{env}},\omega_{\text{env}}$ \cite{kaelbling1998planning}.}

\dayum{
For our real-world setting, we consider 6-monthly clinical data for 1,737 patients in the Alzheimer's Disease Neuroim- aging Initiative \cite{marinescu2018tadpole} study (\textbf{ADNI}). The state space consists of normal function (\smash{$s_{\text{norm}}$}), mild cognitive impairment (\smash{$s_{\text{mild}}$}), and dementia (\smash{$s_{\text{dem}}$}). For the action space, we consider ordering\pix/\pix not ordering an MRI---which yields evidence, but is costly. Results are classified per hippocampal volume: average (\smash{$x_{\text{avg}}^{\scalebox{0.6}{\pix\text{MRI}}}$}), high (\smash{$x_{\text{high}}^{\scalebox{0.6}{\pix\text{MRI}}}$}), low (\smash{$x_{\text{low}}^{\scalebox{0.6}{\pix\text{MRI}}}$}), not ordered (\smash{$x_{\text{none}}^{\scalebox{0.6}{\pix\text{MRI}}}$}); separately, the cognitive dementia rating test result---which is always measured---is classified as normal (\smash{$x_{\text{norm}}^{\scalebox{0.6}{\pix\text{CDR}}}$}), questionable impairment (\smash{$x_{\text{ques}}^{\scalebox{0.6}{\pix\text{CDR}}}$}), and suspected dementia (\smash{$x_{\text{susp}}^{\scalebox{0.6}{\pix\text{CDR}}}$}). So the observation space consists of such 12 combinations.}

\dayum{
In DIAG, our normative specification (for $\upsilon$) is that diagnos- tic tests cost $-$$1$, correct diagnoses award $10$, incorrect $-$$36$, and $\gamma$\pix$=$\pix$0.95$. Accuracies are $70\%$ in both directions (\smash{$\omega_{\text{env}}$}), and patients arrive in equal proportions (\smash{$\tau_{\text{env}}$}). But this is un- known to the agent: We simulate $\tilde{\sigma}$ by discretizing the space of models such that probabilities vary in $\pm10\%$ increments from the (highest-likelihood) truth. In ADNI, the configuration is similar---except each MRI costs $-1$, while $2.5$ is awarded once beliefs reach $>$$90\%$ certainty in any direction; also, \smash{$\tilde{\sigma}$} is centered at the IOHMM learned from the data. For simplicity, for \smash{$\tilde{\pi},\tilde{\varrho}$} we use uniform priors in both settings.}

Computationally, inference is performed via MCMC in log-parameter space (i.e. $\log\alpha,\log\beta,\log\eta$) using standard methods, similar to e.g. Bayesian IRL \cite{ramachandran2007bayesian,dimitrakakis2011bayesian,makino2012apprenticeship}. In DIAG, we use 1,000 generated trajectories as basis for learning. Appendix \ref{app:b} provides further details on experimental setup.

\begin{figure*}
\centering
\captionsetup[subfloat]{labelformat=empty}
\subfloat[
\scalebox{0.9}{(a) Learned $\alpha$ for Various Flexibility Levels}
]{\includegraphics[width=.32\linewidth, trim=1.3em 0.95em 1.2em 4.4em]{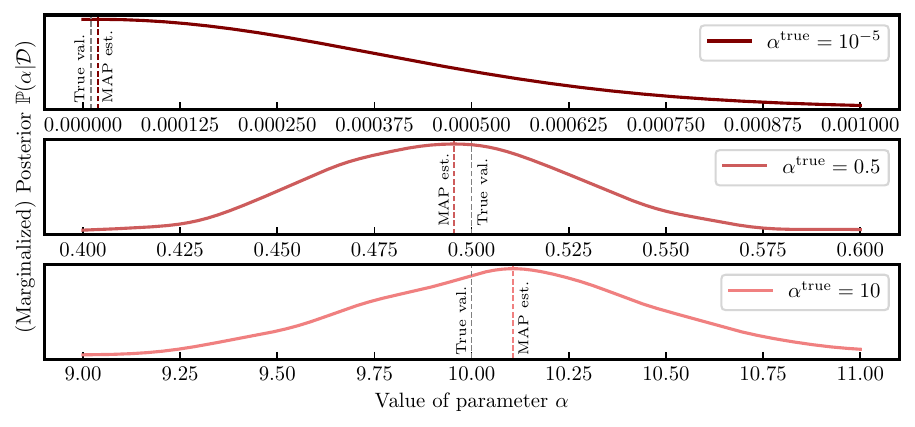}}~
~\subfloat[
\scalebox{0.9}{(b) Learned $\beta,\eta$ for Non-adaptive Behavior}
]{\includegraphics[width=.327\linewidth, trim=0em 4.85em 0.85em 6.3em, clip]{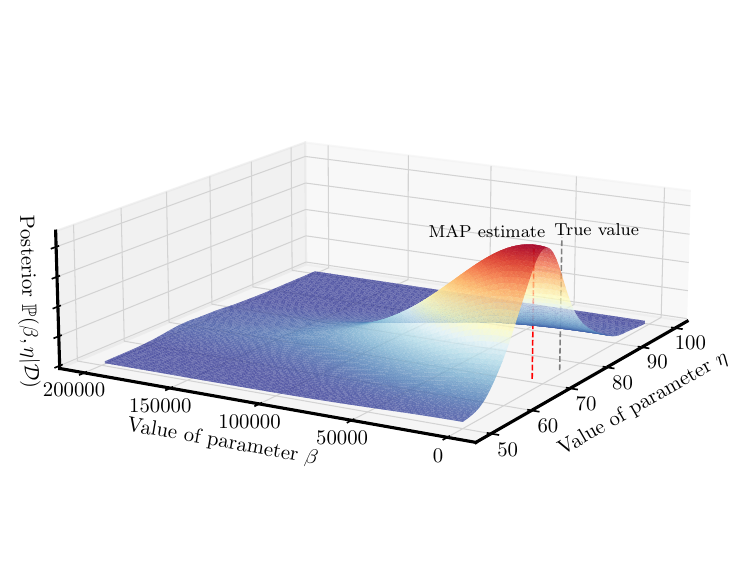}}
~~\subfloat[
\scalebox{0.9}{(c) Learned $\beta,\eta$ for Optimistic Behavior}
]{\includegraphics[width=.313\linewidth, trim=0em 4.85em 0.85em 6.3em, clip]{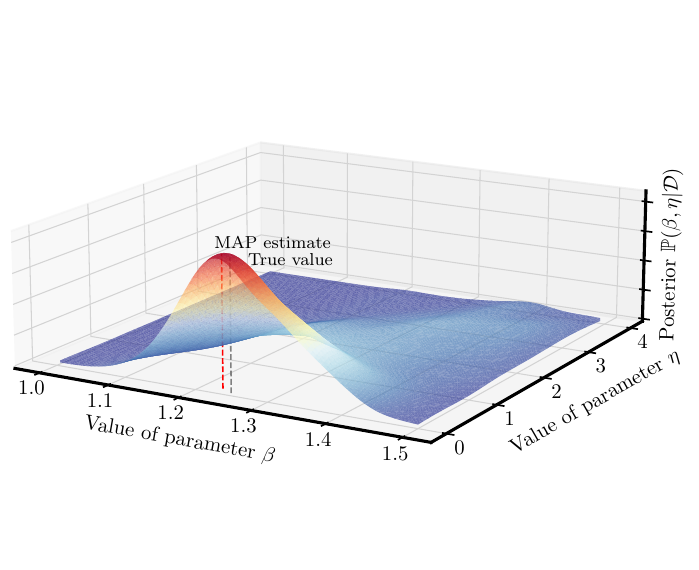}}
\vspace{-0.7em}
\caption{\textit{Inverse Bounded Rational Control}. (a) Posteriors of $\alpha$ learned from extremely flexible (\smash{$\alpha^{\text{true}}$$=$\pix$10^{-5}$}), flexible (\smash{$\alpha^{\text{true}}$$=$\pix$0.5$}), and inflexible (\smash{$\alpha^{\text{true}}$$=$\pix$10$}) behaviors (with $\beta,\eta$ fixed as neutral and adaptive; similar plots can be obtained for those as well). (b) Joint posterior of $\beta,\eta$ for neutral but non-adaptive behavior (\smash{$\beta^{\text{true}}$$=$\pix$10^{3},\eta^{\text{true}}$$=$\pix$75$}), and for (c) optimistic but adaptive behavior (\smash{$\beta^{\text{true}}$$=$\pix$1.25,\eta^{\text{true}}$$=$\pix$10^{-3}$}).}
\vspace{-0.8em}
\label{fig:inverse}
\end{figure*}

\vspace{0.2em}
\dayum{
\textbf{5.1. Interpretability}~
Figure \ref{fig:forward} verifies (for DIAG) that different BRC behaviors accord with our intuitions. First, \textit{cet- eris paribus}, the flexibility ($\alpha$) dimension manifests in how deterministically\pix/\pix stochastically optimal actions are selected (cf. willingness to deviate from action prior \smash{$\tilde{\pi}$}): This is the notion of \textit{behavioral consistency} \cite{karni1991behavioral} in psychology. Sec- ond, the optimism ($\beta$) dimension manifests in the illusion that diagnostic tests are more\pix/\pix less informative for subjective beliefs (cf. willingness to deviate from knowledge prior \smash{$\tilde{\sigma}$}): This is the phenomenon of \textit{over-/underreaction} \cite{daniel1998investor}. Third, the adaptivity ($\eta$) dimension manifests in how much\pix/\pix little evidence is required for declaring a final diagnosis: This corresponds to \textit{base-rate neglect}\pix/\pix\textit{confirmation bias} \cite{tversky1981evidential}.
Hence by \textit{learning} the parameters $\alpha,\beta,\eta$ from data, IBRC provides an \mbox{eminently interpretable example of} behavior re- presentation learning---one that exercises the IDM perspective (much more than just reward learning). Taking a Baye- sian approach to the likelihood (Equation \ref{eq:likelihood}), Figure \ref{fig:inverse}(a) verifies that---as expected---IBRC is capable of recovering different parameter values from their generated behaviors.}

\dayum{
\textbf{5.2. Expressivity}~
Consider (i.) an agent who is biased tow- ards \textit{optimism}, but otherwise flexible and adaptive (Figure \ref{fig:forward}(b), top), and (ii.) an agent who is \textit{non-adaptive}, but otherwise flexible and neutral (\ref{fig:forward}(c), bottom). Now, to an external observer, both types of boundedness lead to similar styles of behavior: They both tend to declare final diagnoses \textit{earlier} than a neutral and adaptive agent would (\ref{fig:forward}(c), top)---that is, \smash{$\pi(u_{\raisebox{1pt}{\scalebox{0.6}{+}}}|z)$\pix$\approx$\pix$1$} after only 2 (not 3) positive tests. Of course, the former does so due to overreaction (evaluating the evidence incorrectly), whereas the latter does so due  to a lower threshold for stopping (despite \mbox{correctly evaluating the evidence)}.
\mbox{As shown by Figures \ref{fig:inverse}(b)--(c)}, IBRC does differentiate between these two different types of biased behaviors: This is revealing, if not necessarily surprising. Crucially, however, this distinction is \textit{not possible} with conventional IRL. All else equal, let us perform Bayesian IRL on the very same behaviors---that is, to learn an effectively skewed $\upsilon$ (while implicitly setting $\alpha,\beta,\eta$ to their perfectly rational limits). As it turns out, the recovered $\upsilon$ for (i.) gives a cost-benefit ratio (of incorrect/correct diagnoses) of $-2.70$\pix$\pm$\pix$0.31$, and the recovered $\upsilon$ for (ii.) gives a ratio of $-2.60$\pix$\pm$\pix$0.29$. Both agents appear to penalize incorrect diagnoses much less than the normative specification of $-3.60$, which is consistent with them tending to commit to final diagnoses earlier than they should. However, this fails to \textit{differentiate} between the two distinct underlying reasons for behaving in this manner.}

\vspace{0.2em}
\textbf{5.3. Applicability}~
Lastly, we highlight the potential utility of IDM in real-world settings as an \textit{investigative device} for auditing and understanding human decision-making. Consider diagnostic patterns for identifying dementia in ADNI, for patients from different risk groups. For instance, we discover that while $\beta$\pix$=$\pix$3.86$ for all patients, clinicians appear to be \textit{significantly less optimistic} when diagnosing patients with the ApoE4 genetic risk factor ($\beta$\pix$=$\pix$601.74$), for female patients ($\beta$\pix$=$\pix$920.70$), and even more so for patients aged $>$\pix$75$ ($\beta$\pix$=$\pix$2,265.30$). Note that such attitudes toward risk factors align with prevailing medical knowledge \cite{allan2011influence,artero2008risk,hua2010sex}.
Moreover, in addition to obtaining such agent-level interpretations of biases (i.e. using the learned parameters), we can also obtain trajectory-level interpretations of decisions (i.e. using the evolution of beliefs). Appendix \ref{app:d} gives examples of ADNI patients using diagrams of trajectories in the belief simplex, to contextualize actions the taken by clinical decision-makers and identify potentially belated diagnoses.

\vspace{-0.25em}
\section{Conclusion}

\dayum{
In this paper, we motivated the importance of descriptive models of behavior as the bridge between normative and prescriptive decision analysis, and formalized a unifying perspective on inverse decision modeling for behavior representation learning.
%
For future work, an important question lies in exploring differently structured parameterizations $\Theta$ that are \textit{interpretable} for different purposes. After all,
IBRC is only one prototype that exercises the IDM formalism more fully. Another question is to what extent different forms of the inverse problem is \textit{identifiable} to begin with. For instance, it is well-known that even with perfect knowledge of a demonstrator's policy, in single environments we can only infer utility functions up to reward shaping. Thus balancing complexity, interpretability, and identifiability of decision models would be a challenging direction of work.}

\clearpage

\section*{Acknowledgments }

We would like to thank the reviewers for their generous feedback. This work was supported by Alzheimer’s Research UK, The Alan Turing Institute under the EPSRC grant EP/N510129/1, the US Office of Naval Research, as well as the National Science Foundation under grant numbers 1407712, 1462245, 1524417, 1533983, and 1722516.

\bibliographystyle{unsrt}

\balance\bibliography{
bib/0-imitate,
bib/1-reinforce,
bib/2-entropy,
bib/3-information,
bib/4-constraints,
bib/5-risk,
bib/6-intrinsic,
bib/7-bounded,
bib/8-identification,
bib/9-interpret,
bib/a-miscellaneous
}

\clearpage\nobalance
\appendix

\section*{Appendices}

\dayum{Appendix \ref{app:a}  gives a longer discussion of merits and caveats;
Appendix \ref{app:b} gives further experiment details; Appendix \ref{app:c} gives derivations of propositions; Appendix \ref{app:d} shows illustrative trajectories; Appendix \ref{app:0} gives a summary of notation.}

\vspace{-0.315em}
\section{Discussion}\label{app:a}

\begin{figure}[t]
\centering
\vspace{-0.7em}
\caption{\textit{Normative, Prescriptive, and Descriptive Modeling}. Re- call the ``lifecycle'' of decision analysis (Section \ref{sec:intro}). As a paradigm of optimal behavior, \textit{normative standards} serve as a theoretical benchmark. To guide imperfect agents toward this ideal, \textit{prescriptive advice} serves to engineer behavior from humans in the loop. Importantly, however, this first requires an understanding of the imperfections---relative to the normative ideal---that require correcting. This is the goal of \textit{descriptive modeling}---that is, to obtain an empirical account of existing behavior from observed data.
Pre-cisely, inverse decision modeling (middle) leverages a normative standard (left) to obtain an interpretable account of demonstrated behavior, thereby enabling the introspection of existing practices, which\hspace{0.25pt} may\hspace{0.25pt} inform\hspace{0.25pt} construction\hspace{0.25pt} of\hspace{0.25pt} prescriptive\hspace{0.25pt} guidelines\hspace{0.25pt} (right).}
\label{fig:pull}
\vspace{0.75em}
\includegraphics[width=\linewidth]{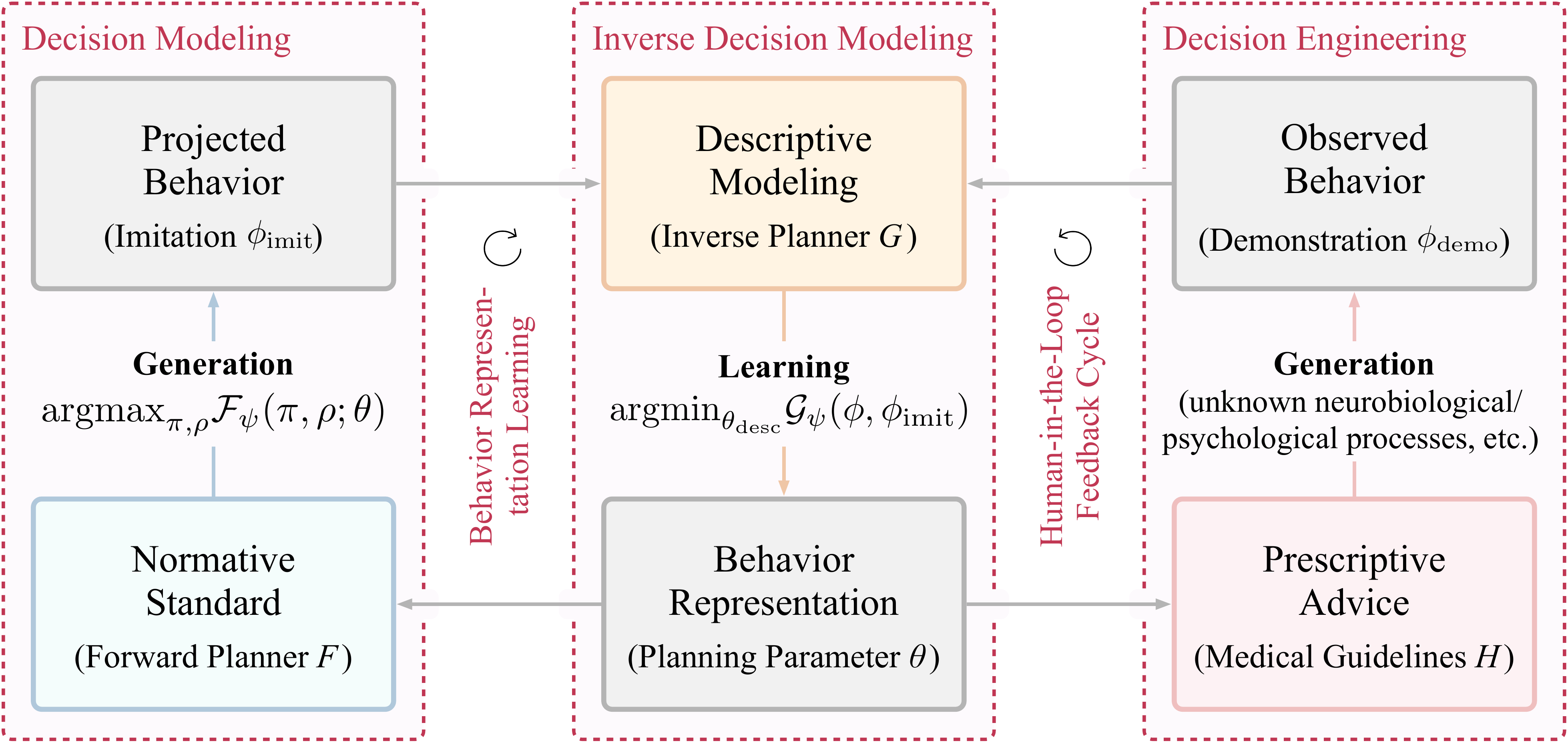}
\vspace{-1.75em}
\end{figure}

\dayum{
In this paper, we motivated the importance of \textit{descriptive} models of behavior as the bridge between normative and prescriptive decision analysis \cite{keller1989role,neumann1947theory,mellers1998judgment} (Figure \ref{fig:pull}). On account of this, we formalized a unifying perspective on inverse decis- ion modeling for behavior representation learning. Precisely, the \textit{inverse decision model} of any observed behavior \smash{$\phi_{\text{demo}}$} is given by its projection \smash{$\phi_{\text{imit}}^{*}=F_{\theta_{\text{norm}}}\circ G_{\theta_{\text{norm}}}(\phi_{\text{demo}})$} onto the space \smash{$\Phi_{\theta_{\text{norm}}}$} of behaviors parameterizable by the structure designed for \smash{$\Theta$} and normative standards \smash{$\theta_{\text{norm}}$} specified.
This formulation is general. For instance, it is agnostic as to the nature of agent and environment state spaces (which--- among other properties---are encoded in $\psi$); it is also agnostic as to whether the underlying forward problem is model-free or model-based (which---among other properties---is encoded in $\theta$). Per the priorities of the investigator (cf. imitation, apprenticeship, understanding, and other objectives), different choices can and should be made to balance the expressivity, interpretability, and tractability of learned models.}

\dayum{
\textbf{Partial Observability}~
At first glance, our choice to accommodate partial observability may have appeared inconsequential. However, its significance becomes immediately apparent once we view an agent's behavior as induced by both a \textit{decision} policy $\pi$ as well as a \textit{recognition} policy $\rho$, and---importantly---that not only may an agent's mapping from internal states into actions be suboptimal (viz. the former), but that their mapping from observations into beliefs may also be subjective (viz. the latter). Therefore in addition to the oft-studied, purely utility-centric nature of (perfectly rational) behavior, this generalized formalism immediately invites consideration of (boundedly rational) behaviors---that is, agents acting under knowledge uncertainty, biased by optimism/robustness, with policies distorted by the complexit- ies of information processing required for decision-making.}

\textbf{Bounded Rationality}~
While the IDM formalism subsumes most standard approaches to imitation learning, apprenticeship learning, and reward learning (cf. Table \ref{tab:project} and Table \ref{tab:inverse}), we emphasize that---with very few exceptions \cite{wu2018inverse,daptardar2019inverse,kwon2020inverse}---the vast majority of original studies in these areas are limited to cases where $\theta_{\text{desc}}$\pix$=$\pix$\upsilon$ alone, or assume fully-observable environments (whence $\mathcal{S}$\pix$=$\pix$\mathcal{X}$\pix$=$\pix$\mathcal{Z}$, and $\rho$ simply being the identity function). Therefore our concrete example of \textit{inverse bounded rational control} was presented as a prototypical instantiation of IDM that much more fully exercises the flexibility afforded by this generalized perspective. Importantly, while our notion of bounded rationality has (implicitly) been present to varying degrees in (forward) control and reinforcement learning (cf. Table \ref{tab:forward} and Table \ref{tab:agent}), ``boundedness'' has largely been limited to mean ``noisy actions''. To be precise, we may differentiate between three ``levels'' of boundedness:

\vspace{-0.8em}
\begin{itemize}[leftmargin=*]
\itemsep-1pt
\item \dayum{\textit{Imperfect Response}: This is the shallowest form of boundedness, and includes Boltzmann-exploratory \cite{heess2013actor,osband2016generalization,cesa2017boltzmann} and (locally) entropy-regularized \cite{odonoghue2017combining} behaviors: It considers first that agents are perfect in their ability to compute the optimal values/policies; however, their actions are ultimately executed with an artificial layer of stochasticity.}
\item \textit{Capacity Constraints}: Given an agent's model (e.g. $\tau,Q$- network, etc.), the information processing needed in computing actions on the go is costly. We may view soft-opt- imal \cite{haarnoja2017reinforcement,haarnoja2018soft,eysenbach2019if,shi2019soft} and KL-regularized \cite{rubin2012trading,galashov2019information,ho2020efficiency,tiomkin2017unified,leibfried2017information} planning and learning as examples. However, these do not model subjectivity of beliefs, adaptivity, or optimism/robustness.
\item \textit{Model Imperfection}: The agent's mental model itself is systematically flawed, due to uncertainty in knowledge, and to biases from optimism or pessimism. We may view certain robust MDPs (with penalties for deviating from priors) \cite{petersen2000minimax,charalambous2004relations,grau2016planning,osogami2012robustness} as examples. However, these still do not account for partial observability (and biased recognition).
\end{itemize}
\vspace{-0.7em}

\dayum{
Now in the inverse direction, imitation/apprenticeship learning has typically viewed reward learning as but an intermediary, so classical methods have worked with perfectly rational planners \cite{ng2000algorithms,abbeel2004apprenticeship,syed2008game,syed2008apprenticeship,klein2011batch,mori2011model,lee2019truly,choi2009inverse,choi2011inverse,chinaei2012inverse,bica2021learning}. Approaches that leverage probabilistic methods have usually simply used Boltzmann-exploratory policies on top of optimal action-value functions (viz. imperfect response) \cite{ramachandran2007bayesian,choi2011map,dimitrakakis2011bayesian,rothkopf2011preference,balakrishnan2020efficient,neu2007apprenticeship,babes2011apprenticeship,tossou2013probabilistic,jain2019model,makino2012apprenticeship,jarrett2020inverse}, or worked within maximum entropy planning/learning frameworks (viz. capacity constraints) \cite{ziebart2008maximum,boularias2011relative,kalakrishnan2013learning,wulfmeier2015maximum,finn2016connection,fu2018learning,qureshi2019adversarial,barde2020adversarial,ziebart2010modeling,zhou2017infinite,lee2018maximum,mai2019generalized}. Crucially, however, the corresponding parameters (i.e. inverse temperatures) have largely been treated as \textit{pre-specified} parameters for learning $\upsilon$ alone---not \textit{learnable} parameters of interest by themselves.
In contrast, what IDM allows (and what IBRC illustrates) is the ``fullest'' extent of boundedness---that is, where stochas- tic actions and subjective beliefs are endogenously the result of knowledge uncertainty and information processing constr- aints. Importantly, while recent work in imitation/appren- ticeship have studied aspects of subjective dynamics that
can be jointly learnable \cite{reddy2018you,herman2016inverse,herman2016thesis,majumdar2017risk,singh2018risk}, they are limited to environments that are fully-observable and/or agents that have point-valued knowledge of environments---substantial simplifications that ignore how humans can and do make imperfect inferences from recognizing environment signals.}

\vspace{-0.18em}
\subsection{Important Distinctions}\label{app:a1}
\vspace{0.20em}

\dayum{
Our goal of \textit{understanding} in IDM departs from the standard objectives of imitation and apprenticeship learning. As a result, some caveats and distinctions warrant special attention as pertains assumptions, subjectivity, and model accuracy.}

\vspace{0.075em}
\textbf{Decision-maker vs. Investigator}~
As noted in Section \ref{sub:project}, the design of \smash{$\Theta$} (and specification of \smash{$\theta_{\text{norm}}$}) are not \textit{assumptions}: We are not making ``factual'' claims concerning the underlying psychological processes that govern human behavior; these are hugely complex, and are the preserve of neuroscience and biology \cite{tomov2020structure}. Instead, such specifications are active \textit{design choices}: We seek to make the ``effective'' claim that an agent is behaving \textit{as if} their generative mechanism were parameterized by the (interpretable) structure we designed for \smash{$\Theta$}. Therefore when we speak of ``assumptions'', it is important to distinguish between assumptions about the \textit{agent} (of which we make none), versus assumptions about the \textit{investigator} performing IDM (of which, by construction, we assume they have the ability to specify values for \smash{$\theta_{\text{norm}}$}).

\vspace{0.075em}
\dayum{
In IBRC, for example, in learning $\beta$ we are asking the question: ``How much (optimistic/pessimistic) deviation from neutral knowledge does the agent appear to tolerate?'' For this question to be meaningfully answered, we---as the inves- tigator---must be able to produce a meaningful value for $\tilde{\sigma}$ to specify as part of \smash{$\theta_{\text{norm}}$}. In most cases, \mbox{we are interested} in deviations from some notion of ``current medical knowledge'', or what knowledge an ``ideal'' clinician may be expected to possess; thus we may---for instance---supply a value for $\tilde{\sigma}$ via models learned a priori from data. Of course, coming up such values for \smash{$\theta_{\text{norm}}$} is not trivial (not to mention entirely dependent on the problem and the investigator's objectives regarding interpretability); however, we emphasize that this does not involve assumptions regarding the \textit{agent}.}

\begin{figure}[t]
\centering
\vspace{-0.7em}
\caption{\textit{Graphical Model}. In general, the environment's states (top) are only accessible via its emissions in response to actions (middle), which the agent incorporates by way of internal states (bottom). However, note that---unlike classic POMDP/IOHMM settings, here the agent's knowledge of the dynamics is subjective.}
\label{fig:pomdp}
\vspace{0.75em}
\includegraphics[width=\linewidth]{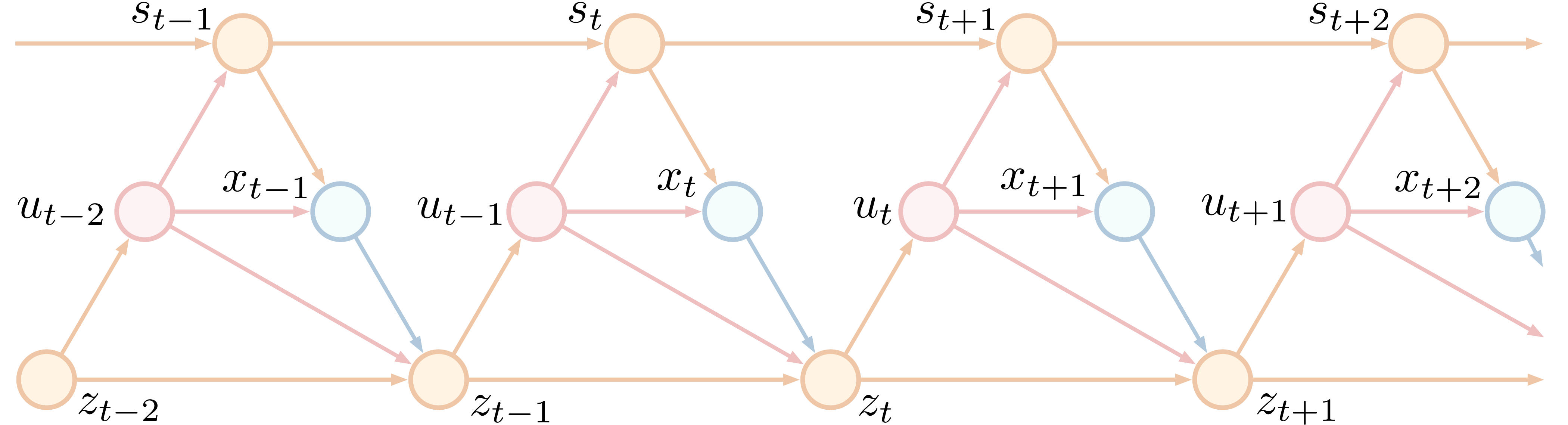}
\vspace{-1.75em}
\end{figure}

\begin{figure}[t]
\centering
\vspace{-0.5em}
\caption{\textit{Backup Diagram}. In IBRC, the backup operation (Theorem \ref{thm:values}) transfers value information across three recursive ``layers''---that is, of successor values for agent states ($V$), state-action pairs ($Q$), and state-action-model tuples ($K$). Indicated below are the utility and penalty terms collected along these backup operations.}
\label{fig:backup}
\vspace{0.75em}
\includegraphics[width=\linewidth]{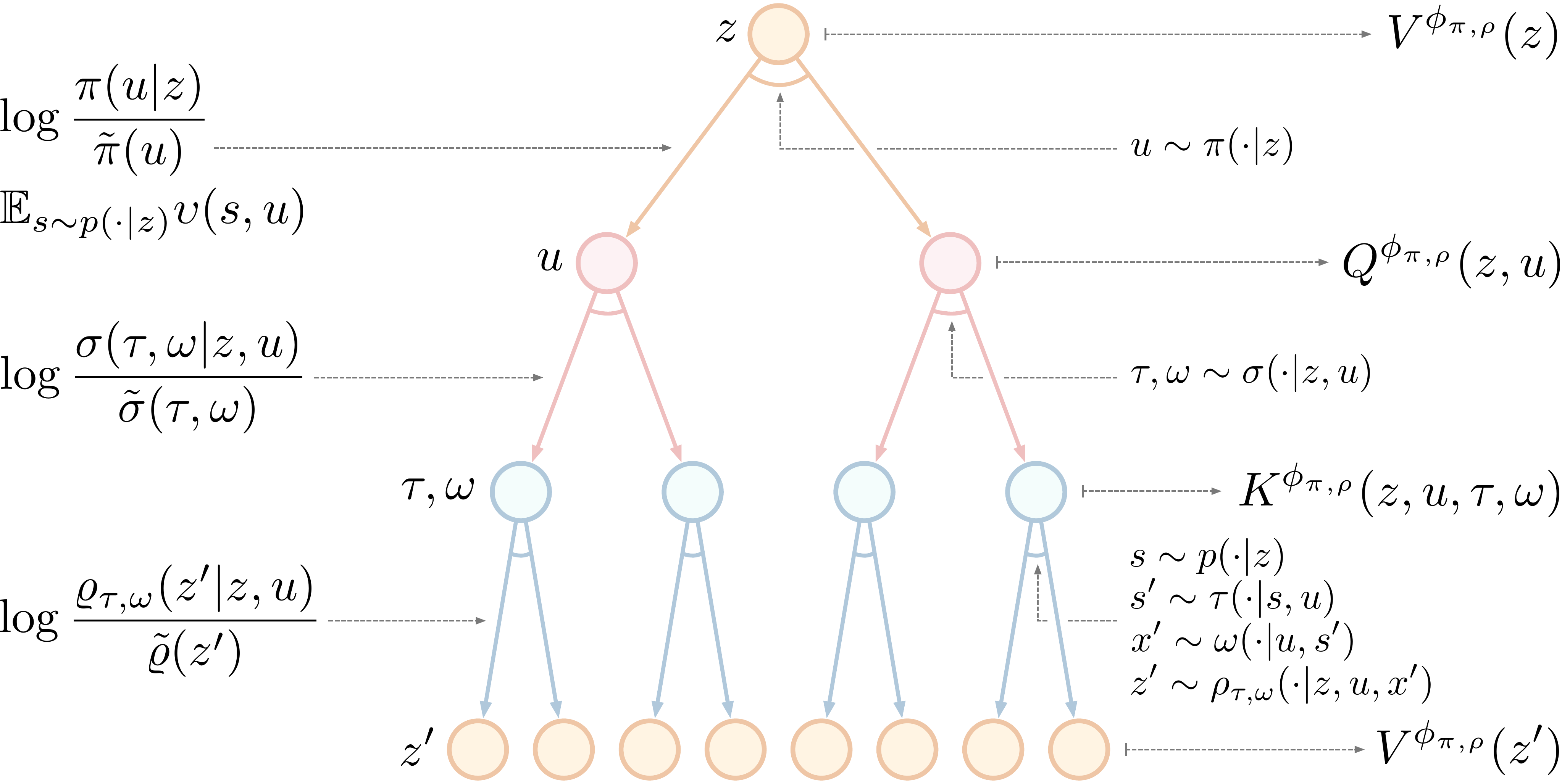}
\vspace{-2.25em}
\end{figure}

\vspace{0.075em}
\textbf{Subjective vs. Objective Dynamics}~
In imitation and app- renticeship learning, parameterizations of utilities and dyna- mics models are simply \textit{intermediaries} for the downstream task (of replicating expert actions or matching expert returns). As a result, no distinction needs be made between the ``external'' environment (with \textit{objective} dynamics \smash{$\tau_{\text{env}},\omega_{\text{env}}$}) and the ``internal'' environment model that an agent works with (with \textit{subjective} dynamics $\tau,\omega$). Indeed, if the learned model were to be evaluated based on live deployment in the real environment (as is the case in IL/IRL), it only makes sense that we stipulate \smash{$\tau,\omega=\tau_{\text{env}},\omega_{\text{env}}$} for the best results.

\dayum{
However, in IDM (and IBRC) we are precisely accounting for how an agent may appear to deviate from such perfect, point-valued knowledge of the environment. Disentangling subjective and objective dynamics is now critical: Both the forward recursion (Lemma \ref{thm:forward}) for occupancy measures and the backward recursion (Theorem \ref{thm:values}) for value functions are computations \textit{internal} to the agent's mind---and need not correspond to any notion of true environment dynamics. The \textit{external} dynamics only comes into play when considering the distribution of trajectories \smash{$h\sim\phi_{\pi,\rho}$} induced by an agent's policies, which---by definition---manifests through (actual or potential) interaction with the real environment.}

\dayum{
\textbf{Demonstrated vs. Projected Behavior}~
As advanced thro- ughout, a primary benefit of the generalized perspective we develop is that we may ask \textit{normative-descriptive questions} taking the form: ``Given that this (boundedly rational) agent should optimize this $\upsilon$, how suboptimally do they appear to behave?'' Precisely, as pertains IBRC we noted that---as the investigator---we are free to specify (what we deem) ``meaningful'' values for $\upsilon$ within \smash{$\theta_{\text{norm}}$}, while recovering one or more behavioral parameters $\alpha,\beta,\gamma$ from \smash{$\theta_{\text{desc}}$}. Clearly, however, we are not at liberty to specify completely random
values for $\upsilon$ (or, more generally, that we are not at liberty to design $\Theta$ and \smash{$\theta_{\text{norm}}$} in an entirely arbitrary fashion). For one, the resulting inverse decision model may simply be a poor reflection the original behavior (i.e. the projection \smash{$\phi_{\text{imit}}^{*}$} onto \smash{$\Phi_{\theta_{\text{norm}}}$} may simply lose too much information from \smash{$\phi_{\text{demo}}$}.\footnote{Abstractly, this is not dissimilar to any type of model fitting problem: If the mismatch between the (unknown) data generating pro- cess and the (imposed) structure of the model is too great, then the quality of the model---by any reasonable measure---would suffer.}}

\dayum{
Without doubt, the usefulness of the inverse decision model (i.e. in providing valid interpretations of observed behavior) depends entirely on the design and specification of $\Theta$ and $\theta_{\text{norm}}$, which requires care in practice. Most importantly, it should be verified that---under our designed parameteriza- tion---the \textit{projected} behavior \smash{$\phi_{\text{imit}}^{*}$} is still a faithful model of the \textit{demonstrated} behavior \smash{$\phi_{\text{demo}}$}. In particular, compared with fitting a black-box model for imitating behavior---or any standard method for imitation/apprenticeship learning, for that matter---it should be verified that our (interpretably parameterized) model does not suffer inordinately in terms of accuracy measures (i.e. in predicting $u$ from $h$); otherwise the model (and its interpretation) would not be meaningful. In Appendix \ref{app:b}, we perform precisely such a sanity check for IBRC, using a variety of standard benchmarks (Table \ref{tab:results}).}

\vspace{-0.20em}
\subsection{Further Related Work}
\vspace{0.20em}

While relevant works have been noted throughout the manus- cript, here we provide additional context for IDM and IBRC, and how notable techniques/frameworks relate to our work.

\textbf{Inverse Decision Modeling}~
Pertinent methods subsumed by our forward and inverse formalisms have been noted in Tables \ref{tab:forward}--\ref{tab:inverse}. In particular, techniques that can be formalized as instantiations of IDM are enumerated in Table \ref{tab:project}. Broadly, for
\textit{imitation learning} these include behavioral cloning-like methods \cite{pomerleau1991efficient,bain1999framework,ross2010efficient,syed2010reduction,syed2007imitation,ross2011reduction,piot2014boosted,jarrett2020strictly
}, as well as distribution-matching methods that directly match occupancy measures \cite{blonde2019sample,kostrikov2019discriminator,ho2016generative,jeon2018bayesian,ghasemipour2019understanding,ghasemipour2019divergence,ke2019imitation,ke2020wafr,kim2018imitation,xiao2019wasserstein,dadashi2021primal,kostrikov2020imitation,arenz2020non,srinivasan2020interpretable,zhang2020f,baram2016model,baram2017model}; we defer to \cite{yue2018imitation,osa2018imitation} for more thorough surveys. For \textit{apprenticeship learning} by inverse reinforcement learning, these include
classic maximum-margin methods based on feature expec- tations \cite{ng2000algorithms,abbeel2004apprenticeship,syed2008game,syed2008apprenticeship,klein2011batch,mori2011model,lee2019truly}, maximum likelihood soft policy matching using Boltzmann-rational policies \cite{neu2007apprenticeship,babes2011apprenticeship}, maximum entropy policies
\cite{ziebart2010modeling,zhou2017infinite,lee2018maximum,mai2019generalized,jain2019model}, and Bayesian maximum a posteriori inference \cite{ramachandran2007bayesian,choi2011map,dimitrakakis2011bayesian,rothkopf2011preference,balakrishnan2020efficient}, as well as methods that leverage preference models and annotations for learning \cite{asri2016score,burchfiel2016distance,jacq2019learning,brown2019extrapolating,brown2020better}.

In this context, the novelty of the IDM formalism is two-fold. First, in defining a unifying framework that generalizes all prior techniques, IDM simultaneously opens up a new class of problems in \textit{behavior representation learning} with consciously designed parameterizations. Specifically, in defining inverse decision models as projections in $\Phi$-space induced by $F,G$, and $\Theta$, the structure and decomposition chosen for \smash{$\Theta_{\text{norm}}\times\Theta_{\text{desc}}$} allows asking normative-descriptive questions that seek to \textit{understand} observed decision-making behavior. Second, in elevating \textit{recognition policies} to first-class citizenship in partially-observable environments, IDM greatly generalizes the notion of ``boundedness'' in decision-making---that is, from the existing focus on noisy optimality in $\pi$, to the ideas of subjective dynamics $\sigma$ and biased belief-updates $\rho$ (viz. discussion in the beginning of this section).

\dayum{
\textbf{Orthogonal Frameworks}~
Multiple studies have proposed frameworks that provide generalized treatments of different aspects of inverse reinforcement learning \cite{neu2009training,choi2011map,ghasemipour2019divergence,ke2020wafr,ni2020f,arenz2020non,jeon2020reward}. However, these are \textit{orthogonal} to our purposes in the sense that they are primarily concerned with establishing connections between different aspects/subsets of the imitation/apprenticeship learning literature. These include loss-function perspectives \cite{neu2009training} and Bayesian MAP perspectives \cite{choi2011map} on inverse reinforcement learning, $f$-divergence minimization perspectives \cite{ghasemipour2019divergence,ke2020wafr} on distribution matching, connections between adversarial and non-adversarial methods for distribution matching \cite{arenz2020non}, as well as different problem settings for learning reward functions \cite{jeon2020reward}. But relative to the IDM formalism, all such frameworks operate within the special case of \smash{$\theta_{\text{desc}}$\pix$=$\pix$\upsilon$} (and full observability).}

\dayum{
\textbf{Case Study: GAIL}~
Beyond aforementioned distinctions, another implication is that IDM defines a single language for understanding key results in such prior works. For example, we revisit the well-known result in \cite{ho2016generative} that gives rise to generative adversarial imitation learning (``GAIL''): It is instructive to recast it in more general---but simpler---terms.
First, consider a \textit{maximum entropy} learner in the MDP setting (cf. Table \ref{tab:forward}), paired with a \textit{maximum margin} identification strategy with a parameter regularizer $\zeta$ (cf. Table \ref{tab:inverse}):}
\vspace{-0.55em}
\begin{gather}
\begin{aligned}
F_{\hspace{-1pt}\theta_{\text{norm}}}^{\scalebox{0.6}{\pix\text{ME}}}\hspace{-2pt}(\theta_{\text{desc}})
\hspace{-2pt}\doteq\hspace{-1pt}
\phi_{\pi^{*}}~\text{where}~\pi^{*}
\hspace{-2pt}\doteq\hspace{-1pt}
\text{argmax}_{\pi}
\mathbb{E}_{z\sim\rho_{0}}\hspace{-2pt}
V_{\text{soft},\theta}^{\phi_{\pi}}(z)~~~~~~~~
\end{aligned}\raisetag{0.91\baselineskip}
\end{gather}
\vspace{-1.9em}
\begin{gather}
\begin{aligned}
G_{\hspace{-1pt}\theta_{\text{norm}}}^{\scalebox{0.6}{\pix\text{MM}}}\hspace{-2pt}(\phi\hspace{0.25pt})
\hspace{-2pt}\doteq\hspace{-2pt}
\text{argmin}_{\theta_{\text{desc}}}\hspace{-2pt}
\mathbb{E}_{z\sim\rho_{0}}\hspace{-1pt}[
V_{\hspace{-1.5pt}\text{soft},\theta}^{\phi_{\text{imit}}}\hspace{-1pt}(z)\hspace{-2.5pt}-\hspace{-3pt}V_{\theta}^{\phi}\hspace{-1pt}(z)
]
\hspace{-2.5pt}+\hspace{-2pt}
\zeta(\hspace{-1pt}\theta_{\text{desc}}\hspace{-0.5pt})~~~~~~~~
\end{aligned}\raisetag{0.98\baselineskip}
\end{gather}
\vspace{-1.6em}

\dayum{
Second, consider a black-box \textit{decision-rule} policy (cf. Table \ref{tab:forward}), where neural-network weights $\chi$ directly parameterize a policy network \smash{$f_{\text{decision}}$} (and \smash{$\theta_{\text{desc}}=\chi$}); this is paired with a \textit{distribution matching} identification strategy (cf. Table \ref{tab:inverse}):}
\vspace{-0.7em}
\begin{equation}
F_{\theta_{\text{norm}}}^{\scalebox{0.6}{\pix\text{DR}}}(\theta_{\text{desc}})
\doteq
\text{argmax}_{\pi}\pix\delta(\pi-f_{\text{decision}}(\chi))
\end{equation}
\vspace{-1.1em}
\begin{equation}
G_{\theta_{\text{norm}}}^{\scalebox{0.6}{\pix\text{DM}}}(\phi\hspace{0.25pt})
\hspace{-2pt}\doteq\hspace{-1pt}
\text{argmin}\hspace{-1pt}\raisebox{-4pt}{$_{\theta_{\text{desc}}}$}\hspace{-10pt}
\zeta^{*}\hspace{-1pt}(\phi_{\text{demo}}
\hspace{-3pt}-\hspace{-2pt}\phi_{\text{imit}})
\hspace{-2pt}-\hspace{-2pt}
\mathcal{H}_{\text{imit}}
\end{equation}
\vspace{-1.4em}

\dayum{
where distance measures are given by the convex conjugate \smash{$\zeta^{*}$}, and \smash{$\mathcal{H}_{\text{imit}}$} gives the causal entropy of the imitating policy.
Now, the primary motivation behind generative adversarial imitation learning is the observation that \smash{$\zeta$}-regularized maximum-margin soft IRL implicitly seeks a policy whose occupancy is close to the demonstrator's as measured by \smash{$\zeta^{*}$}. In IDM, this corresponds to a remarkably simple statement:}

\begin{reproposition}[restate=gail,name=Ho and Ermon{,} Recast]\upshape\label{thm:gail}
\dayum{ Define the \textit{beha- vior projections} induced by the composition of each pairing:}
\vspace{-0.6em}
\begin{equation}
\text{proj}_{\Phi_{\theta_{\text{norm}}}}^{\pix\scalebox{0.6}{\text{ME},\pix\text{MM}}}
\doteq
F_{\theta_{\text{norm}}}^{\scalebox{0.6}{\pix\text{ME}}}
\circ
G_{\theta_{\text{norm}}}^{\scalebox{0.6}{\pix\text{MM}}}
\end{equation}
\vspace{-1.2em}
\begin{equation}
\text{proj}_{\Phi_{\theta_{\text{norm}}}}^{\pix\scalebox{0.6}{\text{DR},\pix\text{DM}}}\doteq
F_{\theta_{\text{norm}}}^{\scalebox{0.6}{\pix\text{DR}}}
\circ
G_{\theta_{\text{norm}}}^{\scalebox{0.6}{\pix\text{DM}}}
\end{equation}
\vspace{-1.5em}

Then these projections are identical: \smash{$
\text{proj}\raisebox{1pt}{$_{\Phi_{\theta_{\text{norm}}}}^{\pix\scalebox{0.6}{\text{ME},\pix\text{MM}}}$}
=
\text{proj}\raisebox{1pt}{$_{\Phi_{\theta_{\text{norm}}}}^{\pix\scalebox{0.6}{\text{DR},\pix\text{DM}}}$}
$} (and inverse decision models thereby obtained are identical).
\end{reproposition}
\vspace{-0.5em}

In their original context, the significance of this lies in the fact that the first pairing explicitly requires parameterizations via reward functions (which---in classic apprenticeship methods---is restricted to be linear/convex), whereas the second pairing allows arbitrary parameterization by neural networks (which---while black-box---are more flexible). In our language, this simply means that the first projection req- uires \smash{$\theta_{\text{desc}}$\pix$=$\pix$\upsilon$}, while the second projection allows \smash{$\theta_{\text{desc}}$\pix$=$\pix$\chi$}.

\textbf{Inverse Bounded Rational Control}~
Pertaining to IBRC, methods that are comparable and/or subsumed have been noted in Tables \ref{tab:project} and \ref{tab:agent}. In addition, the context of IBRC within existing notions of bounded rationality have been discussed in detail in the beginning of this section. Now, more broadly, we note that the study of imperfect behaviors \cite{wheeler2018bounded} spans multiple disciplines: in cognitive science \cite{griffiths2015rational}, biological systems \cite{genewein2015bounded}, behavioral economics \cite{augenblick2018belief}, and information theory \cite{ortega2015information}. Specifically, IBRC generalizes this latter class of information-theoretic approaches to bounded rationality.
First, the notion of \textit{flexibility} in terms of the informational effort in determining successive actions (cf. decision complexity) is present in maximum entropy \cite{haarnoja2017reinforcement,haarnoja2018soft,eysenbach2019if,shi2019soft} and KL-regularized \cite{rubin2012trading,galashov2019information,ho2020efficiency,tiomkin2017unified,leibfried2017information} agents. Second, the notion of \textit{toler- ance} in terms of the statistical surprise in adapting to successive beliefs (cf. recognition complexity) is present in behavioral economics \cite{ely2015suspense,augenblick2018belief} and decision theory \cite{alaa2016balancing,tishby2011information,jarrett2020inverse}. Third, the notions of \textit{optimism} and \textit{pessimism} in terms of the average regret in deviating from prior knowledge (cf. specification complexity) are present in robust planners \cite{petersen2000minimax,charalambous2004relations,grau2016planning,osogami2012robustness}.

\dayum{
On account of this, the novelty of the IBRC example is three-fold. First, it is the first to present generalized recursions incorporating all three notions of complexity---that is, in the mappings into internal states, models, and actions. Second, IBRC does so in the partially-observable setting, which---as noted in above discussions---crucially generalizes the idea of subjective dynamics into subjective beliefs, thereby accounting for boundedness in the recognition process itself. Third (perhaps most importantly), IBRC is the first to consider the \textit{inverse problem}---that is, of turning the entire formalism on its head to \textit{learn} the parameterizations of such boundedness, instead of simply \textit{assuming} known parameters as required by the forward problem. Finally, it is important to note that IBRC is simply one example: There are of course many possibilities for formulating boundedness, including such aspects as myopia and temporal inconsistency \cite{evans2015learning,evans2015learning}; we leave such applications for future work.}

\textbf{Interpretable Behavior Representations}~
Lastly, \mbox{a variety} of works have approached the task of representing behaviors in an interpretable manner. In inverse reinforcement learning, multiple works have focused on the \textit{reward function} itself, specifying interpretable structures that explicitly express a decision-maker's preferences \cite{rothkopf2011preference}, behavior under time pressure \cite{jarrett2020inverse}, consideration of counterfactual outcomes \cite{bica2021learning}, as well as intended goals \cite{zhi2020online}. Separately, another strand of research has focused on imposing interpretable structures onto \textit{policy functions} themselves, such as representing policies in terms of decision trees \cite{bastani2018verifiable} and intended outcomes \cite{yau2020did} in the forward problem, or---in the inverse case---learning imitating policies based on decision trees \cite{bewley2020modelling} or decision boundaries \cite{huyuk2021explaining}. In the context of IDM, both of these approaches can naturally be viewed as instantiations of our more general approach of learning representations of behavior through interpretably parameterized \textit{planners} and \textit{inverse planners}  (as noted throughout Tables \ref{tab:project}--\ref{tab:inverse}). Finally, for completeness also note that an orthogonal branch of research is dedicated to generating \textit{autonomous explanations} of artificial behavior, as suggested updates to human models \cite{chakraborti2017plan,chakraborti2019plan}, and also as responses to human queries in a shared \cite{hayes2017improving} or user-specified vocabulary \cite{sreedharan2020bridging}.

\vspace{-0.4em}
\subsection{Future Work}

\dayum{
A clear source of potential research lies in exploring differ- ently structured parameterizations $\Theta$ to allow interpretable representation learning of behaviors. After all, beyond the black-box and reward-centric approaches in Table \ref{tab:project} and the handful of works that have sought to account for subjective dynamics \cite{huyuk2021explaining,reddy2018you,herman2016inverse,kwon2020inverse}, our example of IBRC is only one such prototype that exercises the IDM formalism more fully.
In developing more complex and/or expressive forward mod- els, an important question to bear in mind is to what extent the inverse problem is identifiable. In most existing cases we have seen, the usual strategies---such as constraining scaling, shifting, reward shaping, as well as the use of Bayesian inference---is sufficient to recover meaningful values. How- ever, we have also seen that in the extreme case of an arbit- rary differentiable planner, any inverse problem immediately falls prey to the ``no free lunch'' result \cite{armstrong2018occam,christiano2015medium,shah2018inferring,shah2019feasibility}.
Thus balancing aspects of complexity, interpretability, and identifiability of decision models would be an interesting direction of work.
Finally, in this work we primarily focused on the idea of limited \textit{intentionality}---that is, in the goal-seeking nature of an agent and how they may be constrained in this respect. But the flip side is also interesting: One can explore the idea of limited \textit{attentionality}---that is, in how an agent may be constrained in their ability to focus on sequences of past events. This idea is explored in \cite{fox2016memoryless,fox2016retentive} by analogy with information bottlenecks in sensors and memory capacities; however, there is much room for developing more human-interpretable parameterizations of how an agent may pay selective attention to observations over time.}

\clearpage
\section{Experiment Details}\label{app:b}

\textbf{Computation}~
In IBRC, we define the space of agent states (i.e. subjective beliefs) as $\mathcal{Z}\doteq\mathbb{R}^{k}$, where $k$ is the number of world states ($k$\pix$=$\pix$3$ for ADNI, and $k$\pix$=$\pix$2$ for DIAG). To implement the \textit{backward recursion} (Theorem \ref{thm:values}), each dimension of $\mathcal{Z}$ is discretized with a resolution of 100, and the values $V(z)$ in the resulting lattice are updated iteratively exactly according to the backup operator $\mathbb{B}^{*}$---until convergence (which is guaranteed by the fact that $\mathbb{B}^{*}$ is contractive, therefore the fixed point is unique; see Appendix \ref{app:c}). For evaluation at any point $z$, we (linearly) interpolate between the closest neighboring grid points. In terms of implement- ing the \textit{inverse problem} in a Bayesian manner (i.e. to recover posterior distributions over $\Theta_{\text{desc}}$), we perform MCMC in log-parameter space (i.e. $\log\alpha,\log\beta,\log\eta$). Specifically, the proposal distribution is zero-mean Gaussian with standard deviation 0.1, with every 10th step collected as a sample. In each instance, the initial 1,000 burn-in samples are discarded, and a total of 10,000 steps are taken after burn-in.

\dayum{
\textbf{Recognition}~
In the manuscript, we make multiple referenc- es to the \textit{Bayes update}, in particular within the context of our (possibly-biased) belief-update (Equation \ref{eq:recognition}). For completeness, we state this explicitly: Given point-valued knowledge of $\tau,\omega$, update \smash{$\rho_{\tau,\omega}(z^{\prime}|z,u,x^{\prime})$} is the Dirac delta centered at}
\vspace{-0.7em}
\begin{gather}
\begin{aligned}
p(s^{\prime}|z,\hspace{-1pt}u,\hspace{-0.5pt}x^{\prime}\hspace{-2pt},\hspace{-1pt}\tau,\hspace{-0.5pt}\omega)
\hspace{-1pt}\doteq\hspace{-1pt}
\mathbb{E}_{s\sim p(\cdot|z)}
\hspace{-2pt}
\bigg[\frac{
\tau(s^{\prime}|s,u)\omega(x^{\prime}|u,s^{\prime})
}{
\mathbb{E}_{s^{\prime}\sim\tau(\cdot|s,u)}\omega(x^{\prime}|u,s^{\prime})
}\bigg]~~~~~~~
\end{aligned}\raisetag{1.25\baselineskip}
\end{gather}
\vspace{-1.em}

\dayum{
and the overall recognition policy is the expectation over such values of $\tau,\omega$ (Equation \ref{eq:recognition}). As noted in Section \ref{sub:rational}, in general $\tilde{\sigma}$ represents any prior distribution the agent is specified to have, and in particular can be some Bayesian posterior $p(\tau,\omega|\mathcal{E})$ given any form of experience $\mathcal{E}$. This can be modeled in any manner, and is not the focus of our work; what matters here is simply that the agent may \textit{deviate} optimistically/pessimistically from such a prior. As noted in Section \ref{sec:simulate}, for our purposes we simulate $\tilde{\sigma}$ by discretizing the space of models such that probabilities vary in $\pm10\%$ increments from the (highest-likelihood) truth. In ADNI, this means \smash{$\tilde{\sigma}$} is centered at the IOHMM learned from the data.}

\dayum{
\textbf{Model Accuracy}~
In Appendix \ref{app:a1} we discussed the caveat: In order for an inverse decision model to provide valid \textit{interpretations} of observed behavior, it should be verified that---under the designed parameterization---the projected behavior \smash{$\phi_{\text{imit}}^{*}$} is still an \textit{accurate} model of the demonstrated behavior \smash{$\phi_{\text{demo}}$}. Here we perform such a sanity check for our IBRC example using the ADNI environment. We consider the following standard \textit{benchmark algorithms}. First, in terms of black-box models for imitation learning, we consider behavioral cloning \cite{bain1999framework} with a recurrent neural network for observation-action histories (\textbf{RNN-Based BC-IL}); an adaptation of model-based imitation learning \cite{babuska2012model} to partially-observable settings, using the learned IOHMM as model (\textbf{IOHMM-Based BC-IL}); and a recently-proposed model-based imitation learning that allows for subjective dynamics \cite{huyuk2021explaining} by jointly learning the agent's possibly-biased internal model and their probabilistic decision boundaries (\textbf{Joint IOHMM-Based BC-IL}). Second, in terms of classic reward-centric methods for apprenticeship learning, we consider Bayesian inverse reinforcement learning in partially-observable environments \cite{jarrett2020inverse} equipped with the learned IOHMM as model (\textbf{Bayesian PO-IRL}); and---analogous to the black-box case---the equivalent of this method that trains the dynamics model jointly along with the agent's apprenticeship policy \cite{makino2012apprenticeship} (\textbf{Joint Bayesian PO-IRL}). Algorithms requiring learned models are given IOHMMs estimated using conventional methods \cite{bengio1995input}---which is the same method by which the true model is estimated in IBRC (that is, as part of the space of candidate models in the support of $\tilde{\sigma}$).}

\begin{table}[t]\small
\newcolumntype{A}{>{          \arraybackslash}m{4.5cm}}
\newcolumntype{B}{>{\centering\arraybackslash}m{2.2cm}}
\newcolumntype{C}{>{\centering\arraybackslash}m{2.2cm}}
\setlength{\cmidrulewidth}{0.5pt}
\setlength\tabcolsep{0pt}
\renewcommand{\arraystretch}{1.025}
\vspace{-0.8em}
\caption{\textit{Comparison of Model Accuracies}. IBRC performs similarly to all benchmark algorithms in matching demonstrated actions. Results are computed using held-out samples based on 5-fold cross-validation. IBRC is slightly better-calibrated, and similar in precision-recall scores (differences are statistically insignificant).}
\vspace{-0.8em}
\label{tab:results}
\begin{center}
\begin{adjustbox}{max width=\linewidth}
\begin{tabular}{ABC}
\toprule
\textbf{Inverse Decision Model} & \textbf{Calibration (Low is Better)} & \textbf{PRC Score (High is Better)}\\
\midrule
\textbf{Black-Box Model}: \\
~~~RNN-Based BC-IL                  & 0.18 $\pm$ 0.05 & 0.81 $\pm$ 0.08 \\
~~~IOHMM-Based BC-IL                & 0.19 $\pm$ 0.07 & 0.79 $\pm$ 0.11 \\
~~~Joint IOHMM-Based BC-IL          & 0.17 $\pm$ 0.05 & 0.81 $\pm$ 0.09 \\
\midrule
\textbf{Reward-Centric Model}: \\
~~~Bayesian PO-IRL                  & 0.23 $\pm$ 0.01 & 0.78 $\pm$ 0.09 \\
~~~Joint Bayesian PO-IRL            & 0.24 $\pm$ 0.01 & 0.79 $\pm$ 0.09 \\
\midrule
\textbf{Boundedly Rational Model}: \\
~~~IBRC (with learned $\alpha,\beta,\eta$) & 0.16 $\pm$ 0.00 & 0.77 $\pm$ 0.01 \\
\bottomrule
\end{tabular}
\end{adjustbox}
\end{center}
\vspace{-1.5em}
\end{table}

\muline{\textit{Results}}. Table \ref{tab:results} shows results of this comparison on predicting actions, computed using held-out samples based on 5-fold cross-validation. Crucially, while IBRC has the advan- tage in terms of interpretability of parameterization, its per- formance---purely in terms of predicting actions---does not degrade: IBRC is slightly better in terms of calibration, and similar in precision-recall (differences are statistically insignificant), which---for our ADNI example---affirms the validity of IBRC as an (interpretable) representation of \smash{$\phi_{\text{demo}}$}.

\dayum{
\textbf{Data Selection}~
From the ADNI data, we first selected out anomalous cases without a cognitive dementia rating test  result, which is almost always taken at every visit by every patient. Second, we also truncated patient trajectories at points where a visit is skipped (that is, if the next visit of a patient does not occur immediately after the 6-monthly period following the previous visit). This selection process leaves 1,626 patients out of the original 1,737, and the median number of consecutive visits for each patient is three. In measuring MRI outcomes, the ``average'' is defined to be within half a standard deviation of the population mean. Note that this is the same pre-processing method employed for ADNI in \cite{huyuk2021explaining}.}

{\setstretch{1.0065}
\dayum{
\textbf{Implementation}~
Details of implementation for benchmark algorithms follow the setup in \cite{huyuk2021explaining}, and are reproduced here:
\muline{\textit{RNN-Based BC-IL}}: We train an RNN whose inputs are the observed histories $h$ and whose outputs are the predicted probabilities $\hat{\pi}(u|h)$ of taking action $u$ given the observed history $h$. The network consists of an LSTM unit of size $64$ and a fully-connected hidden layer of size $64$. The cross-entropy \smash{$\mathcal{L}$\pix$=$\pix$-\sum_{n=1}^N\sum_{t=1}^{T}\sum_{u\in\mathcal{U}}\mathbb{I}\{u_t$\pix$=$\pix$u\}\log\hat{\pi}(u|h)$} is minimized using the Adam optimizer with a learning rate of $0.001$ until convergence (that is, when the loss does not improve for $100$ consecutive iterations).
\muline{\textit{Bayesian PO-IRL}}:
The IOHMM parameters are initialized by sampling uniformly at random. Then, they are estimated and fixed using conventional IOHMM methods. The utility $\upsilon$ is initialized as \smash{$\hat{\upsilon}^0(s,u)$\pix$=$\pix$\varepsilon_{s,u}$}, where \smash{$\varepsilon_{s,u}$\pix$\sim$\pix$\mathcal{N}(0,0.001^2)$}. Then, it is estimated via MCMC sampling, during which new candidate samples are generated by adding Gaussian noise with standard deviation $0.001$ to the previous sample. To form the final estimate, we average every 10th sample among the second set of $500$ samples, ignoring the first $500$ samples. To compute optimal $Q$-values, we use an off-the-shelf POMDP solver \scalebox{0.93}{\texttt{https://www.pomdp.org/code/index.html}}.
\muline{\textit{Joint Bayesian PO-IRL}}:
All parameters are initialized exactly the same way as in Bayesian PO-IRL. Then, both the IOHMM parameters and the utility are estimated jointly via MCMC sampling. In order to generate new candidate samples, with equal probabilities we either sample new IOHMM parameters from the posterior (but without changing $\upsilon$) or obtain a new $\upsilon$ the same way we do in Bayesian PO-IRL (but without changing the IOHMM parameters). A final estimate is formed the same way as in Bayesian PO-IRL.
\muline{\textit{IOHMM-Based BC-IL}}:
The IOHMM parameters are initialized by sampling them uniformly at random. Then, they are estimated and fixed using conventional IOHMM methods. Given the IOHMM parameters, we parameterize policies using the method of \cite{huyuk2021explaining}, with the policy parameters $\{\mu_u\}_{u\in\mathcal{U}}$ (not to be confused with the occupancy measure ``$\mu$'' as defined in the present work) initialized as \smash{$\hat{\mu}^0_u(s)=(1/|S|+\varepsilon_{u,s})/\sum_{s'\in S}(1/|S|+\varepsilon_{u,s'})$}, where \smash{$\varepsilon_{u,s'}$\pix$\sim$\pix$\mathcal{N}(0,0.001^2)$}. Then, they are estimated according solely to the action likelihoods in using the EM algorithm. The expected log-posterior is maximized using the Adam optimizer with learning rate $0.001$ until convergence (that is, when the expected log-posterior does not improve for $100$ consecutive iterations).
\muline{\textit{Joint IOHMM-Based BC-IL}}:
This corresponds exactly to the proposed method of \cite{huyuk2021explaining} itself, which is similar to IOHMM-Based BC-IL except parameters are trained jointly. All parameters are initialized exactly the same way as before; then, the IOHMM parameters and the policy parameters are estimated jointly according to both the action likelihoods and the observation likelihoods simultaneously. The expected log-posterior is again maximized using the Adam optimizer with a learning rate of $0.001$ until convergence (non-improvement for $100$ consecutive iterations).}
}

\section{Proofs of Propositions}\label{app:c}

\forward*

\textit{Proof}. Start from the definition of $\mathbb{M}_{\pi,\rho}$; episodes are res- tarted on completion ad infinitum, so we can write \smash{$\mu_{\pi,\rho}$} as:
\begin{gather}
\begin{aligned}
\mu_{\pi,\rho}(z)
&\doteq
(1-\gamma)
\textstyle\sum_{t=0}^{\infty}\gamma^{t}
p(z_{t}=z|z_{0}\sim\rho_{0}) \\
&=
(1-\gamma)
\textstyle\sum_{t=0}^{\infty}\gamma^{t}
((\mathbb{M}_{\pi,\rho})^{t}\rho_{0})(z)
\end{aligned}
\end{gather}
Then we obtain the result by simple algebraic manipulation:

\vspace{-1em}
\begin{gather}
\begin{aligned}
&~~~~~\hspace{0.75pt}
(1-\gamma)\rho_{0}(z)+\gamma(\mathbb{M}_{\pi,\rho}\mu_{\pi,\rho})(z) \\
&=
(1-\gamma)\rho_{0}(z)+\gamma(1-\gamma)\textstyle\sum_{t=0}^{\infty}\gamma^{t}((\mathbb{M}_{\pi,\rho})^{t+1}\rho_{0})(z) \\
&=
(1-\gamma)(\rho_{0}(z)+\textstyle\sum_{t=0}^{\infty}\gamma^{t+1}((\mathbb{M}_{\pi,\rho})^{t+1}\rho_{0})(z)) \\
&=
(1-\gamma)\textstyle\sum_{t=0}^{\infty}\gamma^{t}((\mathbb{M}_{\pi,\rho})^{t}\rho_{0})(z) \\
&=
\mu_{\pi,\rho}(z)
\end{aligned}\raisetag{0.88\baselineskip}
\end{gather}

\vspace{-1em}
\dayum{
For uniqueness, we use the usual conditions---that is, that the process induced by the environment and the agent's policies is ergodic, with a single closed communicating class.}

\backward*

\textit{Proof}. Start with the Lagrangian, with \smash{$V$\pix$\in$\pix$\mathbb{R}^{\mathcal{Z}}$: $\mathcal{L}_{\pi,\rho}(\mu,V)$}
\begin{gather}
\begin{aligned}
&\doteq
J_{\pi,\rho}-\langle V,\mu-\gamma\mathbb{M}_{\pi,\rho}\mu-(1-\gamma)\rho_{0}\rangle \\
&=
\mathbb{E}_{\substack{
z\sim\mu_{\pi,\rho}\\
s\sim p(\cdot|z)\\
u\sim\pi(\cdot|z)
}}
\upsilon(s,u)
-
\langle V,\mu-\gamma\mathbb{M}_{\pi,\rho}\mu-(1-\gamma)\rho_{0}\rangle \\
&=
\mathbb{E}_{\substack{
z\sim\mu_{\pi,\rho}\\
s\sim p(\cdot|z)\\
u\sim\pi(\cdot|z)
}}
\upsilon(s,u)
+
\mathbb{E}\hspace{-7pt}_{\substack{
z\sim\mu_{\pi,\rho}\\
s\sim p(\cdot|z)\\
u\sim\pi(\cdot|z)\\
\tau,\omega\sim\sigma(\cdot|z,u)\\
s^{\prime}\sim\tau(\cdot|s,u)\\
x^{\prime}\sim\omega(\cdot|u,s^{\prime})\\
z^{\prime}\sim\rho(\cdot|z,u,x^{\prime})
}}\hspace{-8pt}
\gamma
V(z^{\prime}) \\
&~~~~-
\mathbb{E}_{z\sim\mu_{\pi,\rho}}
V(z) 
+
\langle V,(1-\gamma)\rho_{0}\rangle
\end{aligned}\raisetag{0.9\baselineskip}
\end{gather}

\begin{gather}
\begin{aligned}
&=
\mathbb{E}_{\substack{
z\sim\mu_{\pi,\rho}\\
s\sim p(\cdot|z)
}}[
\mathbb{E}_{u\sim\pi(\cdot|z)}[
\upsilon(s,u) \\
&~~~~+
\mathbb{E}_{\substack{
\tau,\omega\sim\sigma(\cdot|z,u)\\
s^{\prime}\sim\tau(\cdot|s,u)\\
x^{\prime}\sim\omega(\cdot|u,s^{\prime})\\
z^{\prime}\sim\rho(\cdot|z,u,x^{\prime})
}}
\gamma
V(z^{\prime})
]
-
V(z)
]
+
\langle V,(1-\gamma)\rho_{0}\rangle
\end{aligned}\raisetag{4\baselineskip}
\end{gather}
Then taking the gradient w.r.t. $\mu$ and setting it to zero yields:
\begin{gather}
\begin{aligned}
V(z)
=
\mathbb{E}_{\substack{
s\sim p(\cdot|z)\\
u\sim\pi(\cdot|z)
}}
[
\upsilon(s,u)
+
\mathbb{E}_{\substack{
\tau,\omega\sim\sigma(\cdot|z,u)\\
s^{\prime}\sim\tau(\cdot|s,u)\\
x^{\prime}\sim\omega(\cdot|u,s^{\prime})\\
z^{\prime}\sim\rho(\cdot|z,u,x^{\prime})
}}
\gamma
V(z^{\prime})
]~~~~~
\end{aligned}\raisetag{2.55\baselineskip}
\end{gather}
For uniqueness, observe as usual that $\mathbb{B}_{\pi,\rho}$ is $\gamma$-contracting:
\begin{gather}
\begin{aligned}
\|\mathbb{B}_{\pi,\rho}&V-\mathbb{B}_{\pi,\rho}V^{\prime}\|_{\infty} \\
&=
\text{max}_{z}\big|\mathbb{E}_{\substack{
u\sim\pi(\cdot|z) \\
\tau,\omega\sim\sigma(\cdot|z,u)\\
z^{\prime}\sim\varrho_{\tau,\omega}(\cdot|z,u)
}}\big[\gamma V(z^{\prime})-\gamma V^{\prime}(z^{\prime})\big]\big| \\
&\leq
\text{max}_{z}\mathbb{E}_{\substack{
u\sim\pi(\cdot|z) \\
\tau,\omega\sim\sigma(\cdot|z,u)\\
z^{\prime}\sim\varrho_{\tau,\omega}(\cdot|z,u)
}}\big[\big|\gamma V(z^{\prime})-\gamma V^{\prime}(z^{\prime})\big|\big] \\
&\leq
\text{max}_{z^{\prime}}\big|\gamma V(z^{\prime})-\gamma V^{\prime}(z^{\prime})\big| \\
&=
\gamma\|V-V^{\prime}\|_{\infty} \\
\end{aligned}
\end{gather}
which allows appealing to the contraction mapping theorem.

\backwardx*

\textit{Proof}. Start with the Lagrangian, now with the new multipliers $\alpha,\beta,\eta\in\mathbb{R}$ in addition to $V\in\mathbb{R}^{\mathcal{Z}}$: $\mathcal{L}_{\pi,\rho}(\mu,\alpha,\beta,\eta,V)$
\begin{gather}
\begin{aligned}
&\doteq
J_{\pi,\rho}-\langle V,\mu-\gamma\mathbb{M}_{\pi,\rho}\mu-(1-\gamma)\rho_{0}\rangle \\
&~~~~-\alpha\cdot(\mathbb{I}_{\pi,\rho}[\pi;\tilde{\pi}]-A)
-\beta\cdot(\mathbb{I}_{\pi,\rho}[\sigma;\tilde{\sigma}]-B) \\
&~~~~-\eta\cdot(\mathbb{I}_{\pi,\rho}[\varrho;\tilde{\varrho}]-C) \\
&=
\mathbb{E}_{\substack{
z\sim\mu_{\pi,\rho}\\
s\sim p(\cdot|z)\\
u\sim\pi(\cdot|z)
}}
\upsilon(s,u)
-
\langle V,\mu-\gamma\mathbb{M}_{\pi,\rho}\mu-(1-\gamma)\rho_{0}\rangle \\
&~~~~-\alpha\cdot(\mathbb{E}_{z\sim\mu_{\pi,\rho}}D_{\text{KL}}(\pi(\cdot|z)\|\tilde{\pi})-A) \\
&~~~~-\beta\cdot(\mathbb{E}_{\substack{z\sim\mu_{\pi,\rho}\\u\sim\pi(\cdot|z)}}D_{\text{KL}}(\sigma(\cdot|z,u)\|\tilde{\sigma})-B) \\
&~~~~-\eta\cdot(\mathbb{E}_{\substack{z\sim\mu_{\pi,\rho}\\u\sim\pi(\cdot|z)\\\tau,\omega\sim\sigma(\cdot|z,u)}}D_{\text{KL}}(\varrho_{\tau,\omega}(\cdot|z,u)\|\tilde{\varrho})-C)
\end{aligned}\raisetag{0.88\baselineskip}
\end{gather}

\begin{gather}
\begin{aligned}
&=
\mathbb{E}_{\substack{
z\sim\mu_{\pi,\rho}\\
s\sim p(\cdot|z)\\
u\sim\pi(\cdot|z)
}}
\upsilon(s,u)
+
\mathbb{E}\hspace{-7pt}_{\substack{
z\sim\mu_{\pi,\rho}\\
s\sim p(\cdot|z)\\
u\sim\pi(\cdot|z)\\
\tau,\omega\sim\sigma(\cdot|z,u)\\
s^{\prime}\sim\tau(\cdot|s,u)\\
x^{\prime}\sim\omega(\cdot|u,s^{\prime})\\
z^{\prime}\sim\rho(\cdot|z,u,x^{\prime})
}}\hspace{-8pt}
\gamma
V(z^{\prime}) \\
&~~~~-
\mathbb{E}_{z\sim\mu_{\pi,\rho}}
V(z) 
+
\langle V,(1-\gamma)\rho_{0}\rangle \\
&~~~~-\alpha\cdot(\mathbb{E}_{\substack{z\sim\mu_{\pi,\rho}\\u\sim\pi(\cdot|z)}}\log\scalebox{1.2}{$\frac{\pi(u|z)}{\tilde{\pi}(u)}$}-A) \\
&~~~~-\beta\cdot(\mathbb{E}_{\substack{z\sim\mu_{\pi,\rho}\\u\sim\pi(\cdot|z)\\\tau,\omega\sim\sigma(\cdot|z,u)}}\log\scalebox{1.2}{$\frac{\sigma(\tau,\omega|z,u)}{\tilde{\sigma}(\tau,\omega)}$}-B) \\
&~~~~-\eta\cdot(\mathbb{E}\hspace{-7pt}_{\substack{z\sim\mu_{\pi,\rho}\\u\sim\pi(\cdot|z)\\\tau,\omega\sim\sigma(\cdot|z,u)\\s^{\prime}\sim\tau(\cdot|s,u)\\x^{\prime}\sim\omega(\cdot|u,s^{\prime})\\z^{\prime}\sim\rho(\cdot|z,u,x^{\prime})}}\hspace{-8pt}\log\scalebox{1.2}{$\frac{\varrho_{\tau,\omega}(z^{\prime}|z,u)}{\tilde{\varrho}(z^{\prime})}$}-C) \\
&=
\mathbb{E}_{\substack{
z\sim\mu_{\pi,\rho}\\
s\sim p(\cdot|z)
}}\big[
\mathbb{E}_{u\sim\pi(\cdot|z)}\big[
\upsilon(s,u)-\alpha\cdot(\log\scalebox{1.2}{$\frac{\pi(u|z)}{\tilde{\pi}(u)}$}-A) \\
&~~~~+
\mathbb{E}_{\substack{
\tau,\omega\sim\sigma(\cdot|z,u)
}}\big[
-\beta\cdot(\log\scalebox{1.2}{$\frac{\sigma(\tau,\omega|z,u)}{\tilde{\sigma}(\tau,\omega)}$}-B) \\
&~~~~+
\mathbb{E}_{\substack{
s^{\prime}\sim\tau(\cdot|s,u)\\
x^{\prime}\sim\omega(\cdot|u,s^{\prime})\\
z^{\prime}\sim\rho(\cdot|z,u,x^{\prime})
}}\big[
-\eta\cdot(\log\scalebox{1.2}{$\frac{\varrho_{\tau,\omega}(z^{\prime}|z,u)}{\tilde{\varrho}(z^{\prime})}$}-C) \\
&~~~~+
\gamma
V(z^{\prime})
\big]\big]\big]
-
V(z)
\big]
+
\langle V,(1-\gamma)\rho_{0}\rangle
\end{aligned}\raisetag{0.88\baselineskip}
\end{gather}

Then taking the gradient w.r.t. $\mu$ and setting it to zero yields:
\begin{gather}
\begin{aligned}
V(z)
=
~\mathbb{E}_{\substack{
s\sim p(\cdot|z)\\
u\sim\pi(\cdot|z)
}}&\big[
-\alpha
\log
\scalebox{1.2}{$
\frac{\pi(u|z)}{\tilde{\pi}(u)}
$}
+
\upsilon(s,u)
+ \\[-2.0ex]
\mathbb{E}_{\tau,\omega\sim\sigma(\cdot|z,u)}&\big[
-\beta
\log
\scalebox{1.2}{$
\frac{\sigma(\tau,\omega|z,u)}{\tilde{\sigma}(\tau,\omega)}
$}
+ \\[-0.8ex]
\mathbb{E}_{\substack{s^{\prime}\sim\tau(\cdot|s,u)\\x^{\prime}\sim\omega(\cdot|u,s^{\prime})\\z^{\prime}\sim\rho_{\tau,\omega}(\cdot|z,u,x^{\prime})}}&\big[
-\eta
\log
\raisebox{-2pt}{\scalebox{1.2}{$
\frac{\varrho_{\tau,\omega}(z^{\prime}|z,u)}{\tilde{\varrho}(z^{\prime})}
$}}
+\pix
\gamma
V(z^{\prime})
\big]\big]\big]
\end{aligned}\raisetag{3.25\baselineskip}
\end{gather}
For uniqueness, observe as before that \smash{$\mathbb{B}_{\pi,\rho}$} is $\gamma$-contracting: \smash{$\|\mathbb{B}_{\pi,\rho}V-\mathbb{B}_{\pi,\rho}V^{\prime}\|_{\infty}\leq\gamma\|V-V^{\prime}\|_{\infty}$}; then appeal to the contraction mapping theorem for uniqueness of fixed point. The only change from before is the additional log terms, which---like the utility term---cancel out of the differences.

For Theorems \ref{thm:values} and \ref{thm:policies}, we give a single derivation for both:

\values*

\policies*

\textit{Proof}. From Proposition \ref{thm:prop}, the (state) value $V^{\phi_{\pi,\rho}}\in\mathbb{R}^{\mathcal{Z}}$ is:
\begin{gather}\label{eq:starv}
\begin{aligned}
V^{\phi_{\pi,\rho}}(z)
=
~\mathbb{E}_{\substack{
s\sim p(\cdot|z)\\
u\sim\pi(\cdot|z)
}}&\big[
-\alpha
\log
\scalebox{1.2}{$
\frac{\pi(u|z)}{\tilde{\pi}(u)}
$}
+
\upsilon(s,u)
+ \\[-2.0ex]
\mathbb{E}_{\tau,\omega\sim\sigma(\cdot|z,u)}&\big[
-\beta
\log
\scalebox{1.2}{$
\frac{\sigma(\tau,\omega|z,u)}{\tilde{\sigma}(\tau,\omega)}
$}
+ \\[-0.8ex]
\mathbb{E}_{\substack{s^{\prime}\sim\tau(\cdot|s,u)\\x^{\prime}\sim\omega(\cdot|u,s^{\prime})\\z^{\prime}\sim\rho_{\tau,\omega}(\cdot|z,u,x^{\prime})}}&\big[
-\eta
\log
\raisebox{-2pt}{\scalebox{1.2}{$
\frac{\varrho_{\tau,\omega}(z^{\prime}|z,u)}{\tilde{\varrho}(z^{\prime})}
$}}
\hspace{-14pt}\raisebox{-13pt}{$+\pix
\gamma
V^{\phi_{\pi,\rho}}(z^{\prime})
\big]\big]\big]$}
\end{aligned}\raisetag{3.25\baselineskip}
\end{gather}
Define (state-action) $Q^{\phi_{\pi,\rho}}$\pix$\in$\pix$\mathbb{R}^{\mathcal{Z}\times\mathcal{U}}$ to be ahead by $\nicefrac{1}{3}$ steps:
\begin{gather}\label{eq:starq}
\begin{aligned}
Q^{\phi_{\pi,\rho}}(z,u)
\doteq
\mathbb{E}_{\substack{
s\sim p(\cdot|z)
}}&\big[
\upsilon(s,u) + \\[-0.25ex]
\mathbb{E}_{\tau,\omega\sim\sigma(\cdot|z,u)}&\big[
-\beta
\log\scalebox{1.2}{$\frac{\sigma(\tau,\omega|z,u)}{\tilde{\sigma}(\tau,\omega)}$}
+ \\[-0.8ex]
\mathbb{E}_{\substack{s^{\prime}\sim\tau(\cdot|s,u)\\x^{\prime}\sim\omega(\cdot|u,s^{\prime})\\z^{\prime}\sim\rho_{\tau,\omega}(\cdot|z,u,x^{\prime})}}&\big[
-\eta
\log
\raisebox{-2pt}{\scalebox{1.2}{$
\frac{\varrho_{\tau,\omega}(z^{\prime}|z,u)}{\tilde{\varrho}(z^{\prime})}
$}}
\hspace{-19pt}\raisebox{-17pt}{$+\pix
\gamma
V^{\phi_{\pi,\rho}}(z^{\prime})
\big]\big]\big]$}
\end{aligned}\raisetag{3.25\baselineskip}
\end{gather}
and (state-action-model) $K^{\phi_{\pi,\rho}}$\pix$\in$\pix$\mathbb{R}^{\mathcal{Z}\times\mathcal{U}\times\mathcal{T}\times\mathcal{O}}$ by $\nicefrac{2}{3}$ steps:
\begin{gather}\label{eq:stark}
\begin{aligned}
K^{\phi_{\pi,\rho}}(z,u,\tau,\omega)
\doteq~~~~~\pix& \\
\mathbb{E\hspace{-7pt}}_{\substack{s\sim p(\cdot|z)\\s^{\prime}\sim\tau(\cdot|s,u)\\x^{\prime}\sim\omega(\cdot|u,s^{\prime})\\z^{\prime}\sim\rho_{\tau,\omega}(\cdot|z,u,x^{\prime})}}\hspace{-7pt}&\big[
-\eta
\log
\raisebox{-2pt}{\scalebox{1.2}{$
\frac{\varrho_{\tau,\omega}(z^{\prime}|z,u)}{\tilde{\varrho}(z^{\prime})}
$}}
\hspace{-19pt}\raisebox{-17pt}{$+\pix
\gamma
V^{\phi_{\pi,\rho}}(z^{\prime})
\big]\big]\big]$}
\end{aligned}\raisetag{3.25\baselineskip}
\end{gather}
\vspace{-1.4em}

\dayum{
The decision and recognition policies seek the optimizations:}
\vspace{-0.8em}
\begin{gather}
\begin{aligned}
&\text{extremize}_{\pi}V^{\phi_{\pi,\rho}}(z) \\
&\pix~~~~~~\text{s.t.}~~~~~~~
\mathbb{E}_{u\sim\pi(\cdot|z)}1=1
\end{aligned}
\end{gather}
\vspace{-1.4em}
\begin{gather}
\begin{aligned}
&\text{extremize}_{\sigma}Q^{\phi_{\pi,\rho}}(z,u) \\
&\pix~~~~~~\text{s.t.}~~~~~~~
\mathbb{E}_{\tau,\omega\sim\sigma(\cdot|z,u)}1=1
\end{aligned}
\end{gather}
Equations \ref{eq:starv}--\ref{eq:stark} are true in particular for optimal values, so
\begin{gather}
\begin{aligned}
V^{*}(z)=\mathbb{E}&_{u\sim\pi^{*}(\cdot|z)}\big[-\alpha\log\scalebox{1.2}{$\frac{\pi^{*}(u|z)}{\tilde{\pi}(u)}$}+Q^{*}(z,u)\big]
\end{aligned}\raisetag{1.025\baselineskip}
\end{gather}
\vspace{-2em}
\begin{gather}
\begin{aligned}
Q^{*}(z,u)&=\mathbb{E}_{s\sim p(\cdot|z)}\big[\upsilon(s,u)\big]+\mathbb{E}_{\tau,\omega\sim\sigma^{*}(\cdot|z,u)}\big[~~ \\
&-\beta\log\scalebox{1.2}{$\frac{\sigma^{*}(\tau,\omega|z,u)}{\tilde{\sigma}(\tau,\omega)}$}+K^{*}(z,u,\tau,\omega)\big]
\end{aligned}
\end{gather}
Therefore for the extremizations we write the Lagrangians
\begin{gather}
\begin{aligned}
\mathcal{L}(\pi^*,\lambda)\doteq V^{*}(z)+\lambda\cdot(\mathbb{E}_{u\sim\pi^*(\cdot|z)}1-1)
\end{aligned}
\end{gather}
\vspace{-2.2em}
\begin{gather}
\begin{aligned}
\mathcal{L}(\sigma^*,\nu)\doteq Q^{*}(z,u)+\nu\cdot(\mathbb{E}_{\tau,\omega\sim\sigma^*(\cdot|z,u)}1-1)
\end{aligned}
\end{gather}
Straightforward algebraic manipulation yields the policies:
\begin{equation}
\pi^{*}(u|z)
=
\scalebox{1.2}{$\frac{\tilde{\pi}(u_{t})}{Z_{Q^{*}}(z)}$}
\exp\big(\tfrac{1}{\alpha}Q^{*}(z,u)\big)
\end{equation}
\vspace{-2em}
\begin{equation}
\sigma^{*}(\tau,\omega|z,u)
=
\scalebox{1.2}{$\frac{\tilde{\sigma}(\tau,\omega)}{Z_{K^{*}}(z,u)}$}
\exp\big(\tfrac{1}{\beta}K^{*}(z,u,\tau,\omega)\big)
\end{equation}
where partition functions $Z_{Q^{*}}(z)$ and $Z_{K^{*}}(z)$ are given by:
\begin{gather}
\begin{aligned}
Z_{Q^{*}}(z)=\mathbb{E}_{u\sim\tilde{\pi}}\exp(\tfrac{1}{\alpha}Q^{*}(z,u))
\end{aligned}
\end{gather}
\vspace{-2em}
\begin{gather}
\begin{aligned}
Z_{K^{*}}(z,u)=\mathbb{E}_{\tau,\omega\sim\tilde{\sigma}}\exp(\tfrac{1}{\beta}K^{*}(z,u,\tau,\omega))
\end{aligned}
\end{gather}
which proves Theorem \ref{thm:policies}. Then Theorem \ref{thm:values} is obtained by plugging back into the backward recursion (Proposition \ref{thm:prop}).

\vspace{0.5em}
For uniqueness, we want \smash{$\|\mathbb{B}V$$-$$\mathbb{B}V^{\prime}\|_{\infty}$$\leq$$\gamma\|V$$-$$V^{\prime}\|_{\infty}$}. Let \smash{$\|V$\hspace{-0.5pt}$-$$V^{\prime}\|_{\infty}$$=$\pix$\varepsilon$} (\smash{$\text{max}_{z^{\prime}}\hspace{-1pt}|V\hspace{-0.5pt}(z^{\prime})$$-$$V^{\prime}(z^{\prime})|$$=$\pix$\varepsilon$}). Now, \smash{$(\mathbb{B}^{*}V)(z)$}
\begin{gather}
\begin{aligned}
&\doteq
\alpha\log\mathbb{E}_{u\sim\tilde{\pi}}\big[\exp\big(\tfrac{1}{\alpha}\big(
\mathbb{E}_{s\sim p(\cdot|z)}\upsilon(s,u) \\
&~~~~+
\beta\log\mathbb{E}_{\tau,\omega\sim\tilde{\sigma}}\big[\exp\big(\tfrac{1}{\beta}
\mathbb{E}_{z^{\prime}\sim\varrho_{\tau,\omega}(\cdot|z,u)}
\big[ \\
&~~~~-\eta\log\scalebox{1.2}{$\frac{\varrho_{\tau,\omega}(z^{\prime}|z,u)}{\tilde{\varrho}(z^{\prime})}$}+\pix\gamma V(z^{\prime})\big]
\big)\big]\big)
\big)\big] \\
&\leq
\alpha\log\mathbb{E}_{u\sim\tilde{\pi}}\big[\exp\big(\tfrac{1}{\alpha}\big(
\mathbb{E}_{s\sim p(\cdot|z)}\upsilon(s,u) \\
&~~~~+
\beta\log\mathbb{E}_{\tau,\omega\sim\tilde{\sigma}}\big[\exp\big(\tfrac{1}{\beta}
\mathbb{E}_{z^{\prime}\sim\varrho_{\tau,\omega}(\cdot|z,u)}
\big[ \\
&~~~~-\eta\log\scalebox{1.2}{$\frac{\varrho_{\tau,\omega}(z^{\prime}|z,u)}{\tilde{\varrho}(z^{\prime})}$}+\pix\gamma(V^{\prime}(z^{\prime})+\varepsilon)\big]
\big)\big]\big)
\big)\big] \\
&=
\alpha\log\mathbb{E}_{u\sim\tilde{\pi}}\big[\exp\big(\tfrac{1}{\alpha}\big(
\mathbb{E}_{s\sim p(\cdot|z)}\upsilon(s,u) \\
&~~~~+
\beta\log\mathbb{E}_{\tau,\omega\sim\tilde{\sigma}}\big[\exp\big(\tfrac{1}{\beta}\gamma\varepsilon+\tfrac{1}{\beta}
\mathbb{E}_{z^{\prime}\sim\varrho_{\tau,\omega}(\cdot|z,u)}
\big[ \\
&~~~~-\eta\log\scalebox{1.2}{$\frac{\varrho_{\tau,\omega}(z^{\prime}|z,u)}{\tilde{\varrho}(z^{\prime})}$}+\pix\gamma V^{\prime}(z^{\prime})\big]
\big)\big]\big)
\big)\big] \\
&=
\alpha\log\mathbb{E}_{u\sim\tilde{\pi}}\big[\exp\big(\tfrac{1}{\alpha}\big(
\mathbb{E}_{s\sim p(\cdot|z)}\upsilon(s,u) \\
&~~~~+
\beta\log\big(\exp(\tfrac{1}{\beta}\gamma\varepsilon)\mathbb{E}_{\tau,\omega\sim\tilde{\sigma}}\big[\exp\big(\tfrac{1}{\beta}
\mathbb{E}_{z^{\prime}\sim\varrho_{\tau,\omega}(\cdot|z,u)}
\big[ \\
&~~~~-\eta\log\scalebox{1.2}{$\frac{\varrho_{\tau,\omega}(z^{\prime}|z,u)}{\tilde{\varrho}(z^{\prime})}$}+\pix\gamma V^{\prime}(z^{\prime})\big]
\big)\big]\big)\big)
\big)\big] \\
&=
\alpha\log\mathbb{E}_{u\sim\tilde{\pi}}\big[\exp\big(\tfrac{1}{\alpha}\gamma\varepsilon+\tfrac{1}{\alpha}\big(
\mathbb{E}_{s\sim p(\cdot|z)}\upsilon(s,u) \\
&~~~~+
\beta\log\mathbb{E}_{\tau,\omega\sim\tilde{\sigma}}\big[\exp\big(\tfrac{1}{\beta}
\mathbb{E}_{z^{\prime}\sim\varrho_{\tau,\omega}(\cdot|z,u)}
\big[ \\
&~~~~-\eta\log\scalebox{1.2}{$\frac{\varrho_{\tau,\omega}(z^{\prime}|z,u)}{\tilde{\varrho}(z^{\prime})}$}+\pix\gamma V^{\prime}(z^{\prime})\big]
\big)\big]\big)
\big)\big] \\
&=
\alpha\log\big(\exp(\tfrac{1}{\alpha}\gamma\varepsilon)\mathbb{E}_{u\sim\tilde{\pi}}\big[\exp\big(\tfrac{1}{\alpha}\big(
\mathbb{E}_{s\sim p(\cdot|z)}\upsilon(s,u) \\
&~~~~+
\beta\log\mathbb{E}_{\tau,\omega\sim\tilde{\sigma}}\big[\exp\big(\tfrac{1}{\beta}
\mathbb{E}_{z^{\prime}\sim\varrho_{\tau,\omega}(\cdot|z,u)}
\big[ \\
&~~~~-\eta\log\scalebox{1.2}{$\frac{\varrho_{\tau,\omega}(z^{\prime}|z,u)}{\tilde{\varrho}(z^{\prime})}$}+\pix\gamma V^{\prime}(z^{\prime})\big]
\big)\big]\big)
\big)\big]\big) \\
&=
\gamma\varepsilon+\alpha\log\mathbb{E}_{u\sim\tilde{\pi}}\big[\exp\big(\tfrac{1}{\alpha}\big(
\mathbb{E}_{s\sim p(\cdot|z)}\upsilon(s,u) \\
&~~~~+
\beta\log\mathbb{E}_{\tau,\omega\sim\tilde{\sigma}}\big[\exp\big(\tfrac{1}{\beta}
\mathbb{E}_{z^{\prime}\sim\varrho_{\tau,\omega}(\cdot|z,u)}
\big[ \\
&~~~~-\eta\log\scalebox{1.2}{$\frac{\varrho_{\tau,\omega}(z^{\prime}|z,u)}{\tilde{\varrho}(z^{\prime})}$}+\pix\gamma V^{\prime}(z^{\prime})\big]
\big)\big]\big)
\big)\big] \\
&=
\gamma\varepsilon+(\mathbb{B}^{*}V^{\prime})(z)
\end{aligned}\raisetag{0.88\baselineskip}
\end{gather}
\dayum{%
Likewise, we can show that \smash{$(\mathbb{B}^{*}V)(z)\geq(\mathbb{B}^{*}V^{\prime})(z)-\gamma\varepsilon$}. Hence $\text{max}_{z}|(\mathbb{B}V)(z)-(\mathbb{B}V^{\prime})(z)|=\|\mathbb{B}V-\mathbb{B}V^{\prime}\|_{\infty}\leq\gamma\epsilon$.}

\clearpage
\onecolumn

\dayum{
\textbf{Note on Equation \ref{eq:24}}: Note that we originally formulated ``soft policy matching'' in Table \ref{tab:inverse} as a forward Kullback-Leibler divergence expression. However, analogously to maximum likelihood in supervised learning, the entropy terms drop out of the optimization, which yields Equation \ref{eq:24}. To see this, note that the causally-conditioned probability is simply the product of conditional probabilities at each time step, and each conditional is ``Markovianized'' using beliefs $z_{t}$ (i.e. Equation \ref{eq:likelihood}).}

\section{Illustrative Trajectories}\label{app:d}

\begin{figure*}[h!]
\centering
\tikzset{
mark0/.pic={\path[fill=blue] (0,0) circle[radius=#1];},
mark1/.pic={
\draw[scale=1.2,rotate=45,color=red] (-#1,0)--(#1,0);
\draw[scale=1.2,rotate=45,color=red] (0,-#1)--(0,#1);},
mark2/.pic={\fill[scale=1.2] (-#1,0)--(0,-#1)--(#1,0)--(0,#1)--cycle;},
arrow/.style={thin,-stealth,shorten <=2,shorten >=2,overlay}}
\newcommand{\tikzsimplex}{
\coordinate (s1) at (0:0);
\coordinate (s2) at (0:100);
\coordinate (s3) at (60:100);

\coordinate (d0) at (54.2,79.0);
\coordinate (d1) at (51.3,80.0);
\coordinate (d2) at (49.1,77.2);
\coordinate (d3) at (49.1,70.7);
\coordinate (d4) at (51.6,67.9);
\coordinate (d5) at (50.0,65.1);
\coordinate (d6) at (45.6,65.1);
\coordinate (d7) at (43.1,67.4);
\coordinate (d8) at (22.7,31.5);
\coordinate (d9) at (28.0,19.9);
\coordinate (da) at (54.0,19.9);
\coordinate (db) at (58.4,15.6);
\coordinate (dc) at (76.4,15.6);
\coordinate (dd) at (79.1,13.2);
\coordinate (de) at (76.6,8.9);
\coordinate (df) at (80.8,1.6);
\coordinate (dg) at (97.2,1.6);
\coordinate (dh) at (98.0,2.9);

\path[pattern=custom dots,pattern color=red!15] (s1) -- (s3) {[rounded corners=2.4] -- (d0) -- (d1) -- (d2) -- (d3) -- (d4) -- (d5) -- (d6) -- (d7) -- (d8) -- (d9) -- (da) -- (db) -- (dc) -- (dd) -- (de) -- (df) -- (dg) -- (dh)} -- (s2) -- (s1);
\path[pattern=custom lines,pattern color=blue!10] {[rounded corners=2.4] (d0) -- (d1) -- (d2) -- (d3) -- (d4) -- (d5) -- (d6) -- (d7) -- (d8) -- (d9) -- (da) -- (db) -- (dc) -- (dd) -- (de) -- (df) -- (dg) -- (dh)} -- (d0);
\draw[rounded corners=2.4,dash pattern=on 1.2 off .8] (d0) -- (d1) -- (d2) -- (d3) -- (d4) -- (d5) -- (d6) -- (d7) -- (d8) -- (d9) -- (da) -- (db) -- (dc) -- (dd) -- (de) -- (df) -- (dg) -- (dh);

\draw (s1) -- (s2) -- (s3) -- cycle;
\draw (s1) node[below,fonttiny,xshift=1,yshift=1,overlay] {NL};
\draw (s2) node[below,fonttiny,xshift=-10,yshift=1,overlay] {Dementia};
\draw (s3) node[above,fonttiny,yshift=-1] {MCI};
\path (0,-6) -- (100,-6);}

\centering
\begin{tikzpicture}
\draw[yshift=1,pattern=custom dots,pattern color=red!30] (-6.0,-3.6) rectangle (6.0,3.6);
\draw[yshift=12+1,pattern=custom lines,pattern color=blue!20] (-6.0,-3.6) rectangle (6.0,3.6);
\node[xshift=4,right] at (0,0) {An MRI is less likely to be ordered.};
\node[xshift=4,right] at (0,12) {An MRI is more likely to be ordered.};
\path (165,0) pic[ultra thick] {mark1=2};
\path (165,12) pic {mark0=2};
\node[xshift=1,right] at (165,0) {An MRI is not ordered.};
\node[xshift=1,right] at (165,12) {An MRI is ordered.};
\draw[xshift=273,thick,-stealth] (-6,0) -- (6,0);
\path (273,12) pic {mark2=2};
\node[xshift=4,right] at (273,0) {Belief updates};
\node[xshift=4,right] at (273,12) {Final beliefs};
\draw[xshift=351,thick,dash pattern=on 2.4 off 1.6] (-6,0) -- (6,0);
\draw[xshift=351,yshift=12-1] (90:5)--(210:5)--(330:5)--cycle;
\node[xshift=4,right] at (351,0) {Decision boundary};
\node[xshift=4,right] at (351,12) {Belief simplex};
\end{tikzpicture}

\subfloat[Patient treated as if ``rationally'']{
\centering
\begin{tikzpicture}[scale=1.2]
\tikzsimplex
\coordinate (b1) at (60.2,17.8);
\coordinate (b2) at (50.5,62.5);
\coordinate (b3) at ($ (49.4,76.6) - (3,0) $);
\coordinate (b4) at (54.3,64.8);
\coordinate (b5) at (49.9,77.1);
\path (b1) pic {mark0=1};
\path (b2) pic {mark0=1};
\path (b3) pic[thick] {mark1=1};
\path (b4) pic {mark0=1};
\path (b5) pic {mark2=1};
\draw[arrow] (b1) to[bend left=15] (b2);
\draw[arrow] (b2) to[bend left=30] (b3);
\draw[arrow] (b3) to[bend left=15] (b4);
\draw[arrow] (b4) to[bend right=45,looseness=1.25] (b5);
\end{tikzpicture}
}
\hfill
\subfloat[Patient not treated ``rationally'']{
\centering
\begin{tikzpicture}[scale=1.2]
\tikzsimplex
\coordinate (b1) at (60.2,17.8);
\coordinate (b2) at (63.5,37.2);
\coordinate (b3) at (61.6,46.9);
\coordinate (b4) at (71.9,25.3);
\coordinate (b5) at (77.5,16.7);
\path (b1) pic[thick] {mark1=1};
\path (b2) pic[thick] {mark1=1};
\path (b3) pic[thick] {mark1=1};
\path (b4) pic[thick] {mark1=1};
\path (b5) pic {mark2=1};
\draw[arrow] (b1) to[bend right=15] (b2);
\draw[arrow] (b2) to[bend left=75,looseness=2] (b3);
\draw[arrow] (b3) to[bend left=45] (b4);
\draw[arrow] (b4) to[bend left=60,looseness=1.75] (b5);
\end{tikzpicture}
}
\hfill
\subfloat[Patient~is diagnosed~belatedly]{
\centering
\begin{tikzpicture}[scale=1.2]
\tikzsimplex
\coordinate (b1) at (60.2,17.8);
\coordinate (b2) at (63.5,37.2);
\coordinate (b3) at (61.6,46.9);
\coordinate (b4) at (77.1,19.4);
\coordinate (b5) at (84.6,10.2);
\path (b1) pic[thick] {mark1=1};
\path (b2) pic[thick] {mark1=1};
\path (b3) pic {mark0=1};
\path (b4) pic {mark0=1};
\path (b5) pic {mark2=1};
\draw[arrow] (b1) to[bend right=15] (b2);
\draw[arrow] (b2) to[bend left=75,looseness=2] (b3);
\draw[arrow] (b3) to[bend left=45] (b4);
\draw[arrow] (b4) to[bend left=60,looseness=1.75] (b5);
\end{tikzpicture}
}
\caption{\textit{Decision Trajectories}. Examples of apparent beliefs and actions of a clinical decision-maker regarding real patients, including cases where: (a)~the clinician's decisions coincide with those that would have been dictated by a ``perfectly-rational'' policy---despite their bounded rationality; (b)~the clinician fails to make ``perfectly-rational'' decisions (in this context, the ``boundedness'' of the clinician could be due to any number of issues encountered during the diagnostic process); and (c)~a patient who---apparently---could have been diagnosed much earlier than they actually were, but for the clinician not having followed the decisions prescribed by the ``perfectly-rational'' policy.}
\label{fig:patients}
\end{figure*}
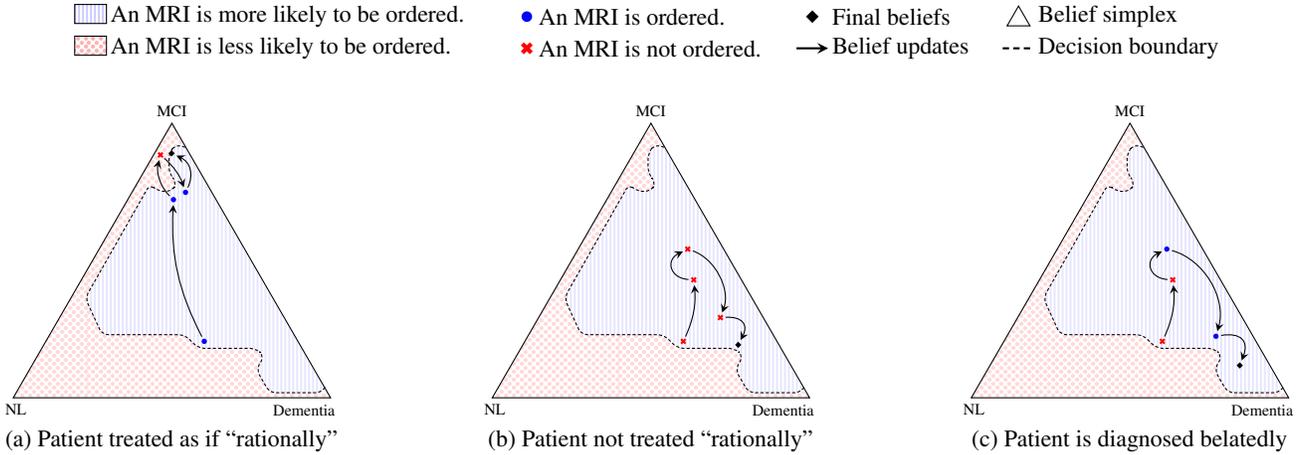

\dayum{
Here we direct attention to the potential utility of IBRC (and---more generally---instantiations of the IDM paradigm) as an ``investigative device'' for auditing and quantifying individual decisions. In Figure \ref{fig:patients}, we see that modeling the evolution of a decision-maker's subjective beliefs provides a concrete basis for analyzing the corresponding sequence of actions chosen. Each vertex of the belief simplex corresponds to one of the three stable Alzheimer's diagnoses, and each point within the simplex corresponds to a unique belief (i.e. probability distribution). The closer the point is to a vertex (i.e. disease state), the higher the probability assigned to that state. For instance, if the belief is located exactly in the middle of the simplex (i.e. equidistant from all vertices), then all states are believed to be equally likely. Note that this is visual presentation is done similarly to \cite{huyuk2021explaining}, where decision trajectories within belief simplices are first visualized in this manner---with the core difference here being that the decision policies (hence decision boundaries thereby induced) are computed using a different technique.}

\section{Summary of Notation}\label{app:0}

\newcolumntype{A}{>{\centering\arraybackslash}m{2.2cm}}
\newcolumntype{B}{>{\centering\arraybackslash}m{4.2cm}}
\newcolumntype{C}{>{\centering\arraybackslash}m{2.2cm}}
\newcolumntype{D}{>{\centering\arraybackslash}m{2.2cm}}
\newcolumntype{E}{>{\centering\arraybackslash}m{4.2cm}}
\newcolumntype{F}{>{\centering\arraybackslash}m{2.2cm}}
\begin{table*}[h!]\small
\setlength\tabcolsep{0pt}
\renewcommand{\arraystretch}{0.95}
\begin{center}
\begin{tabular}{ABC|DEF}
\toprule
Notation & Meaning & (first defined in)
&
Notation & Meaning & (first defined in)
\\
\midrule
$\psi$ & problem setting& Section \ref{sub:forward}
&
$s$ & environment state& Section \ref{sub:forward}
\\
$x$ & environment emission& Section \ref{sub:forward}
&
$z$ & agent state, i.e. belief& Section \ref{sub:forward}
\\
$u$ & agent emission, i.e. action& Section \ref{sub:forward}
&
$\tau_{\text{env}}$ & environment transition& Section \ref{sub:forward}
\\
$\tau$ & subjective transition& Section \ref{sub:forward}
&
$\omega_{\text{env}}$ & environment emission& Section \ref{sub:forward}
\\
$\omega$ & subjective emission& Section \ref{sub:forward}
&
$\upsilon$ & utility (i.e. reward) function& Section \ref{sub:forward}
\\
$\gamma$ & discount factor& Section \ref{sub:forward}
&
$\phi$ & behavior& Section \ref{sub:forward}
\\
$\phi_{\text{demo}}$ & demonstrated behavior& Section \ref{sub:inverse}
&
$\phi_{\text{imit}}$ & imitation behavior& Section \ref{sub:inverse}
\\
$\theta$ & planning parameter& Section \ref{sub:forward}
&
$\theta_{\text{norm}}$ & normative parameter& Section \ref{sub:inverse}
\\
$\theta_{\text{desc}}$ & descriptive parameter& Section \ref{sub:inverse}
&
$\pi$ & decision policy& Section \ref{sub:forward}
\\
$\rho$ & recognition policy& Section \ref{sub:forward}
&
$\sigma$ & specification policy& Section \ref{sub:rational}
\\
$F$ & forward planner& Section \ref{sub:forward}
&
$G$ & inverse planner& Section \ref{sub:inverse}
\\
$\alpha^{-1}$ & flexibility coefficient& Section \ref{sub:bounded}
&
$\beta^{-1}$ & optimism coefficient& Section \ref{sub:bounded}
\\
$\eta^{-1}$ & adaptivity coefficient& Section \ref{sub:bounded}
&
$\tilde{\pi}$ & action prior& Section \ref{sub:bounded}
\\
$\tilde{\sigma}$ & model prior& Section \ref{sub:bounded}
&
$\tilde{\varrho}$ & belief prior& Section \ref{sub:bounded}
\\
\bottomrule
\end{tabular}
\end{center}
\end{table*}
 
\twocolumn

\clearpage
\bibliographystyle{unsrt}

\balance\bibliography{
bib/0-imitate,
bib/1-reinforce,
bib/2-entropy,
bib/3-information,
bib/4-constraints,
bib/5-risk,
bib/6-intrinsic,
bib/7-bounded,
bib/8-identification,
bib/9-interpret,
bib/a-miscellaneous
}

\end{document}